\ifdefined\XeTeXversion\else\ifdefined\pdfoutput
\pdfoutput=1
\fi\fi
\documentclass[10pt,twocolumn,letterpaper]{article}
\usepackage[pagenumbers]{cvpr}
\usepackage{multirow}
\usepackage{array}
\usepackage{microtype}
\usepackage{tikz}
\usepackage{fontawesome5}
\usepackage{enumitem}
\providecolor{tbimg}{HTML}{1F78B4}
\providecolor{tbtxt}{HTML}{7570B3}
\providecolor{tbours}{HTML}{D95F02}
\newcommand\tbicon[1][1.9ex]{%
\begin{tikzpicture}[x=#1,y=#1,baseline=0.14*#1,line cap=round,line join=round]
  \foreach \i in {0,1,2} {
    \fill[tbimg,rounded corners=0.035*#1] (0,\i*0.34) rectangle ++(0.3,0.27);
    \fill[tbtxt,rounded corners=0.035*#1] (1.0,\i*0.34) rectangle ++(0.3,0.27);}
  \draw[tbours,line width=0.105*#1] (0.3,0.36) -- (1.0,0.36);
  \draw[tbours,line width=0.07*#1] (0.3,0.9) .. controls (0.5,0.42) and (0.8,0.42) .. (1.0,0.9);
  \foreach \x/\y in {0.47/0.66,0.65/0.56,0.83/0.66} {\draw[tbours,line width=0.045*#1] (\x,0.36) -- (\x,\y);}
\end{tikzpicture}}

\newcommand{\ours}{TowerBridge\xspace}

\newcommand{\oursi}{\mbox{\tbicon\hspace{0.12em}TowerBridge}\xspace}
\newcommand{\astra}{Astra\xspace}

\newcommand{\dD}{\mathbf{D}}
\newcommand{\ci}[2]{\,[\ensuremath{#1},\,\ensuremath{#2}]}
\newcommand{\tabfont}{\footnotesize\setlength{\tabcolsep}{4pt}\renewcommand{\arraystretch}{1.05}\tabnums}
\newcommand{\tabdense}{\scriptsize\setlength{\tabcolsep}{2.5pt}\renewcommand{\arraystretch}{1.12}\tabnums}
\newcommand{\tabnums}{}%
\newcommand{\hd}[1]{\multicolumn{1}{c}{#1}}
\newcommand{\cit}[2]{{\scriptsize[\ensuremath{#1},\,\ensuremath{#2}]}}
\newcommand{\mk}[1]{\rlap{$^{#1}$}}
\newcommand{\gp}{\hspace{0.8em}}
\newcolumntype{E}{r@{\extracolsep{0pt}\hspace{0.35em}}l@{\extracolsep{\fill}}}
\definecolor{oursc}{HTML}{D95F02}
\definecolor{quotebg}{HTML}{FFF4EA}
\usepackage{iftex}
\ifXeTeX
  \usepackage{fontspec}
  \edef\next{\noexpand\DeclareSymbolFont{tabnum}{TU}{\rmdefault}{m}{n}\noexpand\DeclareSymbolFont{tabnumbf}{TU}{\rmdefault}{b}{n}}\next
  \newcommand{\tabnumcodes}[1]{%
    \count255=`\0 \loop\Umathcode\count255="0 #1 \count255 \ifnum\count255<`\9 \advance\count255 1 \repeat
    \Umathcode`\.="0 #1 `\. \Umathcode`\,="6 #1 `\, \Umathcode`\/="0 #1 `\/
    \Umathcode`\+="2 #1 `\+ \Umathcode`\-="2 #1 "2212 }
  \renewcommand{\tabnums}{\tabnumcodes{\symtabnum}\renewcommand{\mathbf}[1]{{\tabnumcodes{\symtabnumbf}##1}}}
\fi

\definecolor{cvprblue}{rgb}{0.21,0.49,0.74}
\usepackage[pagebackref,breaklinks,colorlinks,allcolors=cvprblue]{hyperref}

\def\confName{CVPR}
\def\confYear{2027}

\title{\texorpdfstring{\tbicon[2.1ex]\hspace{0.15em}}{}TowerBridge: Bridging the Dual-Encoder Modality Gap\\
via Weight-Space Programming}

\author{Hani Alomari \quad Md Atabuzzaman \quad Chris Thomas\\
Department of Computer Science\\
Virginia Tech\\
{\tt\small \{hani, atabuzzaman, christhomas\}@vt.edu}}

\begin{document}
\twocolumn[{%
\renewcommand\twocolumn[1][]{#1}%
\maketitle
\vspace{-1.6em}
\begin{center}
  \centering
  \includegraphics[width=0.95\textwidth]{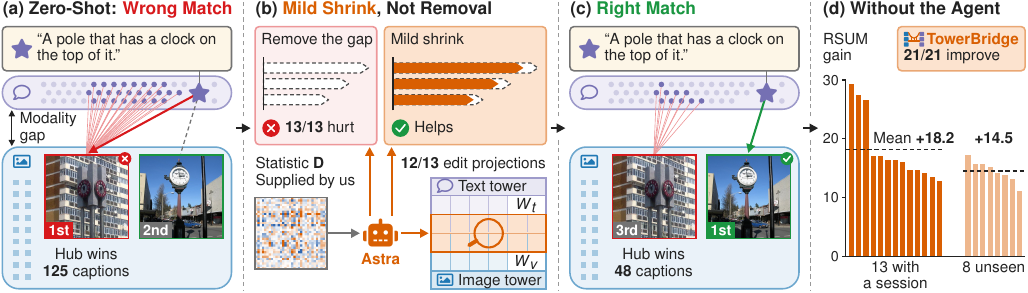}
  \vspace{-0.6em}
  \captionof{figure}{%
  \textbf{A mild shrink of hard-negative directions improves retrieval, whereas removing the modality gap hurts.}
  (a)~A caption retrieves a \emph{hub} photo instead of the correct match.
  (b)~Guided by our paired statistic $\dD$, \astra finds that removing the gap hurts in all 13 sessions, whereas a mild shrink helps; 12 of 13 patches edit an output projection.
  (c)~\astra's ViT-B/32 patch ranks the correct photo first and reduces hub wins.
  (d)~\oursi improves COCO test retrieval on all 21 CLIP checkpoints without the agent.}
  \label{fig:teaser}
\end{center}
\vspace{0.4em}
}]

\begin{abstract}
CLIP-style dual encoders map images and text through two separate towers into a shared embedding space, where the two modalities occupy distinct regions separated by a modality gap.
Retrieval fails when wrong matches, or hard negatives, outrank the correct one, and the gap is the usual suspect.
Yet removing it post hoc degrades retrieval on every CLIP checkpoint we test.
We introduce a paired statistic that identifies directions where the two towers agree more on hard negatives than on matched pairs.
Mildly attenuating these directions improves retrieval while largely preserving the gap.
We arrive at this correction through weight-space programming: guided by the statistic, an LLM agent inspects both towers and writes closed-form weight edits without backpropagation.
Free to edit any component, the agent improves COCO test retrieval on all 13 CLIP checkpoints and edits an output projection in 12 of the 13.
We distill and refine its recurring correction into \oursi, a closed-form recipe motivated by a first-order analysis of the hard-negative margin.
Computed in seconds from 25{,}000 labelled pairs, it requires no model-specific tuning and leaves the architecture and inference cost unchanged.
\oursi{} improves COCO test retrieval by 16.8 RSUM and Flickr30k test retrieval by 6.6 on average across all 21 evaluated CLIP checkpoints, including 8 unseen by the agent, while limiting any drop in ImageNet zero-shot accuracy to 0.5 points.
It also reduces hubness on all 21 checkpoints and improves retrieval on 4 of 5 dual encoders from other families.
\end{abstract}

\section{Introduction}
\label{sec:intro}

Image--text retrieval commonly relies on dual encoders such as CLIP~\cite{clip,openclip}.
Yet a caption can retrieve the wrong image even when its correct match is available (Fig.~\ref{fig:teaser}).
These high-scoring incorrect matches are known as \emph{hard negatives}.
For OpenAI's ViT-B/16, 41\% of text-to-image misses on COCO concentrate on \emph{hub} images that are retrieved for many captions they do not match~\cite{radovanovic2010hubs,qbnorm}.
Each tower maps its input into a shared embedding space, yet the two modalities occupy distinct regions separated by a \emph{modality gap}~\cite{modalitygap}, the usual suspect for such errors.

A natural remedy is to close it.
However, reducing it~\cite{an2025i0t} does not reliably improve zero-shot classification~\cite{sato2026hubness} or retrieval~\cite{grassucci2026closing}, and in our experiments removing it degrades retrieval on every checkpoint.
Fine-tuning~\cite{lora} improves retrieval but requires gradient-based optimization and hyperparameter tuning, while test-time re-ranking~\cite{qbnorm,csls,invsoftmax,nnn,dn} requires a bank of queries or reference samples and leaves the model unchanged.
It remains unclear what should change in the weights where the two towers meet.

In this paper, we show that the directions favoring hard negatives should be mildly attenuated rather than removed (Fig.~\ref{fig:teaser}).
We introduce a paired statistic $\dD$ that identifies these \emph{hard-negative directions}, where the two towers agree more on hard negatives than on matched pairs.
A 10\% shrink of these directions gains $+12.5$ COCO validation RSUM on average, against at most $+0.3$ for as many top principal directions chosen without pair labels.
The two leading directions align with the common mean and the modality gap.
Removing them degrades retrieval on every checkpoint, whereas shrinking them by 10\% improves every one.
These directions carry useful content, which a mild shrink retains.
Applied to the hard-negative directions, it widens hard-negative margins and reduces hubness while largely preserving the gap's size.
We call this effect \emph{bridging}.

We arrive at this correction through \emph{weight-space programming}~\cite{wsp_blog}: an LLM agent, \astra, inspects both towers and writes short programs that directly edit their weights without gradients.
The agent can edit any component of either tower, from attention heads to output projections.
We run one session on each of 13 CLIP checkpoints, and each produces a \emph{patch}, the set of edits the agent keeps.
We design and supply $\dD$ as evidence that the weights alone do not reveal, and the agent chooses where and how to intervene.
In controls, removing $\dD$ reduces the agent's gains more than removing access to the raw weights (Sec.~\ref{sec:f1}).
Of its 13 final patches, 12 rewrite an output projection, and most mildly attenuate the directions identified by $\dD$.

\begin{figure*}[t]
  \centering
  \includegraphics[width=\linewidth]{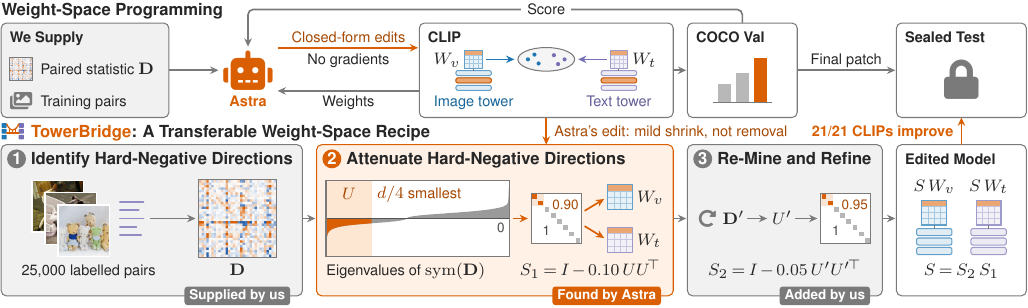}
  \caption{\textbf{From an agent session to the \oursi recipe.}
  \emph{Top:} \astra inspects both towers and writes closed-form weight edits using our paired statistic $\dD$ and COCO validation feedback. \astra's final patch is evaluated once on sealed test sets.
  \emph{Bottom:} \ours combines the agent's projection shrink with our re-mining and refinement step in an agent-free recipe that improves all 21 CLIP checkpoints on COCO test. $\dD$ and its spectrum are real.}
  \label{fig:method}
\end{figure*}

We distill and refine this correction into \oursi, a closed-form recipe with three steps (Fig.~\ref{fig:method}).
It computes $\dD$ from 25{,}000 labelled pairs, applies a 10\% shrink in both output projections to the quarter of directions that most favor hard negatives, and re-mines hard negatives to apply a second, smaller shrink.
The recipe takes seconds for each model and requires no agent.

\astra improves COCO test retrieval on all 13 checkpoints, by $+12.2$ RSUM on average (the sum of recall at 1, 5, and 10 in both directions).
On COCO test, \ours improves all 21 evaluated CLIP checkpoints by $+16.8$ RSUM on average, including 8 unseen by the agent, 3 of them downloaded after the recipe was fixed.
It improves Flickr30k test retrieval by $+6.6$ while limiting any drop in ImageNet zero-shot accuracy to half a percentage point.
Across these checkpoints, the median share of text-to-image misses landing on hubs falls from 28\% to 19\%.
\ours also outperforms the agent's own patches on all 13 checkpoints with a session and improves retrieval on 4 of 5 dual encoders from other families.
\ours involves no gradient-based optimization or hyperparameter search, needs no test-time reference bank, and adds no inference cost.

\noindent Our main contributions are:
\begin{itemize}[noitemsep,leftmargin=*,topsep=2pt]
    \item We show which directions to change and how: removing the leading hard-negative directions, which align with the common mean and the gap, degrades retrieval, whereas mildly shrinking hard-negative directions improves it while largely preserving the gap's size (Sec.~\ref{sec:f2}).
    \item We introduce a paired statistic that identifies these directions and use it to guide an LLM agent that can edit any component of either tower. The agent's edits improve retrieval on all 13 CLIP checkpoints, evaluated on test sets that it never accesses, and converge on the output projections (Sec.~\ref{sec:f1}).
    \item We develop \oursi, an agent-free, closed-form method motivated by a first-order analysis of the hard-negative margin (Sec.~\ref{sec:edit}). It improves all 21 CLIP checkpoints we evaluate and 4 of 5 dual encoders from other families, and outperforms the agent's own patches (Sec.~\ref{sec:f3}).
\end{itemize}

\section{Related Work}
\label{sec:related}

\noindent\textbf{Modality gap and retrieval correction.}
CLIP's image and text embeddings occupy separate regions~\cite{modalitygap}, with recent work relating this gap to information imbalance~\cite{schrodi2025twoeffects}.
Methods reduce the gap~\cite{an2025i0t}, although this does not consistently improve zero-shot classification~\cite{sato2026hubness} or retrieval~\cite{grassucci2026closing}.
We instead identify which directions to change: retrieval improves by mildly attenuating directions that favor hard negatives while largely preserving the gap.
Centering~\cite{suzuki2013centering} and all-but-the-top~\cite{mu2018allbutthetop} remove mean and top-variance directions chosen without pair labels, which lowers CLIP retrieval; shrinking them mildly gains little (see the supplement).
Hubness is a related problem in high-dimensional nearest-neighbor search~\cite{radovanovic2010hubs,dinu2015hubness}.
Inverted softmax~\cite{invsoftmax}, CSLS~\cite{csls}, QB-Norm~\cite{qbnorm}, DBNorm~\cite{dbnorm}, NNN~\cite{nnn}, and DN~\cite{dn} correct rankings using queries or reference samples at test time.
Our correction is written into the checkpoint, adds no inference cost, and complements re-ranking.
Fine-tuning~\cite{lora} also modifies weights but requires gradient-based optimization and hyperparameter search.

\noindent\textbf{Weight-space programming.}
Weight-space programming~\cite{wsp_blog} introduces a new way to modify neural networks: an LLM agent inspects their weights and edits them directly without gradients.
WeightLoop~\cite{weightloop} develops this direction through experimental feedback and reuse of editing programs, alongside broader efforts in LLM-driven discovery~\cite{funsearch,aiscientist}.
We study cross-modal retrieval, where success depends on how two towers represent matched pairs relative to hard negatives.
Our paired statistic provides evidence about this relationship, while the agent chooses where and how to edit.
We use these sessions to discover a recurring correction and turn it into a recipe that transfers to unseen checkpoints without the agent.
Recent questions about the role of weight access and the consistency of agent-written edits~\cite{liu2026computation,ahmed2026llm,hu2026exploring} also motivate our controlled evaluation.

\noindent\textbf{Interpreting and editing CLIP.}
Mechanistic interpretability connects model behavior to internal components~\cite{elhage2021framework,conmy2023acdc}.
For CLIP, this includes attention heads~\cite{textspan}, neurons~\cite{gandelsman2025neurons}, and artifact tokens~\cite{registers,jiang2025registers}; LLM-based tools further automate component analysis~\cite{bills2023neurons,maia}.
The closest editing methods include SITH~\cite{sith}, which rescales singular directions in vision heads, and IsoCLIP~\cite{isoclip}, which analyzes both output projections for intra-modal retrieval.
Our agent can investigate components in either tower, guided by paired evidence that identifies where their agreement favors incorrect matches.
This connects component analysis to explicit weight edits for cross-modal retrieval.
More broadly, model editing changes facts, rules, or task behavior~\cite{rome,memit,santurkar2021editing,taskarith}; our objective concerns the two encoders' joint ranking.

\noindent\textbf{Hard negatives and closed-form metrics.}
Hard negatives~\cite{vsepp,alomari2025maxmatch} and neighborhood losses~\cite{thomas2020preserving} guide retrieval training, while KISSME~\cite{kissme} computes a metric from similar and dissimilar pairs.
Our paired statistic $\dD$ identifies directions where the towers agree more on hard negatives than on matched pairs, providing explicit targets for weight edits.
The agent's recurring correction motivates \ours, which mildly shrinks these directions in both output projections.
We establish the optimality of its first shrink for a bilinear margin surrogate under per-direction and total contraction constraints, with the conditions specified in Sec.~\ref{sec:edit}.
This connects an agent-discovered edit to a closed-form recipe that requires no agent or model-specific tuning.

\section{Method}
\label{sec:method}
We study which weight edits help matched pairs outrank hard negatives in a pretrained dual encoder.
We first define a paired statistic that identifies the directions favoring hard negatives.
Guided by this statistic, \astra inspects the model and writes short, closed-form weight edits without backpropagation.
We then turn its recurring correction to the output projections into \ours, a simple recipe that can be applied to new checkpoints without the agent (Fig.~\ref{fig:method}).

\subsection{Problem Setup and the Paired Statistic}
\label{sec:setup}
Given an image or caption, retrieval aims to rank its matched counterpart above the other candidates.
CLIP computes the two representations separately: $e_v = W_v h_v(x)$ for an image $x$ and $e_t = W_t h_t(y)$ for a caption $y$.
Here, $h_v$ and $h_t$ are the image and text transformers, and $W_v$ and $W_t$ project their outputs into a shared $d$-dimensional space.
Candidates are ranked by cosine similarity $\hat e_v^{\top}\hat e_t$, where hats denote $L_2$ normalization.
We seek weight edits that improve this ranking on held-out pairs while keeping ImageNet zero-shot accuracy within a fixed tolerance.
A set of these edits forms a \emph{patch}; applying it produces an ordinary checkpoint with the same architecture and inference cost.

\noindent\textbf{The paired statistic.}
To guide an edit, we need to identify which parts of the shared representation favor wrong matches.
We therefore compare matched pairs with their \emph{hard negatives}: the incorrect candidates with the highest similarity scores.
These candidates compete most directly with the correct match and reveal where the towers' agreement is misleading.
For each caption $y$, we collect its five highest-scoring wrong images in $\mathcal{H}(y)$; for each image $x$, we collect its 25 highest-scoring wrong captions in $\mathcal{H}(x)$.
For matched pairs $\mathcal{P}$, we define the $d\times d$ paired statistic
\begin{equation}
\label{eq:D}
\begin{aligned}
\dD &= \mathbb{E}_{\mathcal{P}}\big[\hat e_v(x)\,\hat e_t(y)^{\top}\big]\\
 &- \tfrac12\, \mathbb{E}_{y,\,x' \in \mathcal{H}(y)}\big[\hat e_v(x')\, \hat e_t(y)^{\top}\big]\\
 &- \tfrac12\, \mathbb{E}_{x,\,y' \in \mathcal{H}(x)}\big[\hat e_v(x)\, \hat e_t(y')^{\top}\big].
\end{aligned}
\end{equation}
The first term measures agreement on matched pairs; the other two measure agreement on hard negatives, weighted equally across the two retrieval directions.
Along a unit direction $u$, agreement is the product $(u^{\top}\hat e_v)(u^{\top}\hat e_t)$ of the two projected embeddings.
The trace of $\dD$ is the mean similarity margin between matched pairs and hard negatives.
The eigenvalues of $\mathrm{sym}(\dD) = (\dD + \dD^{\top})/2$ decompose this margin into contributions along orthogonal directions.
A negative eigenvalue marks a direction where hard negatives agree more than matched pairs, making it a target for weight edits.

\subsection{Weight-Space Programming with \astra}
\label{sec:harness}
To turn paired evidence into better retrieval, we need to choose where to edit the towers and how to change their weights.
We use \astra (GPT-6-Astra) to investigate these choices by inspecting the model, writing weight edits, and testing their effect on retrieval.
In each \emph{session}, the agent operates on one checkpoint using only our harness tools; its native tools are disabled.
The prompt describes all model components and asks the agent to identify agreement and disagreement between the towers and edit their weights accordingly.
It lists the modality gap and hubness among hypotheses to investigate, while leaving the edit locations to the agent.

\noindent\textbf{Inspecting weights and paired evidence.}
The agent has read-only access to both towers' weights and 35{,}000 COCO training images~\cite{coco,karpathy} with five captions each.
Within fixed budgets, it can compute over the weights without gradients, inspect activations and per-head contributions, and examine $\dD$ for pairs of tower components.
We introduce $\dD$ to give the agent evidence that weights alone do not provide: which directions favor matched pairs and which favor hard negatives.
The agent determines where and how to edit the weights using this evidence.
To assess candidate edits, it receives aggregate scores on 5{,}000 further training images, COCO val, and a 5{,}000-image ImageNet proxy.
Tool definitions and budgets are given in the supplement.

\noindent\textbf{Writing explicit weight edits.}
Each edit is a program that can be inspected and re-run without the agent.
These programs may scale attention heads, remove directions from heads or MLPs, add low-rank updates, apply a shared matrix to both projections, or rewrite up to four whole tensors across the two towers.
Edits must be computed in closed form from weights and training-pair statistics, using operations such as means, covariances and eigendecompositions.
Backpropagation, iterative fitting and regression to data targets are disallowed.
We statically check every code block and manually trace each final patch array to its source code; all 13 final patches pass these checks.

\noindent\textbf{Selecting the final patch.}
A candidate replaces the current patch only if its COCO val score is higher by more than two paired standard errors and its ImageNet proxy score drops by at most 0.5 points relative to the unedited model.
COCO test and Flickr30k test~\cite{flickr30k} are sealed: no tool can read them.
After the session ends, we evaluate the final patch once on these test sets to measure how well the edits generalize.

\subsection{\ours: A Transferable Weight-Space Recipe}
\label{sec:edit}
Nine of the 13 sessions produce projection edits that attenuate directions with negative eigenvalues of $\mathrm{sym}(\dD)$ (Sec.~\ref{sec:f1}).
To apply this correction to new checkpoints without another agent session, we distill the ViT-B/32 session's uniform shrink into \ours.
In that session, the agent sums the supplied componentwise blocks to form $\dD$, then shrinks the directions with the 128 smallest eigenvalues of $\mathrm{sym}(\dD)$ by 10\% in both projections.
These directions span one quarter of the 512-dimensional space.
\ours refines this correction in three steps (Fig.~\ref{fig:method}, bottom).

\noindent\textbf{Step 1: Identify hard-negative directions.}
The directions that favor wrong matches depend on the checkpoint, so we identify them from its own embeddings.
We embed 5{,}000 training images with five captions each (25{,}000 labelled pairs), mine hard negatives and compute $\dD$ (Eq.~\eqref{eq:D}).

\noindent\textbf{Step 2: Attenuate hard-negative directions.}
To reduce misleading agreement while retaining useful content in these directions, we apply a mild shrink to both towers.
Let $U$ contain orthonormal eigenvectors for the $d/4$ smallest eigenvalues of $\mathrm{sym}(\dD)$, all negative on the 21 tested checkpoints.
We apply a shared map of strength $s$:
\begin{equation}
\label{eq:S}
W_v \leftarrow S\,W_v, \quad W_t \leftarrow S\,W_t, \qquad S = I - s\,UU^{\top}.
\end{equation}
The map scales these directions by $1-s$ and leaves the remaining directions unchanged.
For unit input embeddings, the edited dot product is $\hat e_v^{\top} M \hat e_t$, where $M = S^{\top}S = I - (2s - s^2)\,UU^{\top}$ assigns weight $0.81$ to the selected directions at $s=0.10$.
Retrieval uses the cosine of the re-normalized edited embeddings.

\noindent\textbf{Step 3: Re-mine and refine.}
The first shrink changes the rankings, so a second correction should use the hard negatives that now compete with matched pairs.
We apply the first-round map $S_1$, re-normalize the embeddings, and re-mine $\mathcal{H}$ to compute an updated statistic $\dD'$.
Let $U'$ contain orthonormal eigenvectors for the $d/4$ smallest eigenvalues of $\mathrm{sym}(\dD')$.
We add a smaller second shrink:
\begin{equation}
\label{eq:two}
S = S_2 S_1,\;\; S_1 = I - 0.10\,UU^{\top},\;\; S_2 = I - 0.05\,U'U'^{\top}.
\end{equation}
The first round ($S_1$) reproduces the agent's edit; we add the second round using the re-mining operation already available in the harness.
The subspace size, shrink strengths and numbers of hard negatives are fixed across checkpoints, so applying \ours requires no agent or model-specific tuning.
In other words, the recipe carries over between checkpoints, while $S$ is recomputed from each checkpoint's own embeddings.

\noindent\textbf{First-order optimality analysis.}
The uniform shrink has a direct justification: it maximizes the bilinear hard-negative margin under bounds on both the change in each direction and the total contraction.
With fixed embeddings and mined negatives, this margin equals $\langle M, \mathrm{sym}(\dD)\rangle$, where $\langle\cdot,\cdot\rangle$ is the Frobenius inner product.
For a budget of $k$ directions, we constrain the metric to
\[
(1-c)I \preceq M \preceq I, \qquad \mathrm{tr}(I-M) \le ck,
\]
where $0<c\le1$ bounds contraction in each direction, $ck$ bounds total contraction, and $\preceq$ denotes positive-semidefinite order.
When at least $k$ eigenvalues are negative, Ky Fan's theorem~\cite{fan1949weyl} gives a maximizer $M = I - c\,U_kU_k^{\top}$, where $U_k$ contains orthonormal eigenvectors for the $k$ smallest eigenvalues.
Setting $k=d/4$ and $c = 2s - s^2 = 0.19$ recovers the first-round edit.
This guarantee concerns the bilinear surrogate.
Under cosine retrieval, uniform shrinkage also outperforms eigenvalue-weighted shrinkage at the same first-order gain by $8.0$ COCO val RSUM on average across 13 checkpoints (Sec.~\ref{sec:abl}).
For fixed embeddings with negatives re-mined under the bilinear score, the margin is concave in $M$ and the current $\mathrm{sym}(\dD)$ is a supergradient.
This provides first-order information about the margin, motivating recomputation of $\dD$ after the first edit to guide the second (see the supplement).

\section{Experiments}
\label{sec:exp}
We evaluate what should change in a pretrained CLIP model to improve retrieval.
We first measure the gains of \astra's patches and where they edit, then test whether hard-negative directions should be removed or mildly attenuated.
Finally, we test whether \ours transfers the resulting correction to new checkpoints without the agent.

\subsection{Datasets, Checkpoints and Evaluation}
\label{sec:setup_exp}
\noindent\textbf{Checkpoints.}
We run one \astra session on each of 13 CLIP checkpoints available through open\_clip~\cite{openclip}, covering ViT-B/32, B/16 and L/14~\cite{vit} and pretraining on OpenAI's data~\cite{clip}, LAION~\cite{laion400m,laion5b}, DataComp~\cite{datacomp}, DFN~\cite{dfn} and MetaCLIP~\cite{metaclip}.
We then apply \ours to these models and 8 further checkpoints, extending coverage to ViT-H/14 (Table~\ref{tab:transfer}), and to 5 dual encoders from other families.

\noindent\textbf{Datasets and metrics.}
We report RSUM, the sum of recall at 1, 5 and 10 in both retrieval directions, as the change from the zero-shot checkpoint.
COCO Karpathy test~\cite{coco,karpathy} measures retrieval improvement, while Flickr30k test~\cite{flickr30k} measures transfer beyond the COCO training pairs used to construct the edits.
We require ImageNet~\cite{imagenet} zero-shot accuracy to drop by at most 0.5 percentage points, referred to as the accuracy guard, and evaluate compositional understanding with SugarCrepe~\cite{sugarcrepe}.

\noindent\textbf{Settings.}
COCO test and Flickr30k test are inaccessible to the agent; each final patch is evaluated after its session ends.
For \ours, the rule is fixed across checkpoints and its matrices are frozen before COCO test evaluation.
We use validation splits for the analyses below and identify them explicitly.
A gain is significant when its 95\% confidence interval, computed by a paired bootstrap grouped by image, lies above zero.
Details are in Sec.~\ref{sec:harness} and the supplement.

\begin{figure}[t]
  \centering
  \includegraphics[width=0.92\linewidth]{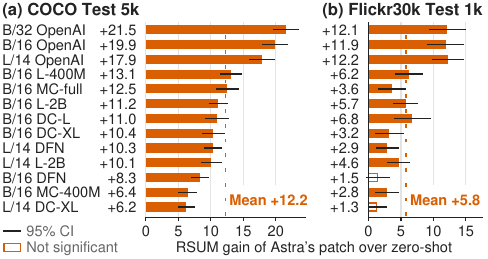}
  \caption{\textbf{Agent-written weight edits improve CLIP retrieval.} Test RSUM gains over zero-shot, with one \astra session for each model. Whiskers: 95\% confidence intervals; dashed lines: means. All COCO gains are significant; Flickr30k evaluates transfer beyond the session's feedback.}
  \label{fig:zoo}
\end{figure}

\subsection{Weight-Space Edits Improve Retrieval}
\label{sec:f1}
\astra's programs improve COCO test retrieval on all 13 checkpoints, each significantly, by $+12.2$ RSUM on average (Fig.~\ref{fig:zoo}).
The gains extend to Flickr30k test, averaging $+5.8$ RSUM and reaching significance on 11 of 13 checkpoints, although no tool reports this dataset to the agent.
Every patch passes the accuracy guard, and each COCO test gain is within $1.5$ RSUM of its validation gain (see the supplement).
These results show that explicit, gradient-free weight edits improve pretrained CLIP models across architectures and pretraining sources, with gains beyond the feedback used to select them.

\begin{figure}[t]
  \centering
  \begin{minipage}[c]{0.6\linewidth}\centering
  \includegraphics[width=\linewidth, trim=0 0 348 10, clip]{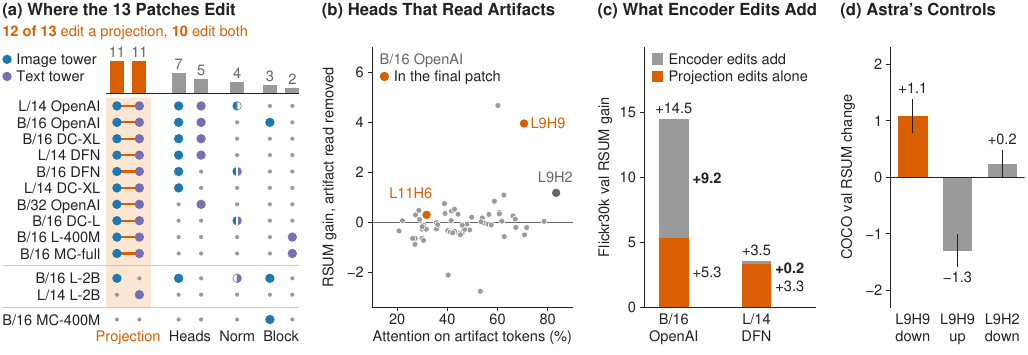}
  \end{minipage}\hfill
  \begin{minipage}[c]{0.37\linewidth}
  \caption{\textbf{Edit locations.} Rows: final patches; bars: patches per site. Blue: image tower; purple: text tower. Norm: final LayerNorm bias; Block: last block.}
  \label{fig:interp}
  \end{minipage}
\end{figure}

Paired evidence helps guide this process.
In controls on OpenAI B/16 and DFN-2B L/14 using a later harness, removing $\dD$ reduces Flickr30k validation gains by $6.3$ and $1.7$ RSUM, respectively.
Removing raw-weight reading has a smaller effect, $0.5$ and $1.3$ RSUM.
Under matched evaluation budgets, a random search over a fixed edit grammar with the same statistic also finds projection corrections, with a similar mean Flickr30k test gain ($+5.7$ against $+5.8$; see the supplement).
The paired statistic thus guides weight-space programming by making these corrections accessible, and \astra uses it to construct explicit weight-editing programs.

\begin{figure}[t]
  \centering
  \includegraphics[width=\linewidth]{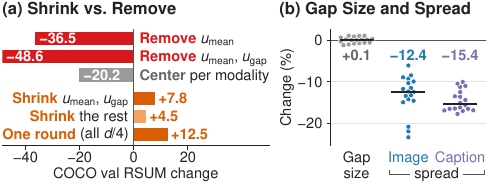}
  \caption{\textbf{Mild attenuation improves retrieval.} (a)~Removing the two leading hard-negative directions, which align with the common mean and the gap, hurts retrieval, as does centering each modality; shrinking them by 10\% helps (24 CLIPs). (b)~\ours largely preserves gap size while reducing the spread of similarities to the other modality's mean. Dots: 18 CLIPs; lines: medians.}
  \label{fig:anatomy}
\end{figure}

Although every session can edit internal components, 12 of the 13 final patches rewrite an output projection (Fig.~\ref{fig:interp}).
All 12 construct a map from $\dD$: 9 attenuate directions with negative eigenvalues as in Eq.~\eqref{eq:S}, and 3 apply rotations.
The remaining patch edits the last vision MLP alone.
Internal edits can also provide additional gains, such as $+9.2$ Flickr30k validation RSUM beyond the projection edits on OpenAI B/16.
When reapplied as a fixed procedure, the OpenAI B/16 MLP rewrite significantly improves on \ours on only one of five additional checkpoints (see the supplement).
The recurring projection correction provides the basis for \ours.

\subsection{Mild Attenuation of Hard-Negative Directions}
\label{sec:f2}
To understand this correction, we test whether its leading directions should be attenuated or removed.
Removing the modality gap~\cite{modalitygap} hurts retrieval in all 13 sessions, consistent with prior observations for classification~\cite{sato2026hubness}.
We extend the comparison to 24 checkpoints using $u_{\mathrm{mean}}$ and $u_{\mathrm{gap}}$, the eigenvectors associated with the two smallest eigenvalues of $\mathrm{sym}(\dD)$, which align with the common-mean and gap directions.
Removing both hurts every checkpoint, by $48.6$ COCO validation RSUM on average, as does per-modality centering~\cite{suzuki2013centering} ($-20.2$); shrinking them by 10\% instead improves all 24, by $+7.8$ (Fig.~\ref{fig:anatomy}a).
Among tested strengths, the best shrink is 10--20\% on every checkpoint, and the mean gain is negative at 30\% (see the supplement).
These two directions alone recover about 60\% of the $+12.5$ gain from one full round.
The comparison supports retaining their useful content while reducing their influence on retrieval scores.
The modality means alone do not explain the gain: shrinking their span gains $+2.3$ COCO validation RSUM on 21 checkpoints, against $+16.1$ for \ours (see the supplement).

\begin{figure}[t]
  \centering
  \includegraphics[width=0.92\linewidth]{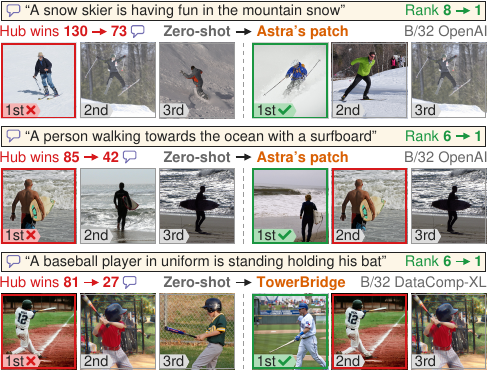}
  \caption{\textbf{Correct images replace hubs.} Top-3 COCO validation retrievals before and after \astra's patch (top) and \ours on a checkpoint without a session (bottom). Green: correct image; red: hub. Hub wins count captions (bubble icon) retrieving that image first. Further examples are in the supplement.}
  \label{fig:qualmain}
\end{figure}

We then examine how the resulting \ours correction changes rankings.
On held-out COCO pairs, the margin to the hardest negative widens in both directions on all 21 checkpoints, and hubness decreases (see the supplement).
The median share of text-to-image misses landing on hubs falls from $28\%$ to $19\%$.
Meanwhile, the modality gap remains largely unchanged: its median size changes by $+0.1\%$ over the 18 checkpoints in Fig.~\ref{fig:anatomy}b.
Figure~\ref{fig:qualmain} illustrates the ranking improvement with both an agent patch and \ours: a correct image replaces a hub that had attracted many unrelated captions.
This is the \emph{bridging} effect: matched pairs become easier to distinguish from hard negatives while the modality gap remains largely unchanged.

\begin{figure*}[t]
  \centering
  \includegraphics[width=0.84\linewidth]{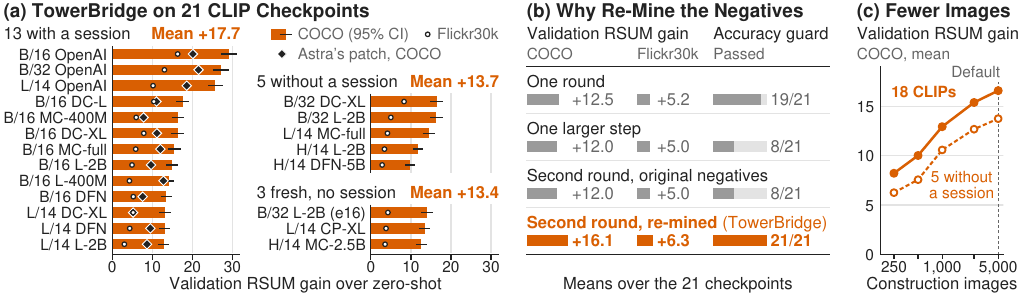}
  \caption{\textbf{\ours transfers the correction without the agent.}
  (a)~COCO and Flickr30k validation gains on 21 CLIPs; fresh checkpoints were downloaded after the recipe was fixed.
  (b)~Re-mining improves on one round while meeting the accuracy guard on all 21.
  (c)~COCO validation gains with fewer construction images.}
  \label{fig:transfer}
\end{figure*}

\begin{table}[t]
\centering
\caption{\textbf{\ours transfers to checkpoints without agent sessions.} RSUM change from zero-shot. 1R, 2R: one and two rounds; other columns use 2R. IN: ImageNet top-1; SC: SugarCrepe add. Fresh: the last three checkpoints, downloaded after the recipe was fixed.}
\label{tab:transfer}
\tabdense
\scalebox{0.9}{%
\begin{tabular*}{\linewidth}{@{\extracolsep{\fill}}l*{8}{r}@{}}
\toprule
 & \multicolumn{2}{c}{COCO val} & \hd{COCO} & \multicolumn{2}{c}{Flickr30k} & & \multicolumn{2}{c@{}}{SC add} \\
\cmidrule(lr){2-3}\cmidrule(lr){4-4}\cmidrule(lr){5-6}\cmidrule(l){8-9}
Checkpoint & \hd{1R} & \hd{2R} & \hd{Test} & \hd{Val} & \hd{Test} & \hd{IN} & \hd{1R} & \multicolumn{1}{c@{}}{2R} \\
\midrule
L/14 MetaCLIP-fullCC & $+11.2$ & $+14.4$ & $+15.8$ & $+4.1$ & $+5.1$ & ${-}0.04$ & ${-}8.1$ & ${-}1.1$ \\
H/14 LAION-2B & $+9.6$ & $+11.7$ & $+13.1$ & $+3.4$ & $+4.8$ & ${-}0.23$ & ${-}4.5$ & ${-}2.1$ \\
B/32 LAION-2B & $+12.9$ & $+16.3$ & $+15.1$ & $+5.0$ & $+6.2$ & ${-}0.01$ & ${-}6.5$ & ${-}2.8$ \\
B/32 DataComp-XL & $+12.4$ & $+16.4$ & $+17.3$ & $+8.3$ & $+7.4$ & ${-}0.23$ & ${-}5.9$ & ${-}1.2$ \\
H/14 DFN-5B & $+6.6$ & $+9.8$ & $+11.1$ & $+2.8$ & $+3.5$ & ${-}0.05$ & ${-}8.5$ & ${-}3.3$ \\
\midrule
B/32 LAION-2B (e16) & $+10.7$ & $+14.1$ & $+15.7$ & $+4.3$ & $+5.2$ & ${-}0.05$ & ${-}5.7$ & ${-}2.0$ \\
L/14 CommonPool-XL & $+10.6$ & $+13.4$ & $+13.9$ & $+3.8$ & $+5.1$ & ${-}0.23$ & ${-}7.4$ & ${-}2.5$ \\
H/14 MetaCLIP-2.5B & $+8.9$ & $+12.6$ & $+14.2$ & $+3.6$ & $+4.6$ & $+0.08$ & ${-}7.8$ & ${-}2.1$ \\
\midrule
Mean, 8 unseen & $+10.4$ & $+13.6$ & $+14.5$ & $+4.4$ & $+5.2$ & ${-}0.10$ & ${-}6.8$ & ${-}2.1$ \\
Mean, 3 fresh & $+10.1$ & $+13.4$ & $+14.6$ & $+3.9$ & $+5.0$ & ${-}0.07$ & ${-}7.0$ & ${-}2.2$ \\
Mean, 13 with a session & $+13.7$ & $+17.7$ & $+18.2$ & $+7.5$ & $+7.5$ & ${-}0.20$ & ${-}4.3$ & ${-}1.8$ \\
Mean, all 21 & $+12.5$ & $+16.1$ & $+16.8$ & $+6.3$ & $+6.6$ & ${-}0.16$ & ${-}5.3$ & ${-}1.9$ \\
\bottomrule
\end{tabular*}%
}
\end{table}

\subsection{\ours Transfers Without the Agent}
\label{sec:f3}
\ours applies the same rule to each checkpoint, computing its correction from that model's own embeddings without model-specific tuning (Sec.~\ref{sec:edit}).
It improves COCO test retrieval on all 21 CLIP checkpoints, each significantly, by $+16.8$ RSUM on average (Table~\ref{tab:transfer}).
All six recalls increase on every checkpoint, with mean recall@1 gains of $+3.8$ points for image-to-text and $+2.8$ for text-to-image retrieval.
The mean Flickr30k test gain is $+6.6$ RSUM, and every checkpoint passes the accuracy guard.

The 8 checkpoints without agent sessions provide a direct test of transfer.
Their gains average $+14.5$ COCO test and $+5.2$ Flickr30k test RSUM, extending to ViT-H/14 and pretraining sources absent from the sessions.
Of these checkpoints, 3 were downloaded only after the recipe was fixed; their mean COCO test gain is $+14.6$.
These results show that applying the discovered correction does not require repeating the agent's search on each model.
The recipe also works with other paired data: built from Flickr30k training pairs, it improves COCO validation retrieval by $8.0$--$22.2$ RSUM on three checkpoints (see the supplement).

The same correction significantly improves Flickr30k validation retrieval while passing the accuracy guard on 4 of 5 evaluated dual encoders from other families: EVA02-B/16~\cite{evaclip}, MobileCLIP-B~\cite{mobileclip}, CoCa B/32~\cite{coca} and SigLIP B/16~\cite{siglip}.
For SigLIP, which lacks a vision projection, we apply $S$ as a linear layer after the vision encoder.
SigLIP2~\cite{siglip2} also improves retrieval but loses $0.70$ ImageNet points, so the accuracy guard retains the zero-shot model (see the supplement).

Distilling the agent's uniform shrink and adding our second-round refinement also improves on the original patches.
On all 13 checkpoints with sessions, \ours significantly exceeds the agent's COCO test gain, averaging $+18.2$ versus $+12.2$ RSUM.
Constructing the edit takes about 8\,s for B/16 and 24\,s for L/14 on one B200, including encoding and both correction rounds but excluding model loading, preprocessing and evaluation of the accuracy guard (see the supplement).
The original agent sessions took 3.6--12.1 hours each.
Once folded into CLIP's projections, the correction yields an ordinary checkpoint with unchanged architecture and inference cost.

\subsection{Ablation Study}
\label{sec:abl}
\begin{table}[t]
\centering
\caption{\textbf{Design choices in \ours.} Means over 18 checkpoints (exploratory). Default: two rounds, $d/4$ directions, both towers, 5{,}000 images. Guard: checkpoints losing ${\le}\,0.5$ ImageNet points. $^{\dagger}$One round on 16 checkpoints, with strength selected on held-out data. $^{\ddagger}$B/16 OpenAI, on our expert patch.}
\label{tab:ablation}
\tabdense
\scalebox{0.9}{%
\begin{tabular*}{\linewidth}{@{\extracolsep{\fill}}l*{5}{r}@{}}
\toprule
Variant & COCO val & Flickr30k val & IN & SC add & Guard \\
\midrule
\ours edit (default) & $\mathbf{+16.6}$ & $\mathbf{+6.7}$ & $\mathbf{{-}0.17}$ & $\mathbf{{-}1.9}$ & $\mathbf{18/18}$ \\
\midrule
One round & $+12.9$ & $+5.6$ & ${-}0.18$ & ${-}5.0$ & $16/18$ \\
Three rounds & $+18.6$ & $+7.4$ & ${-}0.32$ & ${-}2.3$ & $13/18$ \\
\midrule
Subspace $d/16$ & $+15.7$ & $+6.7$ & ${-}0.02$ & ${-}1.7$ & $18/18$ \\
Subspace $d/2$ & $+16.6$ & $+6.7$ & ${-}0.25$ & ${-}2.0$ & $16/18$ \\
\midrule
Image tower only & $+11.9$ & $+5.3$ & ${-}0.05$ & ${-}2.6$ & $18/18$ \\
Text tower only & $+11.7$ & $+5.2$ & $+0.03$ & ${-}2.1$ & $18/18$ \\
Text-only variant ($M$ on $W_t$) & $+15.8$ & $+6.3$ & ${-}0.07$ & ${-}1.0$ & $18/18$ \\
\midrule
Random subspace & ${-}1.0$ & ${-}0.7$ & ${-}0.12$ & ${-}0.0$ & $18/18$ \\
Most \emph{positive} directions & ${-}66.0$ & ${-}34.4$ & ${-}1.43$ & $+3.2$ & $0/18$ \\
\midrule
1{,}000 images & $+12.9$ & $+5.3$ & ${-}0.46$ & ${-}3.8$ & $12/18$ \\
\midrule
Random negatives$^{\dagger}$ & $+3.4$ & $+1.8$ & -- & $+0.7$ & -- \\
PLS / CCA / Procrustes$^{\ddagger}$ & ${-}0.2$ to $+0.8$ & $+0.0$ to $+1.1$ & -- & -- & -- \\
\bottomrule
\end{tabular*}%
}
\end{table}

We test which choices make the correction effective using Table~\ref{tab:ablation} and Fig.~\ref{fig:transfer}b,c.
Table~\ref{tab:ablation} reports means over 18 checkpoints unless noted otherwise.

\noindent\textbf{Hard-negative directions drive the gain.}
Random subspaces of the same size give $-1.0$ COCO validation RSUM, while shrinking the most positive directions gives $-66.0$, compared with $+16.6$ for \ours.
A separate one-round comparison on 16 checkpoints finds that replacing hard negatives with random negatives reduces the mean gain by $7.7$ RSUM.
A 10\% shrink of as many top principal directions, chosen without pair labels, gains at most $+0.3$ on 24 checkpoints, against $+12.5$ for one round of \ours (see the supplement).
The paired statistic identifies useful editing targets beyond generic perturbations of the representation.

\noindent\textbf{A uniform shrink outperforms eigenvalue weighting.}
We test the choice justified for the bilinear margin in Sec.~\ref{sec:edit} under the cosine score used for retrieval.
At matched first-order gain, uniform shrinkage outperforms eigenvalue-weighted shrinkage on all 13 tested checkpoints, by $8.0$ COCO validation RSUM on average, with $0.8$ points less ImageNet loss.
This supports the mild, uniform correction under the deployed scoring rule as well as the constrained surrogate.

\noindent\textbf{Re-mining makes the second round effective.}
The first edit changes which negatives compete with matched pairs, so we compare a larger step, a second round using the original negatives, and a second round with re-mining (Fig.~\ref{fig:transfer}b).
Re-mining replaces about a quarter of the negatives and gains $4.13$ COCO validation RSUM over a second round using the original negatives, on average across 21 checkpoints.
It passes the accuracy guard on all 21; the larger step and stale-negative variants each fail on 13.
A third round raises retrieval further but fails the accuracy guard on 5 of the 18 ablation checkpoints, motivating the two-round recipe.

\noindent\textbf{Both towers contribute; 250 images already help.}
Editing both towers gives $+16.6$ COCO validation RSUM, compared with $+11.9$ for image-only and $+11.7$ for text-only editing, with each variant re-mining its own negatives.
For an existing image index, folding the full metric $M$ into the text tower retains $+15.8$ while leaving image embeddings unchanged.
The correction also works with fewer construction images: 1{,}000 images give $+12.9$ RSUM and 250 give $+8.2$ (Fig.~\ref{fig:transfer}c).
We use 5{,}000 images: all 18 checkpoints pass the accuracy guard, compared with 12 at 1{,}000 images.

\subsection{Comparison and Practical Use}
\label{sec:stands}
\begin{table}[t]
\centering
\caption{\textbf{Comparison with re-rankers and KISSME.} RSUM change vs.\ zero-shot, 8 checkpoints (exploratory). Data: \emph{Pairs} = 25{,}000 labelled COCO pairs, used once; COCO / Flickr30k = a test-time bank from that dataset.}
\label{tab:alt}
\tabfont
\scalebox{0.9}{%
\begin{tabular*}{\linewidth}{@{\extracolsep{\fill}}llrr@{}}
\toprule
Method & Data & Flickr30k val & COCO val \\
\midrule
\ours edit & Pairs & $\mathbf{+7.6}$ & $\mathbf{+18.8}$ \\
Text-only variant & Pairs & $+7.4$ & $+18.2$ \\
KISSME-K~\cite{kissme} & Pairs & $+2.5$ & $+4.3$ \\
\midrule
CSLS~\cite{csls} & COCO & $+6.5$ & $+19.9$ \\
NNN~\cite{nnn} & COCO & $+5.4$ & $+22.0$ \\
QB-Norm~\cite{qbnorm} & COCO & $+1.6$ & $+16.7$ \\
DN~\cite{dn} & COCO & $+2.8$ & $+6.1$ \\
\midrule
CSLS & Flickr30k & $+9.7$ & $+8.0$ \\
NNN & Flickr30k & $+11.2$ & $+3.4$ \\
QB-Norm & Flickr30k & $+8.8$ & $-1.5$ \\
DN & Flickr30k & $+3.8$ & $+5.1$ \\
\midrule
Edit + CSLS & COCO & $+9.6$ & $+28.2$ \\
Edit + NNN & Flickr30k & $+13.5$ & $+14.4$ \\
\bottomrule
\end{tabular*}%
}
\end{table}

\noindent\textbf{\ours outperforms KISSME-K.}
Table~\ref{tab:alt} compares \ours with KISSME-K~\cite{kissme}, using the same labelled pairs and hard negatives on 8 checkpoints.
\ours gains $+7.6$ Flickr30k validation RSUM against $+2.5$ for KISSME-K.

\noindent\textbf{\ours needs no gradient-based optimization.}
On OpenAI B/16, a shared map trained with gradients on the same 5{,}000 images gains $+38.7$ COCO validation RSUM, compared with $+29.1$ for \ours; on DataComp-XL B/16 and DFN-2B L/14, the trained map is level or at most $2.2$ ahead.
LoRA~\cite{lora} achieves a larger gain on OpenAI B/16 ($+46.9$ COCO test RSUM).
\ours provides retrieval improvements without model-specific optimization, a held-out selection split or hyperparameter search (see the supplement).

\noindent\textbf{\ours preserves overall compositional accuracy.}
Across 21 checkpoints, overall SugarCrepe accuracy is nearly unchanged ($+0.06$ points): replace and swap improve by $1.1$ and $1.6$ points, while add decreases by $1.9$.
A constrained variant of Step~2 addresses this tradeoff in a separate four-checkpoint evaluation, retaining nearly the same retrieval gain ($-0.2$ Flickr30k validation RSUM relative to \ours) and the zero-shot add accuracy ($+0.2$ points; see the supplement).

\noindent\textbf{\ours complements test-time re-ranking.}
On the 8 checkpoints in Table~\ref{tab:alt}, \ours outperforms all evaluated re-rankers on Flickr30k validation when their reference bank comes from COCO.
With an in-domain bank, some re-rankers achieve larger gains.
For both bank sources, combining \ours with each evaluated re-ranker adds $2.0$--$4.9$ Flickr30k validation RSUM on average across the 8 checkpoints.

\section{Conclusion}
\label{sec:conclusion}
We showed that the directions of CLIP that favor hard negatives should be mildly attenuated rather than removed.
Our paired statistic identifies these directions, and the leading two align with the common mean and the gap.
Removing them degrades retrieval, whereas a mild shrink improves it while largely preserving the gap.
\astra, an LLM agent that can edit any component of either tower, converges on the output projections and improves COCO test retrieval on all 13 evaluated CLIP checkpoints.
We distill and refine its recurring correction into \ours, which improves all 21 evaluated CLIP checkpoints and 4 of 5 dual encoders from other families without the agent, model-specific tuning or added inference cost.

\noindent\textbf{Limitations.}
The recipe requires labelled image--text pairs and currently targets dual encoders.
Trained maps and re-rankers with an in-domain bank can achieve larger retrieval gains.
We will release code, prompts and logs.

\clearpage
\section*{LLM Usage Statement}
\noindent\textbf{LLMs in the research.}
An LLM is part of the method we study: \astra (GPT-6-Astra) is the agent that inspects model weights and writes the weight edits evaluated in Sec.~\ref{sec:exp}, in the 13 sessions and in the control sessions.
Its prompt, tools and data access are described in Sec.~\ref{sec:harness} and the supplement.
\ours itself is a closed-form procedure and uses no LLM.

\noindent\textbf{Code and experiments.}
We used Claude Code (Anthropic; Claude Fable 5.2 and Claude Opus models) and OpenAI Codex as coding assistants.
They wrote and debugged much of the code for the agent harness, the baselines, the evaluation and the analyses, and launched and monitored experiments that the authors designed or approved.
LLM-generated reviews of our drafts were used to find weaknesses, and several control experiments in the supplement were suggested by these reviews or by the assistants.

\noindent\textbf{Figures.}
All figures were made with LLM assistance: the assistants wrote the plotting and TikZ code that draws each figure from our result files and iterated on its layout under our direction.
No figure contains AI-generated imagery; all photographs are from the evaluated datasets.

\noindent\textbf{Text.}
The same assistants helped draft, restructure and revise all sections of the paper and the supplement, including the abstract, captions and tables, and helped search for related work and check bibliographic entries.

\noindent\textbf{Responsibility.}
The authors designed the study, chose and approved the experiments, verified the code and outputs, checked the citations and reported values, and remain responsible for all scientific claims and submitted content.
No generative model was used to fabricate experimental data or citations.
{
    \small
    \bibliographystyle{ieeenat_fullname}
    \bibliography{main}
}
\clearpage
\raggedbottom
\renewcommand{\thepage}{S\arabic{page}}
\setcounter{page}{1}
\maketitlesupplementary
\setcounter{section}{0}
\renewcommand{\thesection}{\Alph{section}}
\setcounter{table}{0}\renewcommand{\thetable}{S\arabic{table}}
\setcounter{figure}{0}\renewcommand{\thefigure}{S\arabic{figure}}
\setcounter{equation}{0}\renewcommand{\theequation}{S\arabic{equation}}
\renewcommand{\theHsection}{supp.\Alph{section}}\renewcommand{\theHtable}{supp.\arabic{table}}\renewcommand{\theHfigure}{supp.\arabic{figure}}\renewcommand{\theHequation}{supp.\arabic{equation}}

\noindent A caption can retrieve the wrong image because the same few images score highly for many unrelated queries. Our paired statistic identifies directions that favor these hard negatives. Guided by that evidence, the agent wrote a first-round shrink into both projections on OpenAI B/32. We added a second shrink after re-mining the negatives. The resulting correction helps the matching image outrank its competitors while largely preserving the distance between the modality means (Sec.~\ref{supp:bridge}).

A map trained on the same pairs reaches similar retrieval scores on the 15 checkpoints excluded from its schedule selection (Sec.~\ref{supp:fixedmap}). Most of its gain lies in directions that \ours identifies before training (Sec.~\ref{supp:learnedmetric}). On radiology images, rebuilding \ours from in-domain pairs gives a larger test gain than a map tuned in that domain on all 21 checkpoints (Sec.~\ref{supp:ood}).

Unless specified, we average the fixed two-round edit equally over the 21 CLIPs and report changes from zero-shot. Validation ablations show the edit before fallback. For deployment, the ImageNet accuracy guard selects two rounds, one round or zero-shot (Sec.~\ref{supp:guard}). Accuracy changes are in percentage points. Scores and paired differences are rounded separately.

\section{Retrieval Gains and Learned Adaptation}
\label{supp:f2}

\subsection{Retrieval Across CLIP Checkpoints}
All six COCO test recalls improve on every checkpoint. At rank one, the gains average $3.8$ points for image-to-text and $2.8$ for text-to-image retrieval. Summed over ranks 1, 5 and 10 in both directions, recall gains average $16.8$ RSUM on COCO test and $6.6$ on Flickr30k test (Table~\ref{tab:transfer}). Table~\ref{tab:absolute_recall} shows absolute recalls at three starting scores, and Table~\ref{tab:recall_all} gives every checkpoint's changes.

\begin{table*}[t]
\centering
\caption{\textbf{Absolute retrieval on COCO test.} Recall (\%) in a gallery of 5{,}000 images and their captions, before and after the two-round edit. RSUM sums the six unrounded recalls. Table~\ref{tab:recall_all} gives the changes for all 21 checkpoints.}
\label{tab:absolute_recall}
\tabfont\setlength{\tabcolsep}{5pt}
\begin{tabular}{@{}llrrrrrrr@{}}
\toprule
& & \multicolumn{3}{c}{Image $\to$ text} & \multicolumn{3}{c}{Text $\to$ image} & \\
Checkpoint & Method & R@1 & R@5 & R@10 & R@1 & R@5 & R@10 & RSUM \\
\midrule
B/16 OpenAI & Zero-shot & $52.5$ & $76.7$ & $84.7$ & $33.1$ & $58.4$ & $69.0$ & $374.3$ \\
 & \ours & $56.5$ & $80.8$ & $87.9$ & $38.8$ & $64.8$ & $74.8$ & $403.6$ \\
L/14 DFN-2B & Zero-shot & $65.7$ & $85.9$ & $91.2$ & $48.6$ & $73.8$ & $82.4$ & $447.5$ \\
 & \ours & $69.4$ & $87.9$ & $92.8$ & $50.8$ & $76.0$ & $84.1$ & $460.9$ \\
B/32 LAION-2B (e16; fresh) & Zero-shot & $56.3$ & $79.5$ & $87.0$ & $39.1$ & $64.7$ & $74.8$ & $401.4$ \\
 & \ours & $59.8$ & $82.2$ & $88.9$ & $41.8$ & $67.3$ & $77.2$ & $417.2$ \\
\bottomrule
\end{tabular}
\end{table*}

The 13 checkpoints with agent sessions gain $18.2$ COCO test RSUM on average. The eight without sessions gain $14.5$, including $14.6$ on the three downloaded after the recipe was fixed. All three pass the guard and improve Flickr30k val retrieval significantly. Across these three, round two also improves Flickr30k val retrieval and SugarCrepe add accuracy over round one (Table~\ref{tab:ci}). Table~\ref{tab:cohorts} identifies each group.

\subsection{Comparison with a Fixed-Schedule Hinge Map}
\label{supp:fixedmap}
How much retrieval can a shared linear map gain by training on the same pairs? We select one schedule on six development checkpoints and apply it to all 21. The selected map uses a hard-negative hinge loss, 50 Adam steps and no penalty (Sec.~\ref{supp:map_protocol}). Both methods pass the ImageNet guard on all 21 (Table~\ref{tab:fixedmap}).

The hinge map gains $17.1$ COCO val and $7.1$ Flickr30k test RSUM, compared with $16.1$ and $6.6$ for \ours. Its COCO val advantage averages $2.6$\ci{2.0}{3.3} RSUM on the six development checkpoints. On the other 15, hinge minus \ours is $+0.4$\ci{-0.1}{+0.9} on COCO val, $0.0$\ci{-0.6}{+0.6} on Flickr30k val and $-0.2$\ci{-0.8}{+0.5} on Flickr30k test. ImageNet accuracy changes by $-0.12$ points for \ours and $-0.30$ for hinge on these 15. Section~\ref{supp:retention} tests retention on independent images.

We estimate comparable deployment times because both methods encode the inputs and check classification accuracy (Sec.~\ref{supp:cost}). Does training also change the same directions that \ours selects from the unedited model?

\begin{table}[t]
\centering
\caption{\textbf{Fixed-schedule maps and the analytic correction.} Changes from zero-shot in retrieval RSUM, ImageNet top-1 (IN) and SugarCrepe add accuracy (Add). Guard counts checkpoints within the 0.5-point ImageNet tolerance. We selected one schedule per loss on six models. The last two blocks show the 15 excluded from schedule selection and the eight without agent sessions.}
\label{tab:fixedmap}
\tabdense
\begin{tabular*}{\linewidth}{@{\extracolsep{\fill}}l rrr rrr@{}}
\toprule
 & COCO & \multicolumn{2}{c}{Flickr30k} & & & \\
\cmidrule(lr){3-4}
Method & val & val & test & IN & Guard & Add \\
\midrule
\multicolumn{7}{@{}l}{\textit{All 21 CLIP checkpoints}} \\
\ours & $+16.1$ & $+6.3$ & $+6.6$ & $-0.16$ & 21/21 & $-1.9$ \\
Fixed hinge & $+17.1$ & $+6.8$ & $+7.1$ & $-0.22$ & 21/21 & $-1.2$ \\
Fixed InfoNCE & $+11.2$ & $+5.6$ & $+5.7$ & $-0.04$ & 21/21 & $+0.8$ \\
\midrule
\multicolumn{7}{@{}l}{\textit{The 15 checkpoints excluded from schedule selection}} \\
\ours & $+14.1$ & $+5.0$ & $+5.4$ & $-0.12$ & 15/15 & $-2.1$ \\
Fixed hinge & $+14.5$ & $+5.0$ & $+5.2$ & $-0.30$ & 15/15 & $-2.0$ \\
Fixed InfoNCE & $+9.2$ & $+4.1$ & $+4.6$ & $-0.08$ & 15/15 & $+0.3$ \\
\midrule
\multicolumn{7}{@{}l}{\textit{The 8 checkpoints without agent sessions}} \\
\ours & $+13.6$ & $+4.4$ & $+5.2$ & $-0.10$ & 8/8 & $-2.1$ \\
Fixed hinge & $+14.1$ & $+3.8$ & $+4.7$ & $-0.28$ & 8/8 & $-2.3$ \\
Fixed InfoNCE & $+8.5$ & $+3.4$ & $+4.2$ & $-0.06$ & 8/8 & $+0.0$ \\
\bottomrule
\end{tabular*}
\end{table}

\subsection{What the Learned Map Changes}
\label{supp:learnedmetric}
A shared map $A$ determines cosine rankings through $A^{\top}A$, up to a global scale. We scale this metric to be closest to the identity in Frobenius norm, writing $\tilde M_A=\alpha A^{\top}A$, and compare it with the first-round subspace $U$ from the unedited checkpoint.

\noindent\textbf{The learned map contracts the directions selected by \ours.}
The hinge metric gives directions in $U$ a mean weight of $0.90$ and the remaining directions $1.015$. The corresponding weights for \ours are $0.79$ and $1.05$. Along each eigenvector $u_i$ of $\mathrm{sym}(\dD)$, the learned weight $u_i^{\top}\tilde M_Au_i$ tracks $\lambda_i$. Their Spearman correlation averages $0.97$ across checkpoints (minimum $0.84$) and $0.99$ within $U$. The learned metric's $d/4$ most contracted directions overlap $U$ by $0.64$ on average (range $0.55$--$0.67$, random reference $0.25$).

\noindent\textbf{Most of the learned gain lies in these directions.}
We keep the learned change inside a subspace with projector $P$ by applying the square root of $M_P=I+P(\tilde M_A-I)P$ to both towers. Restricting the map to $U$ retains $12.3$ of its $17.1$ COCO val RSUM gain. An equal-rank random subspace retains $1.8$, and the complement of $U$, three times as large, retains $6.0$ (Table~\ref{tab:learnedmetric}). On every checkpoint $U$ retains more gain than its complement or any of the five random draws. The mean retained share per checkpoint is 74\% (range 47--91\%). Restricting the change to $U$ also preserves mean ImageNet accuracy to two decimal places. Each restriction is scored separately, so its recall gain need not add to that of the complementary restriction.

\noindent\textbf{The spectrum tracks the gain across checkpoints.}
By Eq.~\eqref{eq:supp_gain}, the first-order margin gain of a round is proportional to $\sum_{i\le d/4}|\lambda_i|$, which is available from the construction pairs before any edit is evaluated.
Across the 21 checkpoints, this value has Pearson correlation $0.90$ with the measured COCO val gain of \ours and $0.91$ with that of the learned map.
Its partial correlations given the zero-shot RSUM are $0.82$ and $0.84$, so it carries information beyond a checkpoint's starting accuracy.
These associations describe the 21 evaluated checkpoints. Section~\ref{supp:rank} tests whether those directions also favor hard negatives on new pairs.

\begin{table}[t]
\centering
\caption{\textbf{Retrieval retained inside the selected subspace.} Each restricted map keeps the learned metric change in the named subspace and restores the identity elsewhere. $U$ comprises the first $d/4$ directions selected by \ours. Random averages five equal-rank draws per checkpoint. Entries are validation RSUM gains and ImageNet top-1 changes (IN, points) from zero-shot.}
\label{tab:learnedmetric}
\tabfont
\begin{tabular*}{\linewidth}{@{\extracolsep{\fill}}l rrr@{}}
\toprule
Metric change kept & COCO val & Flickr30k val & IN \\
\midrule
\multicolumn{4}{@{}l}{\textit{All 21 CLIP checkpoints}} \\
\gp Everywhere (hinge map) & $+17.1$ & $+6.8$ & $-0.22$ \\
\gp Inside $U$ & $+12.3$ & $+5.2$ & $0.00$ \\
\gp Outside $U$ & $+6.0$ & $+2.8$ & $-0.15$ \\
\gp Inside a random subspace & $+1.8$ & $+0.9$ & -- \\
\gp \ours, for reference & $+16.1$ & $+6.3$ & $-0.16$ \\
\midrule
\multicolumn{4}{@{}l}{\textit{The 15 checkpoints excluded from schedule selection}} \\
\gp Everywhere (hinge map) & $+14.5$ & $+5.0$ & $-0.30$ \\
\gp Inside $U$ & $+11.3$ & $+4.3$ & $0.00$ \\
\gp Outside $U$ & $+5.6$ & $+2.3$ & $-0.10$ \\
\gp Inside a random subspace & $+1.6$ & $+0.7$ & -- \\
\gp \ours, for reference & $+14.1$ & $+5.0$ & $-0.12$ \\
\bottomrule
\end{tabular*}
\end{table}

\subsection{Building the Correction in Other Domains}
\label{supp:ood}
A new domain poses two questions. Does the COCO-built map improve retrieval there? How much can the same rule gain when it sees the domain's own hard negatives? RSICD pairs remote-sensing images with five captions each, and ROCOv2 pairs radiology images with one caption each.\footnote{Lu et al., \emph{Exploring Models and Data for Remote Sensing Image Caption Generation}, IEEE TGRS, 2018. R\"uckert et al., \emph{ROCOv2: Radiology Objects in COntext Version 2}, Scientific Data, 2024.} We evaluate both settings on the same held-out galleries of 1{,}093 RSICD and 1{,}655 ROCOv2 test images. Zero-shot RSUM averages $126.4$ and $96.4$, respectively.

With no target-domain pairs, the COCO-built \ours map gains $3.9$\ci{2.8}{5.1} RSUM on RSICD and $0.8$\ci{0.2}{1.5} on ROCOv2 (Table~\ref{tab:ood}). The paired gains over the COCO-built hinge map are $1.3$\ci{0.5}{2.1} and $0.8$\ci{0.2}{1.4} RSUM, respectively. Rebuilding \ours from 5{,}000 training images per domain, with the same rank, strengths and negative counts, raises those gains to $20.6$\ci{18.2}{23.0} and $12.8$\ci{11.5}{14.1}. All 21 checkpoints improve in both domains. All deploy two rounds on RSICD. On ROCOv2 one falls back to a single round.

We compare two ways of fitting the hinge map on the same in-domain pairs. With its COCO-selected schedule and a 25-step fallback, hinge gains $19.6$ RSUM on RSICD test and $0.4$ on ROCOv2 test. The paired differences (\ours minus hinge) are $1.05$\ci{-1.10}{3.19} and $12.5$\ci{11.2}{13.7} RSUM, respectively. On RSICD validation, hinge leads by $4.1$\ci{2.0}{6.2}.

We next tune the hinge map separately for each checkpoint using four learning rates ($10^{-5},3{\times}10^{-5},10^{-4},3{\times}10^{-4}$) at 5, 10, 25, 50, 100 and 200 steps. We maximize in-domain validation RSUM among settings that pass the guard, using the RSICD validation split or the 1{,}663 ROCOv2 training images outside construction. The tuned map clears the guard on all 21 checkpoints in each domain. On RSICD test its gain is close to the fixed \ours rule (\ours minus hinge, $-0.45$\ci{-2.65}{+1.80} RSUM). On ROCOv2 test, \ours exceeds the tuned map on all 21 checkpoints, by $7.5$\ci{6.6}{8.4} RSUM on average.

\begin{table}[t]
\centering
\caption{\textbf{Using a COCO-built map or rebuilding in the target domain.} Mean test RSUM gains over 21 CLIPs. Guard counts models meeting the ImageNet tolerance before fallback. In-domain rows use 5{,}000 training images. Deployed means include the specified fallback and zero-shot when needed. The final row selects a guard-passing hinge schedule on separate in-domain validation images.}
\label{tab:ood}
\tabfont
\begin{tabular*}{\linewidth}{@{\extracolsep{\fill}}l rr rr@{}}
\toprule
 & \multicolumn{2}{c}{RSICD} & \multicolumn{2}{c@{}}{ROCOv2} \\
\cmidrule(lr){2-3}\cmidrule(l){4-5}
Edit & RSUM & Guard & RSUM & Guard \\
\midrule
\multicolumn{5}{@{}l}{\textit{Before fallback}} \\
\gp \ours, COCO-built & $+3.9$ & 21/21 & $+0.8$ & 21/21 \\
\gp Hinge map, COCO-built & $+2.7$ & 21/21 & $+0.1$ & 21/21 \\
\gp \ours, in-domain & $+20.6$ & 21/21 & $+13.1$ & 20/21 \\
\gp Hinge map, 50 steps & $+20.3$ & 2/21 & $-1.6$ & 8/21 \\
\gp Hinge map, 25 steps & $+20.8$ & 19/21 & $+2.2$ & 19/21 \\
\midrule
\multicolumn{5}{@{}l}{\textit{Deployed}} \\
\gp \ours, in-domain & $+20.6$ & & $+12.8$ & \\
\gp Hinge map, 50 steps & $+2.3$ & & $-1.0$ & \\
\gp Hinge map, 50 then 25 steps & $+19.6$ & & $+0.4$ & \\
\gp Hinge map, domain-tuned & $+21.1$ & & $+5.3$ & \\
\bottomrule
\end{tabular*}
\end{table}

\section{Retention and Compositional Accuracy}
\label{supp:retention}
Does a retrieval edit preserve the decisions made by the original checkpoint? We test its zero-shot classifier on images outside the ImageNet guard, then use caption foils to examine sensitivity to added objects and attributes.

\subsection{Retention on Independent ImageNet-V2 Images}
ImageNet-V2 supplies 10{,}000 new images for the same classes and the same 80-prompt classifier.\footnote{Recht et al., \emph{Do ImageNet Classifiers Generalize to ImageNet?}, ICML 2019.} On its matched-frequency split, mean accuracy changes by $-0.14$\ci{-0.26}{-0.02} points for \ours, close to the $-0.16$ on the guard set, and by $-0.27$\ci{-0.40}{-0.15} for the fixed hinge map. With images resampled jointly across checkpoints, the paired difference favors \ours by $0.13$\ci{0.03}{0.23} points over all 21 and by $0.21$\ci{0.11}{0.32} over the 15 excluded from schedule selection. One \ours checkpoint loses more than half a point (B/32 OpenAI, $0.55$), compared with four hinge checkpoints (largest loss $0.73$).

\subsection{Classification Beyond ImageNet}
\label{supp:suite}
We evaluate the same frozen maps on 13 classification test sets unused for construction, selection or the guard. Ten cover general recognition, fine-grained classes, textures, satellite images and traffic signs. Three test distribution shifts in ImageNet (Table~\ref{tab:suite}). For each set we use its CLIP Benchmark class names, prompts and up to 10{,}000 test images.\footnote{LAION-AI, \emph{CLIP Benchmark}, \url{https://github.com/LAION-AI/CLIP_benchmark}.} Intervals resample images within each set jointly across checkpoints.

Across the 13 sets, mean top-1 accuracy changes by $+0.07$\ci{+0.03}{+0.11} points for \ours and $-0.07$\ci{-0.10}{-0.03} for hinge. The paired advantage is $0.14$\ci{0.11}{0.17} points over all 21 checkpoints and $0.19$\ci{0.16}{0.23} over the 15 excluded from schedule selection. On the three ImageNet shifts, the mean changes are $-0.04$\ci{-0.10}{+0.02} and $-0.27$\ci{-0.34}{-0.20}, respectively. Building \ours from 5{,}000 web pairs instead gives a deployed 13-set mean of $-0.03$\ci{-0.07}{+0.01} points (Sec.~\ref{supp:web}).

The 13-set mean falls by at most $0.30$ points on any checkpoint. Individual tasks vary. The COCO-built correction gains $1.19$ points on EuroSAT and loses $0.54$ on Flowers-102. Its paired advantage over hinge has an interval above zero on seven sets, including all three ImageNet shifts, and below zero on Caltech-101 and Pets.

\begin{table}[t]
\centering
\caption{\textbf{Classification on 13 test sets outside the ImageNet guard.} Zero-shot gives mean top-1 accuracy (\%) over 21 CLIPs. Method columns give changes in percentage points. Diff. is the paired difference, \ours minus hinge. The final rows average the three ImageNet shifts and all 13 sets, respectively, weighting each set equally.}
\label{tab:suite}
\tabfont
\begin{tabular*}{\linewidth}{@{\extracolsep{\fill}}l rrrr@{}}
\toprule
Test set & Zero-shot & \ours & Hinge map & Diff. \\
\midrule
CIFAR-100 & $79.2$ & $-0.04$ & $-0.09$ & $+0.05$ \\
EuroSAT & $58.9$ & $+1.19$ & $+0.77$ & $+0.42$ \\
GTSRB & $51.5$ & $+0.83$ & $+0.14$ & $+0.69$ \\
Caltech-101 & $84.4$ & $-0.37$ & $-0.03$ & $-0.34$ \\
DTD & $58.9$ & $-0.07$ & $-0.15$ & $+0.07$ \\
Oxford Pets & $91.9$ & $-0.25$ & $-0.12$ & $-0.13$ \\
FGVC Aircraft & $32.1$ & $+0.49$ & $+0.56$ & $-0.06$ \\
Stanford Cars & $84.9$ & $-0.04$ & $-0.22$ & $+0.18$ \\
Food-101 & $89.5$ & $-0.16$ & $-0.35$ & $+0.19$ \\
Flowers-102 & $76.3$ & $-0.54$ & $-0.56$ & $+0.02$ \\
\midrule
ImageNet-A & $50.1$ & $-0.13$ & $-0.47$ & $+0.34$ \\
ImageNet-R & $83.3$ & $+0.03$ & $-0.19$ & $+0.22$ \\
ImageNet-Sketch & $60.1$ & $-0.02$ & $-0.14$ & $+0.13$ \\
\midrule
ImageNet shifts (3) & $64.5$ & $-0.04$ & $-0.27$ & $+0.23$ \\
All 13 sets & $69.3$ & $+0.07$ & $-0.07$ & $+0.14$ \\
\bottomrule
\end{tabular*}
\end{table}

\subsection{Extending the Correction to Caption-Foil Distinctions}
\label{supp:addcost}
The default edit changes overall SugarCrepe accuracy by $+0.06$\ci{-0.10}{+0.23} points. Replace and swap improve by $1.1$ and $1.6$ points, while add decreases by $1.9$. When distinguishing an added object or attribute matters, the constrained variant uses caption foils built from the construction data. In each round it selects directions that preserve the mean bilinear margin between true captions and foils (Sec.~\ref{supp:constraint_recipe}).

Across the main 21 checkpoints, the constrained edit gains $15.9$ COCO val and $6.1$ Flickr30k val RSUM. The differences from the default are $-0.20$\ci{-0.40}{+0.00} and $-0.17$\ci{-0.42}{+0.08} RSUM. Mean add accuracy rises $0.34$\ci{0.08}{0.60} points over zero-shot and overall accuracy rises $1.30$ points (Table~\ref{tab:addcost}). Every checkpoint improves on the default's add accuracy and passes the guard. Changes in add accuracy from zero-shot range from $-1.1$ to $+3.4$ points. On the 15 checkpoints excluded from constraint exploration, the mean add change is $+0.45$\ci{+0.17}{+0.72} points.

On the main paper's four prospective checkpoints, the constrained edit gains $1.69$\ci{1.41}{1.96} add points over the default, with a Flickr30k val difference of $-0.23$\ci{-0.71}{+0.28} RSUM. We repeat this comparison with five disjoint retrieval samples and the same foil set, recomputing both statistics each round. All 20 runs pass the guard, improve retrieval on both validation sets and exceed the default on add accuracy. Fifteen also reach or exceed zero-shot add accuracy.

For retrieval, the default uses matched pairs alone and is the rule evaluated on the sealed retrieval tests. For applications that also need caption-foil distinctions, the constrained edit uses rule-generated foils and about one extra second of construction on cached embeddings (Sec.~\ref{supp:constraint_recipe}).

\begin{table}[t]
\centering
\caption{\textbf{Retrieval and SugarCrepe add accuracy after deployment.} The upper block averages the 21 main checkpoints and the lower block four prospective checkpoints (group C, Table~\ref{tab:cohorts}). Retrieval uses RSUM. SugarCrepe add (SC add, 95\% CI) and ImageNet top-1 (IN) use points. Changes are relative to zero-shot. Bold marks the default recipe. REC is the agent-derived recipe in Sec.~\ref{supp:agent_composition}.}
\label{tab:addcost}
\tabdense
\begin{tabular*}{\linewidth}{@{\extracolsep{\fill}}l rr E r@{}}
\toprule
Arm & COCO val & Flickr30k val & \multicolumn{2}{c}{SC add} & IN \\
\midrule
\multicolumn{6}{@{}l}{\textit{21 main-cohort CLIPs}} \\
\ours & $\mathbf{+16.1}$ & $\mathbf{+6.3}$ & $\mathbf{-1.93}$ & \cit{-2.20}{-1.68} & $\mathbf{-0.16}$ \\
Constrained (CON$^2$) & $+15.9$ & $+6.1$ & $+0.34$ & \cit{+0.08}{+0.60} & $-0.12$ \\
\midrule
\multicolumn{6}{@{}l}{\textit{Four prospective checkpoints}} \\
\ours & $\mathbf{+14.1}$ & $\mathbf{+5.6}$ & $\mathbf{-1.45}$ & \cit{-1.83}{-1.08} & $\mathbf{-0.14}$ \\
Constrained (CON$^2$) & $+13.8$ & $+5.4$ & $+0.23$ & \cit{-0.14}{+0.61} & $-0.11$ \\
REC (agent loop) & $+4.6$ & $+2.3$ & $-1.17$ & \cit{-1.50}{-0.85} & $-0.07$ \\
\bottomrule
\end{tabular*}
\end{table}

\section{Transfer and Robustness}
\label{supp:robustness}
We first apply the fixed rule to additional encoders, including models with high starting retrieval scores. We then vary the source of the pairs, the relevance labels and the sample used for construction.

\subsection{Other Dual-Encoder Families}
\label{supp:otherfamilies}
The main paper evaluated five additional encoders. Four improve Flickr30k val retrieval significantly and pass the ImageNet guard (Table~\ref{tab:nonclip}). SigLIP2 B/16 gains retrieval but loses $0.70$ ImageNet points, so it retains zero-shot. Including this fallback, the five-model mean gain is $4.4$\ci{3.5}{5.4} Flickr30k test RSUM. EVA02 and MobileCLIP absorb $S$ into the existing head and bias. SigLIP uses an added linear operation after its vision encoder.

\begin{table}[t]
\centering
\caption{\textbf{The five additional encoders in the main paper.} Flickr30k RSUM gains and ImageNet top-1 changes (IN, points) from zero-shot. Intervals are 95\% CIs. Parentheses ($^{\dagger}$) show the two-round SigLIP2 B/16 edit before fallback. Its deployed change is zero on all three measures.}
\label{tab:nonclip}
\tabfont
\begin{tabular*}{\linewidth}{@{\extracolsep{\fill}}l E rr@{}}
\toprule
Model & \multicolumn{2}{c}{Flickr30k val} & Flickr30k test & IN \\
\midrule
EVA02-B/16 & $+4.3$ & \cit{+2.6}{+6.1} & $+6.6$ & $-0.22$ \\
MobileCLIP-B & $+4.2$ & \cit{+2.7}{+5.7} & $+4.2$ & $-0.40$ \\
CoCa ViT-B/32 & $+7.0$ & \cit{+4.8}{+9.4} & $+6.5$ & $-0.30$ \\
SigLIP B/16 & $+4.3$ & \cit{+2.6}{+6.0} & $+4.8$ & $-0.42$ \\
SigLIP2 B/16$^{\dagger}$ & \llap{(}$+2.3$\rlap{)} & & \llap{(}$+2.6$\rlap{)} & \llap{(}$-0.70$\rlap{)} \\
\bottomrule
\end{tabular*}
\end{table}

\noindent\textbf{Does the correction still help at higher starting accuracy?}
We apply the same rule to MetaCLIP~2 H/14, EVA02-E/14+~\cite{evaclip}, SigLIP2 So400m/14~\cite{siglip2} and Perception Encoder Core L/14.\footnote{Chuang et al., \emph{Meta CLIP 2: A Worldwide Scaling Recipe}, NeurIPS 2025, arXiv:2507.22062. Bolya et al., \emph{Perception Encoder: The Best Visual Embeddings Are Not at the Output of the Network}, NeurIPS 2025, arXiv:2504.13181.} We build each map from the same 5{,}000 COCO training images and apply it to both towers' output embeddings. All four deploy two rounds and improve retrieval on COCO val and both Flickr30k splits (Table~\ref{tab:strong}). The largest of the four COCO gains occurs on MetaCLIP~2 H/14. Even PE Core, starting at $486.3$ COCO val RSUM, gains $8.5$. SigLIP2 So400m/14 meets the ImageNet guard with a change of $-0.03$ points.

\begin{table*}[t]
\centering
\caption{\textbf{Retrieval on four high-accuracy encoders.} Each retrieval column gives the two-round gain over zero-shot in RSUM with a paired 95\% CI. The ImageNet columns give original top-1 accuracy and its change in percentage points. All four edits pass the guard. PE Core uses 336-pixel images.}
\label{tab:strong}
\tabfont
\begin{tabular*}{\linewidth}{@{\extracolsep{\fill}}l E E E rr@{}}
\toprule
& \multicolumn{2}{c}{COCO val} & \multicolumn{2}{c}{Flickr30k val} & \multicolumn{2}{c}{Flickr30k test} & \multicolumn{2}{c}{ImageNet} \\
Model & \multicolumn{2}{c}{RSUM gain} & \multicolumn{2}{c}{RSUM gain} & \multicolumn{2}{c}{RSUM gain} & Zero-shot (\%) & Change \\
\midrule
MetaCLIP 2 H/14 & $+15.2$ & \cit{+13.7}{+16.6} & $+6.5$ & \cit{+4.8}{+8.2} & $+5.9$ & \cit{+4.2}{+7.7} & $81.3$ & $-0.18$ \\
EVA02-E/14+ & $+11.5$ & \cit{+10.4}{+12.7} & $+2.9$ & \cit{+1.9}{+4.0} & $+4.1$ & \cit{+2.9}{+5.4} & $82.0$ & $+0.23$ \\
SigLIP2 So400m/14 & $+11.3$ & \cit{+9.9}{+12.7} & $+2.2$ & \cit{+0.9}{+3.5} & $+2.6$ & \cit{+1.2}{+4.1} & $83.2$ & $-0.03$ \\
PE Core L/14 & $+8.5$ & \cit{+7.2}{+9.7} & $+1.8$ & \cit{+0.7}{+2.9} & $+1.3$ & \cit{+0.3}{+2.4} & $83.5$ & $-0.23$ \\
\bottomrule
\end{tabular*}
\end{table*}

\subsection{Retrieval Gain Across Starting Scores}
\begin{figure}[t]
  \centering
  \includegraphics[width=\linewidth]{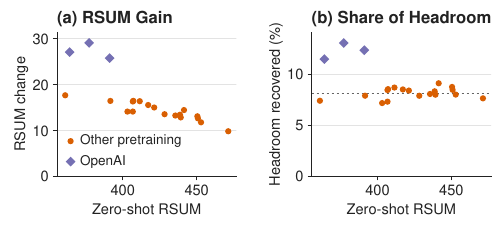}
  \caption{\textbf{Retrieval gain across starting scores.} Each point is one of 21 CLIPs on COCO val. Panel (a) shows the two-round gain in RSUM. Panel (b) divides that gain by $600$ minus zero-shot RSUM. Diamonds mark OpenAI checkpoints. The dashed line averages the other 18.}
  \label{fig:strength}
\end{figure}

Across 21 checkpoints, the COCO validation gain ranges from $9.8$ to $29.1$ RSUM and tends to be smaller at higher zero-shot scores (Pearson $r=-0.79$). Figure~\ref{fig:strength} also expresses the gain as a share of the remaining distance to 600, the maximum RSUM. The 18 checkpoints outside OpenAI pretraining recover $7.2$--$9.1\%$ of that distance and the three OpenAI checkpoints recover $11.5$--$13.0\%$.

\subsection{Retrieval Beyond COCO-Style Captions}
\label{supp:web}
Conceptual Captions~\cite{cc3m} pairs each image with one web alt-text caption. On a gallery of 10{,}000 validation pairs, the COCO-built map improves all 13 checkpoints with agent sessions. Their mean gain is $4.1$\ci{3.6}{4.5} RSUM from a zero-shot mean of $366.5$. Recall@1 rises $1.2$ points for text-to-image and $0.3$ for image-to-text retrieval.

To test whether the web pairs themselves suffice for construction, we divide the gallery into two disjoint halves. On all 21 checkpoints, we build the map from the same 5{,}000 pairs and evaluate on the other 5{,}000. Before fallback it gains $10.4$\ci{9.6}{11.4} RSUM on that gallery, compared with $3.7$ for the COCO-built map (Table~\ref{tab:webbuilt}). The web-built map also improves both COCO val and Flickr30k val on every checkpoint. The two first-round subspaces overlap by $0.35$ (random reference $0.25$).

With 5{,}000 web pairs, 15 checkpoints deploy two rounds, two deploy one and four retain zero-shot. Their mean gain on held-out web pairs is $7.9$\ci{7.2}{8.7} RSUM. Building from all 10{,}000 pairs yields a deployed edit on 20 checkpoints and raises the COCO val gain from $4.3$\ci{3.8}{4.8} to $5.9$\ci{5.3}{6.5}. Only the smaller construction leaves an independent web gallery.

\begin{table}[t]
\centering
\caption{\textbf{Construction from web captions.} Mean changes from zero-shot over 21 CLIPs. COCO and Flickr30k report validation RSUM. Web uses the same 5{,}000 evaluation pairs throughout. Dashes mark overlap with construction. IN is ImageNet top-1 change in points. Guard counts accepted two-round edits. Deployed means include fallbacks.}
\label{tab:webbuilt}
\tabfont
\begin{tabular*}{\linewidth}{@{\extracolsep{\fill}}l rrr rr@{}}
\toprule
Built from & COCO & Flickr30k & Web & IN & Guard \\
\midrule
\multicolumn{6}{@{}l}{\textit{Before fallback}} \\
\gp Web, 5{,}000 pairs & $+5.5$ & $+3.1$ & $+10.4$ & $-0.40$ & 15/21 \\
\gp Web, 10{,}000 pairs & $+6.2$ & $+3.2$ & -- & $-0.26$ & 18/21 \\
\gp COCO, 25{,}000 pairs & $+16.1$ & $+6.3$ & $+3.7$ & $-0.16$ & 21/21 \\
\midrule
\multicolumn{6}{@{}l}{\textit{Deployed}} \\
\gp Web, 5{,}000 pairs & $+4.3$ & $+2.4$ & $+7.9$ & & \\
\gp Web, 10{,}000 pairs & $+5.9$ & $+3.1$ & -- & & \\
\bottomrule
\end{tabular*}
\end{table}

\subsection{Extended Ground Truth on COCO Test}
\label{supp:eccv}
A retrieved caption can describe an image even when it is absent from the five COCO references. ECCV Caption (Chun et al., arXiv:2204.03359) adds verified matches to COCO test. Its 1{,}261 image queries average 17.9 valid captions, and its 1{,}332 caption queries average 8.5 valid images. We score the same frozen maps with the official package, whose tie handling changes standard COCO RSUM by less than $0.2$ relative to our scorer.

All six metrics in Table~\ref{tab:eccv} improve on every checkpoint. Recall@1 rises $3.3$ points in both directions. Text-to-image mAP@R rises from $38.1$ to $40.7$ and image-to-text mAP@R from $27.3$ to $29.0$. On the eight checkpoints without sessions the corresponding gains are $2.5$ and $1.6$ points. Thus the gain persists when multiple items can be correct for a query.

\begin{table}[t]
\centering
\caption{\textbf{Retrieval with additional verified positives on COCO test.} ECCV Caption scores (\%) are averaged over 21 CLIPs. Changes from zero-shot are in points with paired 95\% CIs. Improve counts checkpoints with a positive change.}
\label{tab:eccv}
\tabfont
\begin{tabular*}{\linewidth}{@{\extracolsep{\fill}}lrrr@{\hspace{0.35em}}lr@{}}
\toprule
Metric & Zero-shot & \ours & \multicolumn{2}{c}{Change} & Improve \\
\midrule
\multicolumn{6}{@{}l}{\textit{Image $\to$ text}} \\
\gp mAP@R & $27.3$ & $29.0$ & $+1.7$ & \cit{+1.6}{+1.9} & $21/21$ \\
\gp R-Precision & $37.4$ & $39.2$ & $+1.7$ & \cit{+1.6}{+1.9} & $21/21$ \\
\gp Recall@1 & $73.1$ & $76.4$ & $+3.3$ & \cit{+2.8}{+3.9} & $21/21$ \\
\midrule
\multicolumn{6}{@{}l}{\textit{Text $\to$ image}} \\
\gp mAP@R & $38.1$ & $40.7$ & $+2.7$ & \cit{+2.5}{+2.9} & $21/21$ \\
\gp R-Precision & $47.3$ & $49.8$ & $+2.5$ & \cit{+2.3}{+2.7} & $21/21$ \\
\gp Recall@1 & $79.7$ & $83.0$ & $+3.3$ & \cit{+2.8}{+3.8} & $21/21$ \\
\bottomrule
\end{tabular*}
\end{table}

\subsection{Robustness to the Construction Sample}
\label{supp:samples}
We rebuild the fixed recipe on five further disjoint samples of 5{,}000 COCO training images. Each excludes the original sample and is shared by all 21 checkpoints. Before fallback, all 105 constructions improve both COCO val and Flickr30k val retrieval, each with a 95\% interval above zero (Table~\ref{tab:samples}). Across samples, standard deviations average $0.28$ and $0.37$ RSUM, below the evaluation-image standard errors of $0.78$ and $0.98$. The largest within-checkpoint ranges are $1.2$ and $1.5$ RSUM.

The guard accepts 104 of 105 two-round edits. One B/16 LAION-2B construction loses $0.53$ ImageNet points and reverts to zero-shot after one round also fails. All original-sample edits pass. Across the five new samples the ImageNet change has a mean standard deviation of $0.05$ points, and the mean SugarCrepe add change is $-1.85$ points ($-1.93$ on the original). Across sample pairs the first-round subspaces overlap by $0.48$ (random reference $0.25$), yet retrieval gains remain stable. Section~\ref{supp:rank} tests the sign of each direction on independent pairs.

\begin{table}[t]
\centering
\caption{\textbf{Retrieval across disjoint construction samples.} Each sample contains 5{,}000 COCO training images. Orig. gives the original sample and New the mean $\pm$ standard deviation over five others. Entries are changes from zero-shot before fallback in validation RSUM and ImageNet top-1 (IN, points). B/32 LAION-2B denotes the e16 model. The final row averages checkpoint means and standard deviations separately.}
\label{tab:samples}
\tabdense\setlength{\tabcolsep}{2pt}
\begin{tabular*}{\linewidth}{@{\extracolsep{\fill}}l rr rr rr@{}}
\toprule
 & \multicolumn{2}{c}{COCO val} & \multicolumn{2}{c}{Flickr30k val} & \multicolumn{2}{c}{IN} \\
\cmidrule(lr){2-3}\cmidrule(lr){4-5}\cmidrule(l){6-7}
Checkpoint & Orig. & New & Orig. & New & Orig. & New \\
\midrule
B/16 OpenAI & $+29.1$ & $+28.7\pm0.33$ & $+16.4$ & $+16.3\pm0.35$ & $-0.06$ & $+0.03\pm0.04$ \\
L/14 DFN-2B & $+13.1$ & $+13.0\pm0.35$ & $+4.4$ & $+4.5\pm0.39$ & $-0.14$ & $-0.14\pm0.02$ \\
B/32 LAION-2B & $+14.1$ & $+13.9\pm0.15$ & $+4.3$ & $+4.8\pm0.30$ & $-0.05$ & $-0.11\pm0.04$ \\
\midrule
21 CLIPs, mean & $+16.1$ & $+16.0\pm0.28$ & $+6.3$ & $+6.4\pm0.37$ & $-0.16$ & $-0.16\pm0.05$ \\
\bottomrule
\end{tabular*}
\end{table}

\subsection{Transfer Across Construction Data}
Holding construction size fixed at 5{,}000 five-caption images, we change only the source of the pairs (Table~\ref{tab:calib}). Edits built on Flickr30k improve COCO val by $8.0$--$22.2$ RSUM on the three tested checkpoints. Matching the source to the evaluation adds $5$--$7$ RSUM on COCO and up to $3.6$ on Flickr30k. An equal mixture retains gains on both datasets. All nine constructions pass the ImageNet guard.

\begin{table}[t]
\centering
\caption{\textbf{Transfer across construction data.} RSUM change from zero-shot on COCO and Flickr30k validation. Each edit uses 5{,}000 training images with five captions each, drawn from COCO, Flickr30k or half from each (Mix). Every edit passes the ImageNet guard.}
\label{tab:calib}
\tabfont
\begin{tabular*}{\linewidth}{@{\extracolsep{\fill}}llrr@{}}
\toprule
Checkpoint & Built from & COCO val & Flickr30k val \\
\midrule
B/16 OpenAI & COCO & $+29.1$ & $+16.4$ \\
 & Flickr30k & $+22.2$ & $+19.9$ \\
 & Mix & $+27.2$ & $+19.1$ \\
\midrule
B/16 DataComp-XL & COCO & $+16.4$ & $+7.9$ \\
 & Flickr30k & $+10.1$ & $+9.1$ \\
 & Mix & $+14.0$ & $+8.7$ \\
\midrule
L/14 DFN-2B & COCO & $+13.1$ & $+4.4$ \\
 & Flickr30k & $+8.0$ & $+5.0$ \\
 & Mix & $+12.0$ & $+5.4$ \\
\bottomrule
\end{tabular*}
\end{table}

\section{How the Correction Improves Retrieval}
\label{supp:geometry}
\ours improves retrieval while largely preserving the distance between the modality means.
We examine the directions it attenuates, then connect the resulting score changes to wider hard-negative margins and fewer retrievals captured by hubs.

\subsection{Why Attenuate Rather Than Remove?}
\label{supp:shrink}
\noindent\textbf{Gap removal in the sessions.}
All 13 agents included gap removal in their first batch of edits.
Each estimated $g$, the normalized difference of the modality means, from its first \texttt{crossmap} call on 5{,}000 construction images and 25{,}000 captions.
It applied $I-a\,gg^{\top}$ to both projections, testing full removal ($a=1$) and smaller doses between $0.25$ and $0.75$; most also tested the common mean or the plane of both means.
Full removal cost 44--112 RSUM on the 5{,}000 held-out training images, and every tested dose of at least $0.25$ lowered RSUM.
We extend this analysis to 24 checkpoints on COCO val, using the paired statistic's leading directions.
Removal refers to the projection before cosine normalization; re-normalizing can change the modality means again.

\noindent\textbf{The leading directions of the paired statistic.}
Let $u_1$ and $u_2$ be eigenvectors of the unedited model's $\mathrm{sym}(\dD)$ associated with its two smallest eigenvalues.
Let $\mu_{\mathrm{img}}$ and $\mu_{\mathrm{txt}}$ be the means of the unit image and caption embeddings on COCO val.
We measure alignment of a unit direction $u$ with a vector $w$ by $|\cos|=|u^{\top}w|/\lVert w\rVert$.
Over 24 CLIP checkpoints (13 with sessions and 11 others), $u_1$ aligns with the common mean: its $|\cos|$ with $\mu_{\mathrm{img}}$ is $0.74$.
Its alignment with the first principal component is weaker ($|\cos|=0.19$, measured on 25 checkpoints).
The second direction aligns with the gap, with $|\cos|=0.61$ against $\mu_{\mathrm{img}}-\mu_{\mathrm{txt}}$.
Figure~\ref{fig:anatomy}a names the directions $u_{\mathrm{mean}}$ and $u_{\mathrm{gap}}$ for these alignments; each is an eigenvector of the paired statistic, not the corresponding mean vector itself.
Removing $u_1$, or both $u_1,u_2$, harms every checkpoint, with mean COCO val losses of $36.5$ and $48.6$ RSUM.
Shrinking the two directions by 10\% instead gains $+7.8$, while shrinking the remaining directions alone gains $+4.5$; the full $d/4$ shrink gives $+12.5$ (Fig.~\ref{fig:anatomy}a).
In a separate decomposition over the 13 session checkpoints, $u_1,u_2$ account for 63\% of the one-round COCO val gain, $u_3$--$u_8$ for 15\%, $u_9$--$u_{32}$ for 15\%, and the remaining selected directions for 7\%.
Increasing the rank beyond $d/4$ does not improve mean retrieval.

\begin{figure*}[t]
  \centering
  \includegraphics[width=\linewidth]{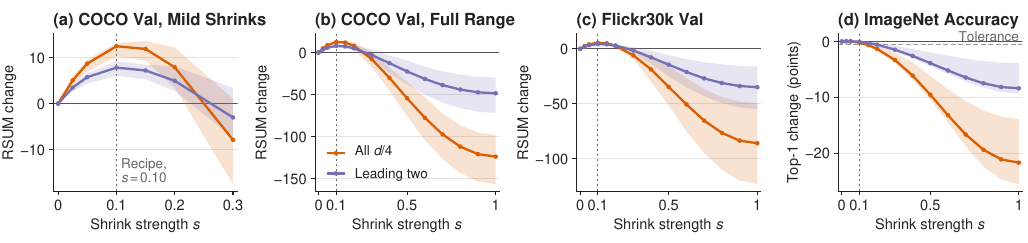}
  \caption{\textbf{Mild attenuation versus removal.} One-round changes from zero-shot when shrinking all $d/4$ hard-negative directions (orange) or only the leading two (purple); $s=1$ removes them. Panel (a) enlarges $s\le0.3$ from (b). Lines are means over 24 CLIPs; bands are interquartile ranges. Panel (d) marks the 0.5-point ImageNet tolerance. Results are shown before fallback.}
  \label{fig:dose}
\end{figure*}

\noindent\textbf{From a mild shrink to removal.}
Figure~\ref{fig:dose} varies one-round strength on the same 24 checkpoints.
For all $d/4$ directions, the mean COCO val gain peaks at $s=0.10$ ($+12.5$ RSUM), remains positive at $0.20$ ($+7.9$), turns negative at $0.30$ ($-7.9$), and reaches $-124.1$ at removal.
Every checkpoint's best tested strength is 0.10--0.20, both for the full shrink and for the leading two directions.
For the full shrink, ImageNet narrows the range: all 24 pass at $s=0.05$, 21 at $0.10$, 9 at $0.15$ and none at $0.30$.
The sweep supports a mild first round; the theory determines its form, not its strength.

\noindent\textbf{What the leading directions carry.}
To see why removal fails where a mild shrink helps, we measure the score contribution $\hat e_v^{\top}P\hat e_t$ of a subspace with projector $P$ on COCO val, for matched pairs, random pairs and the zero-shot hard negatives, over the 21 CLIP checkpoints.
In the plane of $u_1$ and $u_2$, matched pairs score almost exactly as random pairs do ($0.100$ against $0.099$), while hard negatives score higher ($0.119$), on every checkpoint.
Outside the selected subspace the order is reversed: matched pairs score $0.125$, hard negatives $0.086$ and random pairs $-0.025$.
The leading plane therefore carries little pair-specific agreement.
Its coordinates instead follow how often a candidate is retrieved: the absolute Spearman correlation between an image's coordinate and its 10-occurrence count is $0.42$ for $u_1$ and $0.34$ for $u_2$, against $0.02$ for a typical direction.
Across random pairs, the plane's score term is negatively correlated with the term of all other directions ($-0.63$ on average), so it offsets part of their variation.

Table~\ref{tab:removal} shows the corresponding retrieval behavior.
After removal, the skewness of the caption 10-occurrence counts rises from $3.39$ to $5.89$, hubs take 66\% of the top-10 caption slots instead of 33\%, and retrieval loses $49.6$ RSUM.
A 10\% shrink of the same plane lowers image and caption hubness on all 21 checkpoints and gains $+7.8$ RSUM.
These measurements link the leading plane to candidate-level offsets and show that mild attenuation reduces hubness where removal increases it.
They support the ranking interpretation without isolating a causal decomposition of the gain.

\begin{table}[t]
\centering
\caption{\textbf{How strongly should the leading plane contribute?} On 21 CLIPs, we measure COCO val RSUM change and the skewness of top-10 occurrence counts. Lower skewness means fewer retrievals concentrate on hubs. The plane spans $u_1,u_2$, and $U$ contains all $d/4$ selected directions. Figure~\ref{fig:dose} uses the larger cohort of 24.}
\label{tab:removal}
\tabfont
\begin{tabular*}{\linewidth}{@{\extracolsep{\fill}}l rrr@{}}
\toprule
Edit & Image skew & Caption skew & RSUM \\
\midrule
Zero-shot & $2.03$ & $3.39$ & -- \\
10\% shrink of the plane & $1.57$ & $2.66$ & $+7.8$ \\
Removal of the plane & $2.89$ & $5.89$ & $-49.6$ \\
10\% shrink of $U$, one round & $1.46$ & $2.34$ & $+12.5$ \\
\ours & $1.33$ & $1.95$ & $+16.1$ \\
\bottomrule
\end{tabular*}
\end{table}

\subsection{Hard-Negative Agreement on New Pairs}
\label{supp:rank}
For each checkpoint we estimate first-round eigenvectors on one COCO construction sample, then score those directions on another, using six ordered pairs of disjoint samples (Sec.~\ref{supp:samples}). If $u_i$ is an eigenvector from the first sample, its score $u_i^{\top}\mathrm{sym}(\dD_{\rm held})u_i$ measures how strongly the same direction favors hard negatives on the second. All eight leading directions retain a negative sign, as do $99.8\%$ of ranks 9--32. Ranks 1--2, 3--8 and 9--32 retain $98\%$, $81\%$ and $52\%$ of their in-sample magnitude. Only $7\%$ of ranks above $d/8$ through $d/4$ remain negative on the second sample. The selected quarter as a whole retains $68\%$ of its in-sample value.

Two directions per round already recover $10.0$ COCO val RSUM. At $d/16$ the gain reaches $15.3$, about 95\% of the default's $16.1$, with a mean ImageNet change that rounds to zero (Table~\ref{tab:rank}). Every checkpoint passes the guard at all ranks through $d/4$.

Can the construction pairs suggest a rank? We split the 5{,}000 images in half, count the negative directions from one half that also score negative on the other, and average the counts in both orders. We then build both rounds from all 5{,}000 images at that rank. It averages $0.064d$ (range $0.035d$--$0.123d$) and passes the guard on all 21 checkpoints. Relative to $d/4$, the changes are $-0.78$\ci{-0.96}{-0.61} COCO val RSUM, $-0.07$\ci{-0.28}{+0.14} Flickr30k val RSUM and $+0.14$\ci{+0.12}{+0.16} ImageNet points. The smaller subspace retains about 95\% of the COCO val gain. The main paper uses the $d/4$ rule fixed before test evaluation.

\begin{table}[t]
\centering
\caption{\textbf{Retrieval as the number of attenuated directions increases.} Both rounds use $k$ directions, with fresh negatives for round two. Entries are mean changes from zero-shot over 21 CLIPs before fallback. Retrieval uses RSUM and IN uses ImageNet top-1 points. Guard counts checkpoints losing at most 0.5 points.}
\label{tab:rank}
\tabfont
\begin{tabular*}{\linewidth}{@{\extracolsep{\fill}}lrrrr@{}}
\toprule
Size $k$ & COCO val & Flickr30k val & IN & Guard \\
\midrule
$2$ & $+10.0$ & $+5.1$ & $+0.04$ & $21/21$ \\
$8$ & $+12.4$ & $+5.7$ & $+0.09$ & $21/21$ \\
$d/32$ & $+14.3$ & $+6.1$ & $+0.03$ & $21/21$ \\
$d/16$ & $+15.3$ & $+6.3$ & $+0.00$ & $21/21$ \\
$d/8$ & $+15.8$ & $+6.4$ & $-0.06$ & $21/21$ \\
$d/4$ (\ours) & $+16.1$ & $+6.3$ & $-0.16$ & $21/21$ \\
$3d/8$ & $+16.1$ & $+6.3$ & $-0.20$ & $20/21$ \\
$d/2$ & $+16.1$ & $+6.3$ & $-0.23$ & $19/21$ \\
\bottomrule
\end{tabular*}
\end{table}

\subsection{Paired Evidence Beyond the Modality Means}
\label{supp:meanspan}
Because the two leading directions align with the modality means, we test whether a correction built from the means alone would suffice.
The control shrinks $\mathrm{span}\{\mu_{\mathrm{img}},\mu_{\mathrm{txt}}\}$, with the means computed on the construction pairs, using the recipe's rounds and strengths: $0.10$, then $0.05$ after the means are recomputed on the edited, re-normalized embeddings.
It differs from \ours only in replacing the $d/4$ hard-negative directions with these two mean directions.
Shrinking the mean span gains $2.3$\ci{1.8}{2.8} COCO val and $0.9$\ci{0.4}{1.4} Flickr30k val RSUM (Table~\ref{tab:meanspan}). The paired correction gains more on all 21 checkpoints on both datasets, by $13.8$\ci{13.0}{14.5} and $5.4$\ci{4.6}{6.2} RSUM on average. The mean span captures about 14\% of the COCO gain.

We also test whether useful corrections remain outside the mean span.
In each round, let $Q$ be an orthonormal basis of the current construction embeddings' mean span and $C=I-QQ^{\top}$ its orthogonal-complement projector.
We replace $\mathrm{sym}(\dD)$ with $C\,\mathrm{sym}(\dD)\,C$ and take its $d/4$ smallest-eigenvalue directions, keeping the same strengths and re-mining schedule.
This control retains $+5.8$\ci{+5.2}{+6.3} COCO val RSUM, about 36\% of the gain.
The paired hard-negative statistic thus identifies useful corrections beyond a mean-only adjustment, while excluding the mean span also reduces the gain.

\begin{table}[t]
\centering
\caption{\textbf{Paired directions versus the modality means.} Mean change from zero-shot over 21 CLIPs, before fallback: retrieval RSUM and ImageNet top-1 (IN, points). Mean span uses the recipe's strengths on $\mathrm{span}\{\mu_{\mathrm{img}},\mu_{\mathrm{txt}}\}$; Without mean span restricts the paired statistic to its orthogonal complement. Improve counts positive COCO val changes. Bold marks the largest retrieval gains.}
\label{tab:meanspan}
\tabfont
\begin{tabular*}{\linewidth}{@{\extracolsep{\fill}}lrrrr@{}}
\toprule
Edit & COCO val & Flickr30k val & IN & Improve \\
\midrule
\multicolumn{5}{@{}l}{\textit{One round}} \\
\ours & $+12.5$ & $+5.2$ & $-0.18$ & $21/21$ \\
Mean span & $+2.4$ & $+0.9$ & $+0.04$ & $21/21$ \\
\midrule
\multicolumn{5}{@{}l}{\textit{Two rounds}} \\
\ours & $\mathbf{+16.1}$ & $\mathbf{+6.3}$ & $-0.16$ & $21/21$ \\
Mean span & $+2.3$ & $+0.9$ & $-0.02$ & $20/21$ \\
Without mean span & $+5.8$ & $+1.7$ & $-0.20$ & $21/21$ \\
\bottomrule
\end{tabular*}
\end{table}

\subsection{Checking the Correction Under Cosine Retrieval}
\label{supp:cosine}
The theory optimizes a bilinear margin, while retrieval normalizes the edited embeddings.
To test this distinction directly, we construct a statistic from the exact cosine derivative (derived in Sec.~\ref{supp:cosine_derivation}).
For a small metric change $M=I+\varepsilon E$ with symmetric $E$, replace each product $\hat e_v\hat e_t^{\top}$ in Eq.~\eqref{eq:D} by
\[
\begin{aligned}
G(\hat e_v,\hat e_t)={}&\mathrm{sym}\big(\hat e_v\hat e_t^{\top}\big)\\
&-\tfrac12\big(\hat e_v^{\top}\hat e_t\big)
\big(\hat e_v\hat e_v^{\top}+\hat e_t\hat e_t^{\top}\big)
\end{aligned}
\]
Averaging $G$ over matched pairs and hard negatives as in Eq.~\eqref{eq:D} gives the cosine-margin derivative for fixed negatives. We use this matrix in both rounds, re-mining as usual. Across 21 checkpoints its first-round subspace overlaps that of \ours by $0.92$ on average (random reference $0.25$). Both edits pass the guard throughout. The cosine-based edit differs from \ours by $-0.07$\ci{-0.20}{+0.05} COCO val and $-0.13$\ci{-0.30}{+0.04} Flickr30k val RSUM.

We can also vary the numerator and normalization separately in the final cosine score (Eq.~\eqref{eq:supp_cos}). On OpenAI B/16, DataComp-XL B/16 and DFN-2B L/14, changing only the dot product gains $9.2$ Flickr30k val RSUM on average. Changing only the norms gives $-1.1$, and changing both gives $+9.5$. On these checkpoints, most of the retrieval gain is already present in the edited dot product.

\subsection{Comparison with Centering and All-but-the-Top}
\label{supp:centering}
Centering~\cite{suzuki2013centering} subtracts the mean embedding, and all-but-the-top~\cite{mu2018allbutthetop} then removes top principal directions.
Both use unpaired mean and variance statistics.
We test them on the 24 checkpoints of Fig.~\ref{fig:anatomy}a, using unit embeddings of the 5{,}000 construction images and 25{,}000 captions to estimate means and directions (Table~\ref{tab:centering}).

Per-modality centering subtracts each modality's own mean; a halfway shift moves each mean toward the other while preserving their common mean; common centering subtracts $(\mu_{\mathrm{img}}+\mu_{\mathrm{txt}})/2$ from both.
The first two close the construction-set gap before re-normalization.
We also test means estimated on the evaluation set and the original centering score, an inner product without re-normalization.
Per-modality centering costs $20.2$\ci{18.8}{21.5} COCO val and $12.4$\ci{10.7}{14.0} Flickr30k val RSUM under cosine scoring; without re-normalization, the losses are $7.6$\ci{6.2}{8.8} and $6.2$\ci{4.6}{7.9}.
No centering variant improves a checkpoint on COCO val; evaluation-set means improve 2 of 24 on Flickr30k val under cosine scoring.
Per-modality centering followed by re-normalization is also the post-hoc operation in I0T~\cite{an2025i0t}.
We apply that operation to unedited OpenAI B/16 using one caption per image on Flickr30k val and evaluation-set means.
Text-to-image R@1 rises from 71.0 to 73.7, image-to-text R@1 falls from 73.2 to 69.1, and RSUM changes from 516.1 to 514.3.
This check isolates the post-hoc operation and evaluation convention; it does not reproduce I0T's trained variant or its other checkpoints.
The direction-specific gain shows why closing the gap and improving bidirectional retrieval are separate outcomes.

For all-but-the-top, we remove $D=1$ or $D=d/100$ principal directions after centering, either separately per modality or from pooled embeddings with equal modality weight.
Every variant lowers COCO val retrieval on every checkpoint.
One round of \ours instead improves all 24, by $+12.5$ RSUM.

\noindent\textbf{Mild shrinks without pair labels.}
All-but-the-top removes its directions, so we also shrink top principal directions mildly, as \ours does with hard-negative directions.
We shrink 2, $d/100$ or $d/4$ directions by 5--50\%, taking them from the pooled uncentered second moment, the pooled covariance, or each modality's covariance, and apply the result as a linear map without translation.
The best of these 36 settings gains $+2.2$ COCO val RSUM (two directions of the pooled second moment at 10\%), and a 10\% shrink of $d/4$ directions gains at most $+0.3$, against $+12.5$ for one round of \ours at the same strength and rank (Table~\ref{tab:centering}).
The paired statistic identifies useful editing directions beyond these unpaired mean and variance controls.

\begin{table}[t]
\centering
\caption{\textbf{Centering, all-but-the-top and unpaired shrinks.} Mean RSUM change from zero-shot over the 24 CLIPs in Fig.~\ref{fig:anatomy}a. Means and directions use construction embeddings unless marked eval.-set. Cosine scoring re-normalizes edited embeddings; inner-product scoring does not. Gap closure is before re-normalization. $D$: removed principal directions. Improve counts positive COCO val changes; bold marks the largest gains.}
\label{tab:centering}
\tabfont\setlength{\tabcolsep}{3pt}
\begin{tabular*}{\linewidth}{@{\extracolsep{\fill}}lrrr@{}}
\toprule
Edit & COCO val & Flickr30k val & Improve \\
\midrule
\multicolumn{4}{@{}l}{\textit{Centering, cosine score}} \\
Per modality & $-20.2$ & $-12.4$ & $0/24$ \\
Per modality, eval.-set means & $-20.1$ & $-9.2$ & $0/24$ \\
Common mean & $-35.0$ & $-34.4$ & $0/24$ \\
Mean shift, gap closed & $-69.8$ & $-55.3$ & $0/24$ \\
\midrule
\multicolumn{4}{@{}l}{\textit{Centering, inner-product score}} \\
Per modality & $-7.6$ & $-6.2$ & $0/24$ \\
Per modality, eval.-set means & $-7.5$ & $-7.2$ & $0/24$ \\
Common mean & $-42.0$ & $-39.7$ & $0/24$ \\
\midrule
\multicolumn{4}{@{}l}{\textit{All-but-the-top}} \\
Per modality, $D=1$ & $-21.1$ & $-13.6$ & $0/24$ \\
Per modality, $D=d/100$ & $-33.2$ & $-23.5$ & $0/24$ \\
Pooled, $D=1$ & $-22.7$ & $-12.1$ & $0/24$ \\
Pooled, $D=d/100$ & $-38.2$ & $-25.4$ & $0/24$ \\
\midrule
\multicolumn{4}{@{}l}{\textit{10\% shrink of top principal directions}} \\
Pooled second moment, 2 directions & $+2.2$ & $+0.9$ & $24/24$ \\
Pooled second moment, $d/4$ & $+0.3$ & $+0.4$ & $16/24$ \\
Pooled covariance, 2 directions & $-4.5$ & $-3.1$ & $0/24$ \\
Pooled covariance, $d/4$ & $-0.2$ & $+0.0$ & $8/24$ \\
Per modality, 2 directions & $-0.9$ & $-0.3$ & $5/24$ \\
Per modality, $d/4$ & $-3.0$ & $-1.0$ & $0/24$ \\
\midrule
\ours, one round & $\mathbf{+12.5}$ & $\mathbf{+5.5}$ & $24/24$ \\
\bottomrule
\end{tabular*}
\end{table}

\subsection{Why Use a Mild, Uniform Shrink?}
\label{supp:uniform}
We also test whether the uniform step justified by the bilinear analysis remains useful under the deployed cosine score.
On 13 checkpoints, it exceeds eigenvalue-weighted shrinkage at matched first-order bilinear-margin gain by $8.0$ COCO val RSUM on average, with $0.8$ points less ImageNet loss.
Doubling the metric contraction from $c=0.19$ to $c=0.38$ on $d/8$ directions loses $6.3$ RSUM relative to the uniform $d/4$ shrink, despite an 87\% larger first-order value.

\noindent\textbf{The eigenvalue-weighted step.}
The weighted step acts on the same $k=d/4$ directions as the uniform step but contracts each in proportion to its eigenvalue.
Let $\lambda_1\le\dots\le\lambda_k<0$ be the $k$ smallest eigenvalues of $\mathrm{sym}(\dD)$ for the unedited model, with eigenvectors $u_1,\dots,u_k$, computed on the 5{,}000 construction images.
The metric of the weighted step is
\begin{equation}
\label{eq:supp_weighted}
\begin{aligned}
M_w &= I-\sum_{i\le k}c_i\,u_iu_i^{\top},\qquad c_i=\eta\,|\lambda_i|,\\
\eta &= c\,\sum_{i\le k}|\lambda_i|\Big/\sum_{i\le k}\lambda_i^2 ,
\end{aligned}
\end{equation}
with $c=0.19$, the contraction of the uniform step.
The scalar $\eta$ is the only normalization.
By Statement~1 in Sec.~\ref{supp:theory}, the weighted step changes the bilinear margin, with negatives held fixed, by $\sum_{i\le k}c_i|\lambda_i|=\eta\sum_{i\le k}\lambda_i^2=c\sum_{i\le k}|\lambda_i|$, which is the value of the uniform step in Eq.~\eqref{eq:supp_gain}.
Weights proportional to $|\lambda_i|$ are the choice that maximizes this margin on the selected subspace when the Frobenius norm of $M-I$ is bounded in place of the contraction in each direction.
The step is applied like the uniform one, as a symmetric map on both towers, $S_w=I-\sum_{i\le k}\big(1-\sqrt{1-c_i}\big)u_iu_i^{\top}$, so that $S_w^{\top}S_w=M_w$.
The implementation caps each $c_i$ at $0.99$; the cap was not reached on any checkpoint.

The weighted step concentrates contraction on the leading directions: $c_1=0.40$--$0.58$, $c_2=0.13$--$0.23$ and $c_3=0.03$--$0.13$ across the 13 session checkpoints, with smaller values afterward.
The uniform step instead uses $0.19$ throughout.
Both are one-round edits scored by cosine on COCO val, holding the subspace and first-order bilinear gain fixed.
The complementary comparison, contraction $0.38$ on $d/8$ directions, instead holds total contraction fixed at $0.19d/4$.
Distributing a mild correction across more directions works better here, despite equal or larger bilinear gains for the alternatives.
The bound motivates this form; the cosine experiments determine the operating strength.

\noindent\textbf{Relation to spectral discrimination.}
Contrastive PCA compares target and background covariance, and Fisher discrimination compares between-class and within-class scatter.\footnote{Abid et al., \emph{Exploring patterns enriched in a dataset with contrastive principal component analysis}, 2018. Fisher, \emph{The use of multiple measurements in taxonomic problems}, 1936.} In vision--language models, Chuang et al.\ use biased prompts to construct a projection that removes biased directions from text embeddings.\footnote{Chuang et al., \emph{Debiasing Vision-Language Models via Biased Prompts}, 2023, \href{https://arxiv.org/abs/2302.00070}{arXiv:2302.00070}.} Here the contrast comes from the two towers' agreement on matches and mined competitors. A mild contraction changes their shared score, and re-mining identifies the competitors for the next round.

\subsection{Directions Shared Across Checkpoints}
\label{supp:shared_directions}
Checkpoints have arbitrary coordinates and different dimensions, so their eigenvectors cannot be compared entry by entry.
We instead compare the responses of their subspaces to the same inputs.
Take the $n=30{,}000$ COCO val items: 5{,}000 images and their 25{,}000 captions.
For checkpoint $a$, let $P_a$ be the $n\times d_a$ matrix of unit embeddings and $B_a$ an orthonormal basis of the subspace, with two columns for the $(u_1,u_2)$ plane or $d_a/4$ columns for all of $U$.
The scores $Z_a=P_aB_a$ give each item's coordinates in that subspace.
We center the columns of $Z_a$ and let $Q_a$ be an orthonormal basis of their span.
For checkpoints $a,b$, the singular values $\rho_1\ge\rho_2\ge\dots$ of $Q_a^{\top}Q_b$ are the canonical correlations of $Z_a,Z_b$.
We measure their mean square,
\begin{equation}
\label{eq:supp_cca}
\overline{\rho^2}(a, b) = \frac{1}{j}\sum_{i = 1}^{j}\rho_i^2 \in [0, 1],
\end{equation}
where $j$ is the smaller subspace dimension.
This measure does not depend on either checkpoint's coordinates.
For equal dimensions it is $1$ when the subspaces compute the same functions of the inputs up to an invertible linear map and a constant; for unequal dimensions, $1$ indicates containment of the smaller centered score space.
Across the 300 pairs of 25 CLIPs, the $(u_1,u_2)$ plane has mean squared canonical correlation $0.74$, versus $0.03$ for random planes.
The full $d/4$ subspace has a much smaller separation from its random-subspace reference ($0.42$ versus $0.39$).
The broadly shared leading plane and the more model-dependent remaining directions support transferring the rule while recomputing its subspace for each checkpoint.
Part of the leading plane's agreement across models could come from separating images from captions. After centering the two modalities separately, its mean squared canonical correlation is $0.58$, compared with $0.02$ for random planes. Agreement persists within the images alone ($0.41$) and within the captions alone ($0.62$).

\subsection{Candidate Scores Across the Modality Gap}
\begin{table}[t]
\centering
\caption{\textbf{Gap size and candidate-score variation on COCO val.} Gap is $\lVert\mu_{\mathrm{img}}-\mu_{\mathrm{txt}}\rVert$ for unit embeddings. The first numeric column is its zero-shot value; the remaining columns are relative percentage changes after two rounds. Spread is the standard deviation of $\hat e_v^{\top}\mu_{\mathrm{txt}}$ over image candidates (T$\to$I) or $\hat e_t^{\top}\mu_{\mathrm{img}}$ over caption candidates (I$\to$T). $^{\dagger}$No agent session. The gap stays nearly constant while both spreads fall on all 18 checkpoints.}
\label{tab:gap}
\tabfont
\begin{tabular*}{\linewidth}{@{\extracolsep{\fill}}l*{4}{r}@{}}
\toprule
 & Gap & \multicolumn{3}{c@{}}{Change after the edit (\%)} \\
\cmidrule(lr){2-2}\cmidrule(l){3-5}
Checkpoint & Zero-shot & Gap & Spread T$\to$I & Spread I$\to$T \\
\midrule
B/16 DataComp-L & $0.749$ & $+1.1$ & ${-}14.2$ & ${-}15.7$ \\
B/16 DataComp-XL & $0.800$ & $+0.2$ & ${-}11.8$ & ${-}17.8$ \\
B/16 DFN-2B & $0.853$ & ${-}0.7$ & ${-}12.5$ & ${-}17.1$ \\
B/16 LAION-2B & $0.778$ & ${-}0.3$ & ${-}11.5$ & ${-}16.6$ \\
B/16 LAION-400M & $0.710$ & $+0.7$ & ${-}9.9$ & ${-}10.1$ \\
B/16 MetaCLIP-400M & $0.768$ & $+0.2$ & ${-}14.6$ & ${-}11.3$ \\
B/16 MetaCLIP-fullCC & $0.781$ & $+0.2$ & ${-}12.4$ & ${-}13.5$ \\
B/16 OpenAI & $0.825$ & $+1.1$ & ${-}22.0$ & ${-}11.0$ \\
B/32 OpenAI & $0.825$ & $+1.1$ & ${-}23.4$ & ${-}10.5$ \\
L/14 DataComp-XL & $0.806$ & ${-}0.6$ & ${-}9.3$ & ${-}15.1$ \\
L/14 DFN-2B & $0.848$ & ${-}1.0$ & ${-}8.5$ & ${-}16.6$ \\
L/14 LAION-2B & $0.769$ & ${-}0.6$ & ${-}8.9$ & ${-}14.4$ \\
L/14 OpenAI & $0.823$ & $+0.9$ & ${-}20.9$ & ${-}16.9$ \\
L/14 MetaCLIP-fullCC$^{\dagger}$ & $0.809$ & $+0.0$ & ${-}15.0$ & ${-}13.6$ \\
H/14 LAION-2B$^{\dagger}$ & $0.699$ & ${-}0.8$ & ${-}6.1$ & ${-}12.5$ \\
B/32 LAION-2B$^{\dagger}$ & $0.735$ & $+0.0$ & ${-}13.3$ & ${-}16.2$ \\
B/32 DataComp-XL$^{\dagger}$ & $0.823$ & $+0.3$ & ${-}13.4$ & ${-}16.9$ \\
H/14 DFN-5B$^{\dagger}$ & $0.767$ & ${-}1.2$ & ${-}10.4$ & ${-}16.5$ \\
\midrule
Median & $0.790$ & $+0.1$ & ${-}12.4$ & ${-}15.4$ \\
Lower after edit &  & $7/18$ & $18/18$ & $18/18$ \\
\bottomrule
\end{tabular*}
\end{table}

The distance $\lVert\mu_{\mathrm{img}}-\mu_{\mathrm{txt}}\rVert$ remains largely unchanged after editing: its median change is $+0.1\%$ over 18 checkpoints, and it decreases on 7 (Table~\ref{tab:gap}).
To understand how rankings can improve nevertheless, write each unit embedding as its modality mean plus a deviation, $\hat e_v=\mu_{\mathrm{img}}+\delta_v$ and $\hat e_t=\mu_{\mathrm{txt}}+\delta_t$.
The score separates into four terms:
\begin{equation}
\label{eq:supp_gap}
\hat e_v^{\top}\hat e_t =
\underbrace{\mu_{\mathrm{img}}^{\top}\mu_{\mathrm{txt}}}_{\text{constant}}
+ \underbrace{\delta_v^{\top}\mu_{\mathrm{txt}}}_{\text{per image}}
+ \underbrace{\mu_{\mathrm{img}}^{\top}\delta_t}_{\text{per caption}}
+ \underbrace{\delta_v^{\top}\delta_t}_{\text{per pair}} .
\end{equation}
For a fixed text query, the constant and per-caption terms cancel when comparing images. The per-image term can favor the same image across many queries. The roles reverse for an image query. Table~\ref{tab:gap} measures the spread of these candidate terms across the gallery. After editing, it falls on every checkpoint, by a median of 12\% for images and 15\% for captions. The per-pair term also enters every ranking, so the change in spread alone does not determine the recall gain.

\subsection{Margins, Hubness and Modality Mixing}
\label{supp:bridge}
Table~\ref{tab:bridge} measures how the frozen two-round maps change held-out rankings, hubness and modality geometry.

\noindent\textbf{Hard-negative margins.}
For text-to-image retrieval, the margin is the matched image's score minus the highest score of a wrong image.
For image-to-text retrieval, it is the highest matched-caption score minus the highest wrong-caption score.
We average these margins over queries.
The hard-negative mass is the fraction of the absolute spectrum of $\mathrm{sym}(\dD)$ contributed by negative eigenvalues, $\sum_{\lambda_i<0}|\lambda_i|/\sum_i|\lambda_i|$.

\noindent\textbf{Hubness.}
An item's $k$-occurrence counts how often it appears among the top $k=10$ results of queries from the other modality~\cite{radovanovic2010hubs,feldbauer2019hubness}.
We report the skewness of these counts, the share of top-10 slots occupied by items with counts greater than twice the mean (hub occurrence), and their Robin Hood index~\cite{feldbauer2018fast}.
For counts $n_i$ with mean $\bar n$, the last quantity is $\sum_i|n_i-\bar n|/(2\sum_i n_i)$.
For the fraction of misses landing on a hub, we use a separate, fixed hub set: items at or above the zero-shot model's 95th percentile of occurrence counts.
Keeping this set fixed measures how often the edit redirects queries away from the original hubs.

\noindent\textbf{Gap, alignment and mixing.}
We measure gap size by the distance between the modality means, alignment by mean squared distance between matched unit embeddings, and mixing by the fraction of other-modality items among each item's ten nearest neighbors in the image--caption union, excluding itself.
Table~\ref{tab:bridge} reports the image-side mean of the mixing fraction.

\begin{table*}[t]
\centering
\caption{\textbf{Wider retrieval margins and lower hubness.} Frozen two-round edits on validation data. Each pair of columns gives the median across checkpoints (zero-shot $\to$ edited) and the count that improves. For gap, alignment and mixing, arrows count decreases/increases without assigning a preferred direction; uncounted checkpoints are unchanged. Definitions are in Sec.~\ref{supp:bridge}. Other families are EVA02, MobileCLIP, CoCa, SigLIP and SigLIP2; this diagnostic includes SigLIP2's rejected edit before fallback.}
\label{tab:bridge}
\tabfont\setlength{\tabcolsep}{6pt}
\begin{tabular}{@{}l cc cc cc@{}}
\toprule
& \multicolumn{2}{c}{21 CLIPs, COCO val} & \multicolumn{2}{c}{21 CLIPs, Flickr30k val} & \multicolumn{2}{c@{}}{5 other families, COCO val} \\
\cmidrule(lr){2-3}\cmidrule(lr){4-5}\cmidrule(l){6-7}
Measure & Median & Better & Median & Better & Median & Better \\
\midrule
\multicolumn{7}{@{}l}{\textit{Hubness ($k{=}10$)}} \\
\gp $k$-occurrence skewness, images & $1.88$ $\to$ $1.29$ & $21/21$ & $2.03$ $\to$ $1.32$ & $21/21$ & $1.90$ $\to$ $1.06$ & $5/5$ \\
\gp $k$-occurrence skewness, captions & $3.32$ $\to$ $1.90$ & $21/21$ & $2.31$ $\to$ $1.41$ & $21/21$ & $3.22$ $\to$ $2.01$ & $5/5$ \\
\gp Hub occurrence, images & $0.27$ $\to$ $0.15$ & $21/21$ & $0.19$ $\to$ $0.13$ & $21/21$ & $0.25$ $\to$ $0.08$ & $5/5$ \\
\gp Robin Hood index, images & $0.29$ $\to$ $0.23$ & $21/21$ & $0.25$ $\to$ $0.21$ & $21/21$ & $0.27$ $\to$ $0.19$ & $5/5$ \\
\gp T$\to$I misses on a hub (\%) & $28.1$ $\to$ $18.7$ & $21/21$ & $18.4$ $\to$ $14.1$ & $21/21$ & $25.7$ $\to$ $14.3$ & $5/5$ \\
\gp I$\to$T misses on a hub (\%) & $42.8$ $\to$ $29.7$ & $21/21$ & $39.4$ $\to$ $28.3$ & $21/21$ & $43.8$ $\to$ $27.9$ & $5/5$ \\
\midrule
\multicolumn{7}{@{}l}{\textit{Agreement on matches (held out)}} \\
\gp Margin T$\to$I ($\times 10^{3}$) & $-6.2$ $\to$ $-2.3$ & $21/21$ & $24.3$ $\to$ $27.7$ & $21/21$ & $2.4$ $\to$ $5.9$ & $5/5$ \\
\gp Margin I$\to$T ($\times 10^{3}$) & $11.7$ $\to$ $15.9$ & $21/21$ & $38.4$ $\to$ $44.4$ & $21/21$ & $14.9$ $\to$ $20.0$ & $5/5$ \\
\gp Hard-negative mass of $\dD$ & $0.44$ $\to$ $0.37$ & $21/21$ & -- & -- & $0.42$ $\to$ $0.34$ & $5/5$ \\
\midrule
\multicolumn{7}{@{}l}{\textit{Gap and alignment}} \\
\gp Gap size $\lVert\mu_{\mathrm{img}}-\mu_{\mathrm{txt}}\rVert$ & $0.78$ $\to$ $0.78$ & $8{\downarrow}\;13{\uparrow}$ & $0.75$ $\to$ $0.74$ & $17{\downarrow}\;4{\uparrow}$ & $0.85$ $\to$ $0.86$ & $2{\downarrow}\;3{\uparrow}$ \\
\gp Matched-pair distance$^2$ (alignment) & $1.38$ $\to$ $1.40$ & $1{\downarrow}\;20{\uparrow}$ & $1.37$ $\to$ $1.37$ & $10{\downarrow}\;11{\uparrow}$ & $1.55$ $\to$ $1.56$ & $1{\downarrow}\;4{\uparrow}$ \\
\gp Other modality among 10-NN (\%) & $0.0$ $\to$ $0.0$ & $0{\downarrow}\;11{\uparrow}$ & $0.0$ $\to$ $0.0$ & $0{\downarrow}\;11{\uparrow}$ & $0.0$ $\to$ $0.0$ & $0{\downarrow}\;2{\uparrow}$ \\
\bottomrule
\end{tabular}
\end{table*}

On all 21 CLIPs, hard-negative margins widen in both directions and hubness decreases. On COCO val, the median share of text-to-image misses landing on a zero-shot hub falls from $28.1\%$ to $18.7\%$. The initial share on OpenAI B/16 is $41.2\%$. The two modalities still rarely mix among nearest neighbors, and matched-pair distances often increase. The matching item can move up in rank even as its absolute distance grows.

The agent's patches also lower image hubness on all 13 session checkpoints. Their gains and those of CSLS correlate with the original share of misses landing on hubs ($\rho=0.82$ and $0.95$, respectively). Section~\ref{supp:rerank} tests the two approaches together.

\subsection{Retrievals Before and After}
\label{supp:qual}
\begin{figure*}[t]
  \centering
  \includegraphics[width=\linewidth]{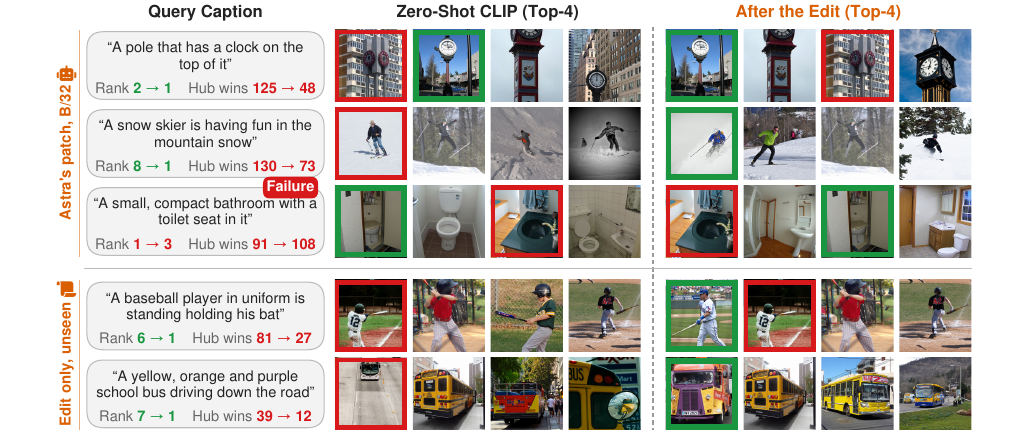}
\caption{\textbf{How a caption's top four images change.} The first three COCO val examples use \astra's OpenAI B/32 patch. The last two use \ours on B/32 DataComp-XL, which had no agent session. Green marks the labelled match and red a hub. Arrows give the match's rank and the number of the 25{,}000 captions retrieving the marked hub first, before and after the edit. In row three, the patch promotes a bathroom hub.}
  \label{fig:qual}
\end{figure*}

Figure~\ref{fig:qual} extends the main paper's examples with agent and recipe edits on COCO val.
Correct images move above hubs that previously attracted unrelated captions.
This pattern also appears on B/32 DataComp-XL, which had no agent session.
Across all validation captions, the agent's B/32 patch fixes 1{,}723 text-to-image R@1 errors while introducing 586, so corrected rankings substantially outnumber new errors.
Misses landing on hubs fall from 39\% to 23\%: 6{,}773 of 17{,}353 misses before editing, versus 3{,}675 of 16{,}216 afterward.
On DataComp-XL, \ours fixes 1{,}286 errors while introducing 554.

\section{Further Comparisons and Practical Use}
\label{supp:ft}
We compare ways to adapt the same checkpoint using paired training images, then ask what a reference bank can add at retrieval time.

\subsection{Learned-Map Construction and Selection}
\label{supp:map_protocol}

The model is a shared $d\times d$ map $A$ on both towers, trained on the same 5{,}000 construction images as \ours with either InfoNCE or a max-violation hinge loss~\cite{vsepp}.
We initialize $A=I$ and use full-batch Adam with learning rate $10^{-4}$, one caption per image per step and hinge margin $0.2$.
For each loss, we test penalties $\lambda\lVert A-I\rVert_F^2$ with $\lambda\in\{0,10^{-3},10^{-2},0.1,1\}$ and stopping steps in $\{25,50,100,200,300,500,750,1000,1500,2000,3000,4000\}$.
Of these 120 settings, we select the highest mean RSUM on 5{,}000 held-out training images over the six development checkpoints, subject to passing the accuracy guard on all six.
Forty hinge and six InfoNCE settings pass on all six; the selected schedule is the hinge loss without a penalty, stopped after 50 steps.

The loss matters for this result.
With InfoNCE, only the largest penalty passes the guard on all six checkpoints, and the best such schedule gains $+11.2$ COCO val RSUM, $5.0$\ci{4.6}{5.4} below \ours and behind it on every checkpoint.
InfoNCE depends on the temperature stored in each checkpoint, which the hinge loss does not use: on the two DFN checkpoints, whose released weights store a temperature of $0.07$, the fixed InfoNCE map gains only $+1.7$ and $-0.2$ COCO val RSUM.
For the hinge map, the guard affects which schedule is selected.
The setting with the highest development RSUM regardless of the guard, a penalty of $10^{-3}$ and 100 steps, fails the guard on 14 of the 21 checkpoints.
Keeping the selected zero penalty but choosing each checkpoint's stopping step by held-out retrieval RSUM alone fails on 8.

\noindent\textbf{Retrieval gain against ImageNet change.}
Table~\ref{tab:frontier} varies the two shrink strengths together for \ours and the number of steps for hinge. Both reach similar retrieval gains at small ImageNet losses. At twice its default strength, \ours gains $18.4$ COCO val RSUM and loses $0.68$ ImageNet points. At the largest tested strength, $s=0.30$, one round loses $8.4$ COCO val RSUM on these 21 checkpoints, while two rounds gain $16.3$. The second subspace overlaps the first by $0.54$, compared with $0.82$ at the default strength. For hinge, doubling the training steps has little effect on COCO val but lowers Flickr30k val retrieval and leaves only three guard passes.

\begin{table}[t]
\centering
\caption{\textbf{Sampled retrieval--accuracy tradeoffs.} Mean changes before fallback on 21 CLIPs: retrieval RSUM and ImageNet top-1 (IN, points). Guard counts individual checkpoints within the 0.5-point tolerance. \ours uses strengths $(s,s/2)$ with re-mining; the hinge map varies training steps without a penalty. $^{\ast}$The fixed defaults. This two-round sweep differs from the one-round sweep in Fig.~\ref{fig:dose}.}
\label{tab:frontier}
\tabfont
\begin{tabular*}{\linewidth}{@{\extracolsep{\fill}}l rrrr@{}}
\toprule
Setting & COCO val & Flickr30k val & IN & Guard \\
\midrule
\ours, $s=0.05$ & $+11.6$ & $+5.0$ & $+0.01$ & 21/21 \\
\ours, $s=0.075$ & $+14.3$ & $+5.8$ & $-0.05$ & 21/21 \\
\ours, $s=0.10^{\ast}$ & $+16.1$ & $+6.3$ & $-0.16$ & 21/21 \\
\ours, $s=0.15$ & $+17.9$ & $+6.5$ & $-0.38$ & 15/21 \\
\ours, $s=0.20$ & $+18.4$ & $+6.3$ & $-0.68$ & 4/21 \\
\ours, $s=0.30$ & $+16.3$ & $+4.6$ & $-1.49$ & 0/21 \\
\midrule
Hinge map, 25 steps & $+15.0$ & $+6.7$ & $-0.02$ & 21/21 \\
Hinge map, 50 steps$^{\ast}$ & $+17.1$ & $+6.8$ & $-0.22$ & 21/21 \\
Hinge map, 100 steps & $+17.2$ & $+5.5$ & $-0.68$ & 3/21 \\
Hinge map, 200 steps & $+12.8$ & $+1.6$ & $-1.78$ & 0/21 \\
Hinge map, 500 steps & $-1.6$ & $-10.2$ & $-4.84$ & 0/21 \\
\bottomrule
\end{tabular*}
\end{table}

\subsection{A Matched Comparison on Three Checkpoints}
\label{supp:panel}
Table~\ref{tab:panel} places the closest alternatives side by side on three checkpoints, all built from the 5{,}000 construction images of \ours and all passing the accuracy guard.
LoRA is retrained here on these images with the configuration of Table~\ref{tab:ft}; its stopping step is selected from $\{100,250,500,1000\}$ by COCO val RSUM among those that pass the guard.
The trained methods gain more on OpenAI B/16; the gaps are smaller on DataComp-XL B/16 and DFN-2B L/14.
The table also reports their different effects on SugarCrepe.

LoRA uses four COCO val and four ImageNet evaluations to choose its stopping step, and its final step fails the accuracy guard on both B/16 checkpoints.
On prepared inputs, training and these evaluations take 394--1{,}630\,s per checkpoint, excluding the zero-shot reference evaluation.
Construction and the shared original/edited guard pass take 63--243\,s for \ours; training alone takes 142--662\,s for LoRA, against 8--27\,s for constructing \ours.
These H200 measurements exclude loading and preprocessing, which are included in the separate deployment study of Sec.~\ref{supp:cost}.
LoRA timings come from one run per checkpoint, with OpenAI B/16 sharing the GPU during part of training.
For \ours, construction times are medians of three runs on an otherwise idle GPU, and the guard is timed once.

LoRA and KISSME-K improve add accuracy on all three checkpoints; the composition constraint in Sec.~\ref{supp:addcost} extends \ours to this objective.

\begin{table}[t]
\centering
\caption{\textbf{Matched construction data on three checkpoints.} Changes from zero-shot in retrieval RSUM and accuracy (points): ImageNet top-1 (IN), SugarCrepe add (Add) and overall (All). Every method uses the same 5{,}000 images and passes the guard. KISSME-K uses the extended grid; Fixed map uses the shared hinge schedule; Tuned map is fitted per checkpoint with InfoNCE, or with hinge where marked ($^{\ast}$). LoRA's stopping step is chosen on the reported COCO val set; Flickr30k and SugarCrepe are unused in selection.}
\label{tab:panel}
\tabdense
\begin{tabular*}{\linewidth}{@{\extracolsep{\fill}}l rr rrr@{}}
\toprule
Method & COCO val & Flickr30k val & IN & Add & All \\
\midrule
\multicolumn{6}{@{}l}{\textit{B/16 OpenAI}} \\
\ours & $+29.1$ & $+16.4$ & $-0.06$ & $+0.6$ & $+1.5$ \\
KISSME-K & $+8.4$ & $+6.1$ & $-0.28$ & $+2.8$ & $+1.4$ \\
Fixed map & $+32.6$ & $+18.7$ & $+0.50$ & $+4.6$ & $+2.8$ \\
Tuned map & $+38.7$ & $+21.3$ & $+0.22$ & $+8.3$ & $+4.7$ \\
LoRA & $+52.9$ & $+31.4$ & $-0.45$ & $+9.8$ & $+5.1$ \\
\midrule
\multicolumn{6}{@{}l}{\textit{B/16 DataComp-XL}} \\
\ours & $+16.4$ & $+7.9$ & $-0.30$ & $-3.3$ & $-0.2$ \\
KISSME-K & $+3.8$ & $+1.8$ & $-0.14$ & $+0.4$ & $+0.5$ \\
Fixed map & $+16.9$ & $+7.1$ & $-0.16$ & $-2.5$ & $+0.1$ \\
Tuned map & $+16.1$ & $+6.4$ & $-0.42$ & $-0.1$ & $+1.5$ \\
LoRA & $+16.3$ & $+8.1$ & $-0.36$ & $+1.9$ & $+1.9$ \\
\midrule
\multicolumn{6}{@{}l}{\textit{L/14 DFN-2B}} \\
\ours & $+13.1$ & $+4.4$ & $-0.14$ & $-3.0$ & $-0.5$ \\
KISSME-K & $+2.1$ & $+0.4$ & $-0.15$ & $+1.1$ & $+0.8$ \\
Fixed map & $+15.1$ & $+4.0$ & $-0.47$ & $-3.4$ & $+0.2$ \\
Tuned map$^{\ast}$ & $+15.2$ & $+4.5$ & $-0.42$ & $-3.3$ & $+0.2$ \\
LoRA & $+14.7$ & $+5.3$ & $-0.44$ & $+1.9$ & $+2.4$ \\
\bottomrule
\end{tabular*}
\end{table}

\subsection{Learning the Map with Gradients}
Table~\ref{tab:ft} gives the main paper's OpenAI B/16 comparison. All methods use 5{,}000 COCO training images and pass the guard. The learned maps share the construction images of \ours and select from a $5\times12$ grid per loss on held-out training pairs. LoRA and full fine-tuning use a separate subset of the same size. The shared map gains $9.6$\ci{8.0}{11.3} COCO val RSUM over \ours on OpenAI B/16 and $8.6$ on OpenAI B/32. The differences are smaller on DataComp-XL B/16 ($-0.3$\ci{-1.7}{+1.1}) and DFN-2B L/14 ($+2.2$\ci{+1.0}{+3.4} for hinge, with Flickr30k level).

The original B/16 LoRA run gains $46.9$ COCO test RSUM. Training takes 99\,s for LoRA and 114\,s for full fine-tuning, or 393\,s and 470\,s including checkpoint evaluations. These are single runs, with LoRA on a shared GPU. Section~\ref{supp:panel} compares LoRA and \ours on identical construction images.

\begin{table*}[t]
\centering
\caption{\textbf{Trained alternatives on OpenAI B/16.} Gains from zero-shot in validation RSUM (95\% CIs where available) and ImageNet top-1 (IN, points). Learned maps select a penalty and stopping step on 1{,}000 separate training images. LoRA and full fine-tuning select on the reported COCO val split. All selected models pass the guard.}
\label{tab:ft}
\tabfont\setlength{\tabcolsep}{6pt}
\begin{tabular}{@{}ll r@{\hspace{0.35em}}l r@{\hspace{0.35em}}l r@{}}
\toprule
Method & Tuning & \multicolumn{2}{c}{COCO val} & \multicolumn{2}{c}{Flickr30k val} & IN \\
\midrule
\ours (closed form) & None & $\mathbf{+29.1}$ & \cit{+27.1}{+31.1} & $\mathbf{+16.4}$ & \cit{+13.8}{+18.9} & $\mathbf{-0.06}$ \\
Learned $d{\times}d$ map, both towers & Gradient & $+38.7$ & \cit{+36.3}{+41.1} & $+21.3$ & \cit{+18.2}{+24.6} & $+0.22$ \\
Learned map, text side only & Gradient & $+40.0$ & \cit{+37.4}{+42.7} & $+20.3$ & \cit{+16.9}{+23.7} & $-0.42$ \\
LoRA & Gradient & $+48.8$ & \cit{+46.3}{+51.4} & $+29.5$ & \cit{+25.8}{+33.5} & $-0.19$ \\
Full fine-tuning & Gradient & $+58.8$ & & $+33.2$ & & $+0.07$ \\
\bottomrule
\end{tabular}
\end{table*}

\subsection{The KISSME-K Baseline}
\label{supp:kissme}
Table~\ref{tab:alt} compares \ours with KISSME~\cite{kissme}, a closed-form metric computed from similar and dissimilar pairs.
We first describe the cosine-adapted baseline used in the main paper, then test a larger strength grid and native distance scoring.
All variants operate in the full shared embedding space, on the eight checkpoints listed in Table~\ref{tab:cohorts}.

\noindent\textbf{Pairs.}
The similar pairs are the same 25{,}000 labelled construction pairs as \ours.
The dissimilar pairs are the zero-shot hard negatives of Eq.~\eqref{eq:D}: 5 wrong images per caption and 25 wrong captions per image, pooled with equal weight across directions.

\noindent\textbf{Metric.}
For a pair of an image and a caption, let $\delta=\hat e_v-\hat e_t$.
Let $\Sigma_{+}$ and $\Sigma_{-}$ be the means of $\delta\delta^{\top}$ over the similar and the dissimilar pairs; they are second moments, without centering.
Each is regularized by a ridge, $\Sigma\leftarrow\Sigma+10^{-3}\,(\mathrm{tr}\,\Sigma/d)\,I$.
The metric is
\[
M_{\mathrm{K}}=\mathrm{sym}\big(\Sigma_{+}^{-1}-\Sigma_{-}^{-1}\big).
\]

\noindent\textbf{Positive semidefiniteness.}
$M_{\mathrm{K}}$ can have negative eigenvalues.
Writing $M_{\mathrm{K}}=Q\,\mathrm{diag}(\mu)\,Q^{\top}$, we clip them at zero and take the square root, scaled to spectral norm one:
\[
L_{\mathrm{K}}=\frac{Q\,\mathrm{diag}\big(\sqrt{\max(\mu_i,0)}\big)\,Q^{\top}}{\max_i\sqrt{\max(\mu_i,0)}} .
\]

\noindent\textbf{Application and scoring.}
The baseline applies the map $S_{\mathrm{K}}=(1-\gamma)I+\gamma L_{\mathrm{K}}$ to both towers, in the same way as $S$ in Eq.~\eqref{eq:S}.
Retrieval then uses the cosine of the mapped embeddings, $(S_{\mathrm{K}}\hat e_v)^{\top}(S_{\mathrm{K}}\hat e_t)/(\lVert S_{\mathrm{K}}\hat e_v\rVert\,\lVert S_{\mathrm{K}}\hat e_t\rVert)$.

\noindent\textbf{Strength selection.}
For each checkpoint, choose $\gamma\in\{0.025,0.05,0.1,0.2,0.3,0.5\}$ by RSUM on 5{,}000 further COCO training images, subject to the full ImageNet guard.
If no strength passed, selection would take the smallest ImageNet loss; if the selected map failed to improve selection RSUM, it would return zero-shot.
Neither case occurred.
B/16 and B/32 OpenAI select $0.3$; the other six select $0.5$.
Retrieval validation and test data are unused in this choice.

\noindent\textbf{Larger strengths.}
Extending the grid with $\gamma\in\{0.6,0.7,0.8,0.9,1.0\}$ lets L/14 DFN-2B and two fresh checkpoints select $0.6$; none selects more.
Three checkpoints pass the guard at $0.6$ and $0.7$, and none at $0.8$ or above.
The expanded-grid gains are $+4.5$ COCO val and $+2.3$ Flickr30k val RSUM, versus $+4.3$ and $+2.5$ originally.
\ours remains ahead on all eight checkpoints, by $+14.2$\ci{+13.5}{+15.0} and $+5.3$\ci{+4.4}{+6.1} RSUM.
Selecting without the guard raises KISSME-K only to $+5.6$ and $+2.9$.
At $\gamma=1$, it loses 94 COCO val RSUM and 18.7 ImageNet points on average.

\noindent\textbf{Native scoring.}
KISSME originally ranks by the distance $\delta^{\top}M\delta$ rather than by a cosine.
We therefore also rank by this distance on the unit embeddings, without a map or renormalization, using $M_{\gamma}=(1-\gamma)I+\gamma\,M_{\mathrm{K}}^{+}/\lVert M_{\mathrm{K}}^{+}\rVert_2$, where $M_{\mathrm{K}}^{+}$ is $M_{\mathrm{K}}$ with its negative eigenvalues clipped at zero.
The clipping is needed because 44--53\% of the eigenvalues of $M_{\mathrm{K}}$ are negative.
At $\gamma=0$ the ranking equals that of the cosine; $\gamma=1$ uses the full PSD-clipped metric with no identity component.
ImageNet classification assigns each image to the nearest class embedding under the same distance.
With $\gamma$ selected as above over the extended grid, the checkpoints choose $0.7$ to $0.9$ and all pass the guard.
Native scoring gains $+3.0$\ci{+2.6}{+3.4} COCO val and $+2.1$\ci{+1.6}{+2.7} Flickr30k val RSUM, less than the cosine form, and \ours is ahead by $+15.8$\ci{+14.9}{+16.6} and $+5.4$\ci{+4.5}{+6.4}.
The full PSD-clipped metric, $\gamma=1$, loses 139 COCO val RSUM and 27 ImageNet points.
Interpolation with the identity helps under both scoring rules.
The advantage of \ours in Table~\ref{tab:alt} persists with the larger strength grid and with native distance scoring.

\subsection{Reference Banks and Re-ranking}
\label{supp:rerank}
Re-ranking corrects candidate scores at retrieval time, while \ours changes the representations that produce those scores.
Table~\ref{tab:alt} compares these approaches on the same eight checkpoints as KISSME-K.
Each reference bank contains 5{,}000 training images and their 25{,}000 captions, sampled from COCO or Flickr30k.
The COCO bank uses Karpathy restval images for the session checkpoints and the construction subset for the three fresh checkpoints; the Flickr30k bank uses its training split.
The banks contain no evaluation images.
\ours uses pair identities during construction. CSLS and QB-Norm use unpaired bank scores at application. NNN and DN also use labelled bank folds to set their parameters. In text-to-image retrieval the reference queries are bank captions. In image-to-text retrieval they are bank images.
\noindent\textbf{Choosing the re-ranking parameters.}
CSLS uses $k=50$ neighbors and QB-Norm uses temperature parameter $\beta=50$; the QB-Norm rows in Table~\ref{tab:alt} use dynamic inverted softmax (DIS).
These values were chosen on COCO val of OpenAI B/16 and then held fixed across checkpoints, so that comparison uses development feedback for these baselines.
NNN chooses its subtraction strength and neighbor count separately for each retrieval direction, checkpoint and bank using five folds of the labelled bank, with recall@1 as the selection objective.
Its grid is $\alpha\in\{0.25,0.375,\ldots,1.5\}$ and $k\in\{1,2,4,\ldots,512\}$.
DN subtracts $\lambda$ times the corresponding bank mean from each unit embedding and scores their inner product without re-normalization.
It selects $\lambda\in\{0,0.05,\ldots,1\}$ by RSUM on the same bank folds.
No parameter is chosen on Flickr30k val or on either test set.

\noindent\textbf{Combining the edit with re-ranking.}
For a combined method, queries, candidates and the reference bank all use the edited embeddings.
Table~\ref{tab:rerank_stack} reports the gain from adding \ours to each re-ranker with the same bank source.
Every pooled Flickr30k val contrast is positive, ranging from $+1.95$ to $+4.90$ RSUM, which gives the rounded $+2.0$--$+4.9$ range in the main paper.
These are means across checkpoints, rather than a claim that every individual checkpoint improves under every combination.
CSLS, QB-Norm and NNN retain bank-dependent candidate statistics; DN needs only the two bank means, so a full reference bank need not remain available at inference.
DN subtracts these means after unit normalization and then uses an inner product; moving the translation into a pre-normalization head bias would change that score in general.
\ours folds a linear map into the existing projections, and its gains remain useful when these score corrections are added.

\begin{table}[t]
\centering
\caption{\textbf{Adding \ours to re-ranking.} Mean Flickr30k val RSUM difference between the combined method and the re-ranker alone, on the eight checkpoints of Table~\ref{tab:alt}, with paired joint-bootstrap 95\% CIs. Bank sources match within each contrast. Every pooled interval is above zero.}
\label{tab:rerank_stack}
\tabfont
\begin{tabular*}{\linewidth}{@{\extracolsep{\fill}}l E E@{}}
\toprule
Re-ranker & \multicolumn{2}{c}{COCO bank} & \multicolumn{2}{c@{}}{Flickr30k bank} \\
\midrule
CSLS & $+3.12$ & \cit{+2.49}{+3.79} & $+3.31$ & \cit{+2.66}{+3.97} \\
QB-Norm (DIS) & $+3.70$ & \cit{+2.97}{+4.42} & $+1.95$ & \cit{+1.30}{+2.62} \\
NNN & $+3.75$ & \cit{+3.08}{+4.45} & $+2.32$ & \cit{+1.67}{+2.97} \\
DN & $+4.90$ & \cit{+4.11}{+5.71} & $+4.23$ & \cit{+3.51}{+4.98} \\
\bottomrule
\end{tabular*}
\end{table}

\subsection{How Paired Data Determine the Correction}
To understand what a pretrained model still allows an output-space edit to change, we examine its loss along a shared weight update.
For an InfoNCE loss $L$ evaluated on a specified set of pairs, let $W_v(\varepsilon)=(I+\varepsilon E)W_v$ and $W_t(\varepsilon)=(I+\varepsilon E)W_t$, where $E$ is any $d\times d$ matrix.
The chain rule gives
\begin{equation}
\label{eq:supp_stat}
\frac{dL}{d\varepsilon}\Big|_{\varepsilon = 0} = \big\langle \nabla_{W_v}L,\ E\,W_v\big\rangle + \big\langle \nabla_{W_t}L,\ E\,W_t\big\rangle .
\end{equation}
At a stationary point with respect to both projection matrices, $\nabla_{W_v}L=\nabla_{W_t}L=0$, so every shared-map derivative vanishes.
On 5{,}000 held-out CC3M~\cite{cc3m} web pairs, directions derived from the InfoNCE loss provide a first-order editing value comparable to that obtained from $\dD$ on COCO.
Their subspace differs (overlap 0.34; chance 0.25), and an edit constructed from the web-pair statistic gains $+4.2$ COCO val RSUM.
This measurement concerns the evaluation loss on these pairs, not stationarity of the original pretraining objective.
It illustrates why paired data matter: a residual can be present in different domains, while the useful editing directions depend on the retrieval task.

\subsection{Editing During and After Fine-Tuning}
An agent edit at step 250 of a 500-step full fine-tuning run of OpenAI B/16 adds $+2.1$\ci{+1.2}{+3.0} COCO val RSUM immediately.
After the remaining 250 steps, the advantage is $+0.45$\ci{-0.23}{+1.12} over three seeds, indicating that continued fine-tuning recovers much of the same gain.
For already fine-tuned models, NegCLIP~\cite{negclip} and a COCO fine-tune of B/16, the COCO residual is threefold smaller.
\ours still recovers 2.5\% and 3.9\% of the remaining RSUM headroom, respectively, compared with 6.9--13\% on 25 CLIPs.
Together, these measurements locate the correction in a residual that task-specific training also reduces.

\noindent\textbf{Scope of the adaptation comparison.}
Other approaches also address adaptation and retention: WiSE-FT interpolates pretrained and fine-tuned weights, and FLYP retains a contrastive training objective.\footnote{Wortsman et al., \emph{Robust Fine-Tuning of Zero-Shot Models}, CVPR 2022, arXiv:2109.01903; Goyal et al., \emph{Finetune Like You Pretrain: Improved Finetuning of Zero-Shot Vision Models}, CVPR 2023, arXiv:2212.00638.}
Those protocols are outside the present comparison.

\section{Constructing and Deploying \ours}
\label{supp:edit}
This section specifies the fixed recipe, its development data, the deployment guard and the cost of applying it.

\subsection{The Two-Round Recipe}
The procedure below uses the notation of main Sec.~\ref{sec:method}.
Its subspace size, shrink strengths and negative counts are fixed across checkpoints; only the map is recomputed.

\noindent\fbox{\begin{minipage}{0.96\linewidth}\small
\textbf{\ours} (inputs: frozen checkpoint; 5{,}000 training images with five captions each)\\[2pt]
1.~Embed the images and captions.\\
2.~$S \leftarrow I$. For round $r = 1, 2$ with strength $s_r = 0.10, 0.05$:\\
\hspace*{1em}a.~apply the current $S$ to the original embeddings and L2-normalize;\\
\hspace*{1em}b.~mine, on these embeddings, the 5 hardest wrong images per caption and the 25 hardest wrong captions per image;\\
\hspace*{1em}c.~compute $\dD$ (Eq.~\eqref{eq:D}) on these embeddings;\\
\hspace*{1em}d.~$U \leftarrow$ the $d/4$ eigenvectors of $\mathrm{sym}(\dD)$ with the smallest eigenvalues;\\
\hspace*{1em}e.~$S \leftarrow (I - s_r UU^{\top})\, S$.\\
3.~Fold: $W_v \leftarrow S W_v$, $W_t \leftarrow S W_t$.\\[2pt]
No model-specific tuning or backpropagation is used; $S$ is computed in closed form from the labelled pairs.
\end{minipage}}

On all 21 CLIPs and all six COCO construction samples, the $d/4$ selected eigenvalues are negative in both rounds. Across the original 21 first-round spectra, $42$--$48\%$ of the eigenvalues are negative. The first-round spectra in both specialized domains satisfy the same condition. The optimality result in Sec.~\ref{supp:theory} gives the uniform shrink as its maximizer when at least $d/4$ eigenvalues are negative. A new checkpoint with fewer would fall outside that guarantee at the fixed rank.

\subsection{Development and Application}
\label{supp:development}
Absence of hyperparameter search refers to applying the completed recipe.
A second-round pilot used B/16 OpenAI, DataComp-XL and LAION-2B; B/32 OpenAI; and L/14 OpenAI and DFN-2B, with 5{,}000 COCO restval construction images and the ImageNet 5k proxy.
The schedule and controls were then fixed before evaluation on the common training construction set.
All 18 original ablation checkpoints had appeared in earlier one-round analyses; the first five without sessions are therefore not prospective for development.
Of the 21 CLIPs, only the three fresh checkpoints were named and downloaded after the completed rule was fixed (Table~\ref{tab:protocol_roles}).

\begin{table*}[t]
\centering
\caption{\textbf{Which data determine each decision?} The three prospective checkpoints were downloaded after the rule was fixed. Test retrieval is measured after the edits are frozen. Full ImageNet validation accuracy determines deployment. Independent retention is measured on ImageNet-V2 and 13 further test sets (Sec.~\ref{supp:retention}).}
\label{tab:protocol_roles}
\tabfont
\begin{tabular}{@{}p{0.18\linewidth}p{0.40\linewidth}p{0.37\linewidth}@{}}
\toprule
Stage & Data or checkpoints & Role \\
\midrule
Original agent sessions & 35k COCO training images for construction; another 5k for selection; COCO val and ImageNet 5k proxy for feedback & Choose and accept edits within each of 13 sessions; no Flickr30k or retrieval-test feedback \\
Recipe development & Six-checkpoint pilot; 18 checkpoints previously examined in one-round analyses & Choose the two-round schedule and strengths; the 18-model evaluation is on previously studied models \\
Frozen construction & 5k COCO training images, five captions each, with the same rule on every checkpoint & Compute each checkpoint's map without a retrieval validation split \\
Deployment guard & Full ImageNet val, 50k labelled images and 80 prompts per class & Select two rounds, otherwise one round, otherwise zero-shot; the same set measures ImageNet retention \\
Final retrieval reporting & Sealed COCO test and Flickr30k test; correction maps fixed before evaluation & Measure retrieval gains; scores do not choose the deployed edit \\
Prospective transfer & Three checkpoints with revisions pinned before download & Evaluate the rule after development; these form a subset of the eight checkpoints without sessions \\
\bottomrule
\end{tabular}
\end{table*}

\subsection{The Accuracy Guard}
\label{supp:guard}
The deployment guard compares edited and zero-shot top-1 accuracy on all 50{,}000 ImageNet validation images, using 80 prompts per class.
An edit passes if its loss is at most 0.5 points.
The fixed order is two rounds, otherwise one round, otherwise zero-shot; it uses no retrieval scores.
All 21 original CLIP constructions deploy two rounds.
Neither the two-round nor the one-round edit passes for SigLIP2 B/16, which retains zero-shot (Sec.~\ref{supp:otherfamilies}).
The reported ImageNet changes measure this same deployment constraint; Sec.~\ref{supp:retention} evaluates retention independently on ImageNet-V2 and on 13 further classification sets.

\noindent\textbf{A retrospective split-half check.}
A fixed split places 25 images per class in each half (seed 0).
Selecting on either half deploys both rounds for all 21 CLIPs; opposite-half mean changes are $-0.15$ and $-0.16$ points, with worst loss $0.48$.
The hinge map deploys all edits in one direction, with two opposite-half losses of $0.56$ and $0.55$; swapping halves rejects those two edits.
This is a partition-stability check on previously examined data; independent retention is measured in Sec.~\ref{supp:retention}.

\noindent\textbf{The guard in agent sessions.}
During a session, the \texttt{test} tool reports the ImageNet change on a fixed subset of 5{,}000 validation images, and the rule for replacing the incumbent applies the 0.5-point threshold to this proxy (Sec.~\ref{supp:protocol}).
The ImageNet changes of all 13 final patches in Table~\ref{tab:zoo} were measured after the sessions on the full validation set and range from $-0.48$ to $+0.67$ points.
All 13 final patches pass the full-set check.

\subsection{Constructing the Optional Composition Constraint}
\label{supp:constraint_recipe}
We construct foils from COCO training captions: each foil $f$ adds an object not annotated in the photo, or an unstated color or material.
No SugarCrepe item is used in construction.
Let $\hat e_f$ be the unit embedding of the foil.
The statistic $\dD_{\mathrm{add}}=\mathrm{sym}\,\mathbb{E}[\hat e_v(\hat e_t-\hat e_f)^{\top}]$ measures agreement with the true caption relative to its foil.
By Statement~1 in Sec.~\ref{supp:theory}, $\langle M,\dD_{\mathrm{add}}\rangle$ is their mean bilinear margin and $\langle M-I,\dD_{\mathrm{add}}\rangle$ is the change caused by metric $M$.

Each round shrinks the directions associated with the $d/4$ smallest eigenvalues of $\mathrm{sym}(\dD)+\beta\,\dD_{\mathrm{add}}$.
We search for the smallest $\beta\ge0$ for which the round's metric satisfies $\langle M-I,\dD_{\mathrm{add}}\rangle\ge0$, preserving the mean bilinear caption-foil margin on the construction pairs.

\noindent\textbf{Choosing the coefficient.}
For a coefficient $\beta\ge0$, let $U_{\beta}$ contain the eigenvectors of $\mathrm{sym}(\dD)+\beta\,\dD_{\mathrm{add}}$ for its $d/4$ smallest eigenvalues, and let $a(\beta)=\langle U_{\beta}U_{\beta}^{\top},\dD_{\mathrm{add}}\rangle$.
The round's metric is $M=I-c\,U_{\beta}U_{\beta}^{\top}$, so $\langle M-I,\dD_{\mathrm{add}}\rangle=-c\,a(\beta)$ and the constraint reads $a(\beta)\le0$.
Since $U_{\beta}$ minimizes $\langle UU^{\top},\mathrm{sym}(\dD)+\beta\,\dD_{\mathrm{add}}\rangle$ over matrices $U$ with $d/4$ orthonormal columns, $a(\beta)$ does not increase with $\beta$, and we locate the feasibility boundary by bisection.

If $a(0)\le0$, the round is unchanged from \ours.
Otherwise, set $\beta=t\sigma$ with $\sigma=\lVert\mathrm{sym}(\dD)\rVert_F/\lVert\dD_{\mathrm{add}}\rVert_F$, start at $t=1$ and double until feasible or $t>10^6$.
A feasible bracket is refined by 60 bisection steps, returning its feasible upper endpoint without a stopping tolerance.
If doubling finds no feasible coefficient, the implementation uses the largest tried coefficient and flags the round as infeasible; it adds no further fallback.
This case did not occur.

All 42 rounds on the main cohort were feasible. With cached embeddings, median construction takes 2.6\,s (maximum 7.9\,s), compared with 1.3\,s for the default in this experiment.

Both rounds recompute the retrieval and foil statistics on their current normalized embeddings, with fresh retrieval negatives in round two.
The implementation supports $a(\beta)\le\rho a(0)$; the reported $\rho=0$ was fixed before the four prospective checkpoints were scored, after exploration on six session checkpoints.
The constraint is active and feasible in all eight prospective rounds, with returned $a(\beta)$ zero to numerical precision.

\subsection{Placing the Correction in the Weights}
\label{supp:placement}
For CLIP, fold $S$ into each existing output projection and, where present, its bias.
This preserves architecture and inference cost.
SigLIP's vision tower instead needs an added output layer (Sec.~\ref{supp:otherfamilies}).

An existing image index may make a text-only update useful.
To distinguish the placement of an edit from its strength, let $M=S^{\top}S$ denote its score metric.
For one round, $S = I-s\,UU^{\top}$ is symmetric, and the three placements give
\begin{equation}
\label{eq:supp_towers}
\begin{aligned}
\text{$S$ on both towers:}\quad & (S\hat e_v)^{\top}(S\hat e_t) = \hat e_v^{\top}M\,\hat e_t ,\\
\text{$S$ on one tower:}\quad & (S\hat e_v)^{\top}\hat e_t = \hat e_v^{\top}S\,\hat e_t ,\\
\text{$M$ on one tower:}\quad & \hat e_v^{\top}(M\hat e_t) = \hat e_v^{\top}M\,\hat e_t .
\end{aligned}
\end{equation}
The full $M$ on one tower recovers the two-tower numerator; $S$ alone gives a smaller contraction, $s$ instead of $2s-s^2$.
Its cosine normalization differs, however: the text-only version divides by $\lVert\hat e_v\rVert\,\lVert M\hat e_t\rVert$ rather than $\lVert S\hat e_v\rVert\,\lVert S\hat e_t\rVert$.
For two rounds, $S=S_2S_1$ need not be symmetric.
The first and third identities in Eq.~\eqref{eq:supp_towers} still hold with $M=S^{\top}S$; applying $S$ to only the image or text tower gives the cross-score matrix $S^{\top}$ or $S$, respectively.

Across 18 checkpoints, \ours gains $16.6$ COCO val RSUM with both towers edited (Table~\ref{tab:ablation}). Recomputing the recipe on the image or text tower alone yields $11.9$ or $11.7$. Applying the same two-tower map $S$ to just one tower yields $10.8$ or $10.3$, respectively.

For an existing image index, placing the full metric $M$ on the text tower keeps every indexed image vector valid. It retains $15.8$ of the $16.6$ COCO val gain and $6.3$ of the $6.7$ Flickr30k val gain, and all 18 checkpoints pass the guard. Placing $M$ on the image side gives $16.3$ COCO val RSUM and 13 guard passes.

\subsection{Re-mining and Construction Choices}
\label{supp:construction_controls}
To test why a second round helps, we separate refreshing the hard negatives from simply increasing the first shrink.
Let $P=UU^{\top}$ be the first-round projector.
Two shrinks along this same subspace give $(I-0.05P)(I-0.10P)=I-0.145P$.
This is the larger-step control in Table~\ref{tab:remine}; it is exactly equivalent to retaining the original directions for round two.
A separate control re-normalizes the edited embeddings and recomputes $\dD'$ while keeping the original negative identities.
\ours also refreshes those identities, so its second round responds to the wrong candidates that are now hardest.

\begin{table}[t]
\centering
\caption{\textbf{The contribution of re-mining.} Mean RSUM change from zero-shot on 21 CLIPs, before fallback. Guard counts checkpoints within the 0.5-point ImageNet tolerance. Larger step retains the original subspace; fixed negatives recomputes the statistic but keeps the original negative identities; re-mined refreshes both.}
\label{tab:remine}
\tabfont
\begin{tabular*}{\linewidth}{@{\extracolsep{\fill}}lrrr@{}}
\toprule
Correction & COCO val & Flickr30k val & Guard \\
\midrule
One round, $s=0.10$ & $+12.46$ & $+5.18$ & $19/21$ \\
Larger step, $s=0.145$ & $+11.98$ & $+5.00$ & $8/21$ \\
Round two, fixed negatives & $+11.98$ & $+4.96$ & $8/21$ \\
Round two, re-mined (\ours) & $+16.11$ & $+6.29$ & $21/21$ \\
\bottomrule
\end{tabular*}
\end{table}

After round one, the identities of 21\% of text-to-image negatives and 26\% of image-to-text negatives change. Re-mining produces a second subspace with mean overlap $0.82$ with the first, compared with $0.97$ when the negatives are held fixed (Sec.~\ref{supp:roundtwo}). Updating the negatives gains $4.13$\ci{3.64}{4.64} COCO val and $1.33$\ci{0.74}{1.97} Flickr30k val RSUM over keeping them fixed. Mean SugarCrepe add loss also falls from $5.3$ points after one round to $1.9$ after two. Only the retrieval comparison holds the rest of round two fixed.

The main ablations (Table~\ref{tab:ablation}, 18 checkpoints) use a seeded random orthonormal basis for the random-subspace control and the $d/4$ largest eigenvalues for the positive-direction control. A third round adds a $0.05$ shrink with fresh negatives. For the data-size ablation we draw images without replacement and retain all five captions per image. A single draw of 250 or 1{,}000 images gives $8.2$ or $12.9$ COCO val RSUM before fallback. At 1{,}000 images, 12 of 18 pass the guard, compared with all 18 at 5{,}000.

We also compare procedures after fallback using the same three draws per size (seeds 700--702) on all 21 checkpoints. At 250 images, COCO val gains are $4.8$ for \ours, $1.4$ for 50-step hinge and $7.6$ for hinge with a 25-step fallback. At 1{,}000 images the gains are $10.6$, $7.4$ and $13.1$. The fallback hinge schedule leads on COCO at both sizes. On Flickr30k val its gains ($2.2$ and $4.3$) are close to those of \ours ($1.9$ and $4.4$), with both paired intervals including zero. The 50-step hinge gains $0.1$ and $2.3$ on Flickr30k val.

Two rows of Table~\ref{tab:ablation} use separate controls. On 16 checkpoints, a one-round correction gains $3.4$ COCO val RSUM with random negatives and $11.0$ with hard negatives. Both choose strength on held-out training pairs under the guard and include fallbacks. The paired difference is $7.7$ RSUM. On OpenAI B/16, the PLS, CCA and Procrustes row ($\ddagger$) measures the incremental effect of a rank-8 fit on the earlier expert patch (R in Table~\ref{tab:zoo}).

\subsection{Construction and Deployment Time}
\label{supp:cost}
From prepared inputs on one B200, \ours encodes 5{,}000 images and 25{,}000 captions, mines negatives, computes both maps and folds them into the projections. OpenAI B/16 and L/14 take 7.92 and 23.60\,s on the first pass (7.21 and 22.44\,s, medians of seven further runs). These are the construction times in the main paper. Model loading, preprocessing and guard evaluation are timed separately below.

Table~\ref{tab:cost_components} follows deployment from raw files on an idle H200. Overlapping loading with encoding gives totals of 75\,s for B/16 and 252\,s for L/14. With stages run sequentially, the totals are 93 and 270\,s, of which the guard accounts for 78\% and 85\%. The original, one-round and two-round models share an encoder pass because only their outputs differ. Scoring all three takes 0.02\,s. Mining and computing the maps take about 0.2\,s. These measurements use a warm file cache. A cold-cache B/16 run on a shared GPU took 153\,s. Internal encoder edits require fresh features for the guard.

On cached embeddings the 50-step hinge fit takes 0.41/0.46\,s on B/16 and L/14, compared with 0.05/0.07\,s for \ours (three warmed runs on an idle H200). Substituting these fitting times into the same deployment stages changes the estimated total by less than 0.5\,s. The 200-step InfoNCE fit takes 0.9/1.1\,s and per-checkpoint schedule selection takes 62/84\,s. The fixed hinge schedule skips that selection. The original agent sessions took 3.6--12.1 hours each.

\begin{table}[t]
\centering
\caption{\textbf{Deployment cost from raw files, including the guard.} Seconds on OpenAI checkpoints: medians of three runs on an otherwise idle H200 (fp32, 16 loading workers, warm file cache). Staged totals run the listed operations sequentially; streamed totals overlap loading and encoding. Stage medians need not sum exactly to the total median.}
\label{tab:cost_components}
\tabfont
\begin{tabular*}{\linewidth}{@{\extracolsep{\fill}}lrr@{}}
\toprule
Stage & B/16 & L/14 \\
\midrule
Interpreter start and imports & $4.6$ & $4.1$ \\
Model loading & $2.4$ & $3.4$ \\
Construction data: read, decode, resize, tokenize & $4.4$ & $6.0$ \\
Construction: encode, both rounds, fold & $8.6$ & $26.8$ \\
Guard data: 50k images, 80k prompts & $18.1$ & $18.9$ \\
Guard encoding & $54.8$ & $210.3$ \\
Guard scoring: original, two-round, one-round & $0.02$ & $0.02$ \\
\midrule
End to end, staged & $93.4$ & $269.8$ \\
End to end, streamed & $75.5$ & $252.4$ \\
\bottomrule
\end{tabular*}
\end{table}

\section{From Agent Exploration to a Reusable Correction}
\label{supp:f1}
The B/32 session produced a short program that shrinks hard-negative directions in both projections. We trace how that edit recurs across sessions, how our refinement transfers it to new models, and what the agent found when it also edited the encoders.

\begin{table}[t]
\centering
\caption{\textbf{Sealed-test results from one agent session per checkpoint.} Z: zero-shot COCO test RSUM; A: \astra's final patch; R: the earlier fixed expert recipe, distinct from \ours. Retrieval columns show gains or paired differences on COCO test 5k and Flickr30k test 1k. $^{\circ}$The image-grouped 95\% CI includes zero; other retrieval differences exclude zero. IN is A's top-1 change on all 50k ImageNet validation images (points). Caveat denotes selection optimism in an otherwise valid patch audit. Rows are sorted by Z.}
\label{tab:zoo}
\tabdense
\begin{tabular*}{\linewidth}{@{\extracolsep{\fill}}l*{6}{r}l@{}}
\toprule
 & \hd{Z} & \multicolumn{3}{c}{COCO test 5k} & \hd{Flickr30k} & & \\
\cmidrule(lr){2-2}\cmidrule(lr){3-5}\cmidrule(lr){6-6}
Checkpoint & \hd{COCO} & \hd{A$-$Z} & \hd{R$-$Z} & \hd{A$-$R} & \hd{A$-$Z} & \hd{IN} & Audit \\
\midrule
B/32 OpenAI & $362.0$ & $+21.5$ & $+7.3$ & $+14.3$ & $+12.1$ & $+0.33$ & Caveat \\
B/16 DataComp-L & $362.3$ & $+11.0$ & $+4.4$ & $+6.6$ & $+6.8$ & $+0.03$ & Valid \\
B/16 OpenAI & $374.3$ & $+19.9$ & $+18.4$ & $+1.5$\mk{\circ} & $+11.9$ & $+0.67$ & Valid \\
L/14 OpenAI & $391.0$ & $+17.9$ & $+5.8$ & $+12.1$ & $+12.2$ & $-0.37$ & Valid \\
B/16 LAION-400M & $397.8$ & $+13.1$ & $+3.1$ & $+10.0$ & $+6.2$ & $-0.41$ & Valid \\
B/16 MetaCLIP-400M & $404.1$ & $+6.4$ & $+2.1$ & $+4.4$ & $+2.8$ & $-0.38$ & Valid \\
B/16 DataComp-XL & $408.4$ & $+10.4$ & $+3.6$ & $+6.7$ & $+3.2$ & $-0.33$ & Caveat \\
B/16 MetaCLIP-fullCC & $413.4$ & $+12.5$ & $+2.2$ & $+10.4$ & $+3.6$ & $-0.23$ & Caveat \\
B/16 LAION-2B & $417.1$ & $+11.2$ & $+3.0$ & $+8.1$ & $+5.7$ & $-0.48$ & Valid \\
B/16 DFN-2B & $424.4$ & $+8.3$ & $+0.2$\mk{\circ} & $+8.1$ & $+1.5$\mk{\circ} & $-0.21$ & Caveat \\
L/14 DataComp-XL & $433.0$ & $+6.2$ & $-0.4$\mk{\circ} & $+6.6$ & $+1.3$\mk{\circ} & $+0.09$ & Valid \\
L/14 LAION-2B & $435.6$ & $+10.1$ & $+2.0$ & $+8.1$ & $+4.6$ & $-0.08$ & Caveat \\
L/14 DFN-2B & $447.5$ & $+10.3$ & $+0.1$\mk{\circ} & $+10.2$ & $+2.9$ & $-0.06$ & Valid \\
\midrule
Mean & & $+12.2$ & $+4.0$ & $+8.2$ & $+5.8$ & & \\
\bottomrule
\end{tabular*}
\end{table}

\begin{figure*}[t]
  \centering
  \includegraphics[width=0.95\linewidth]{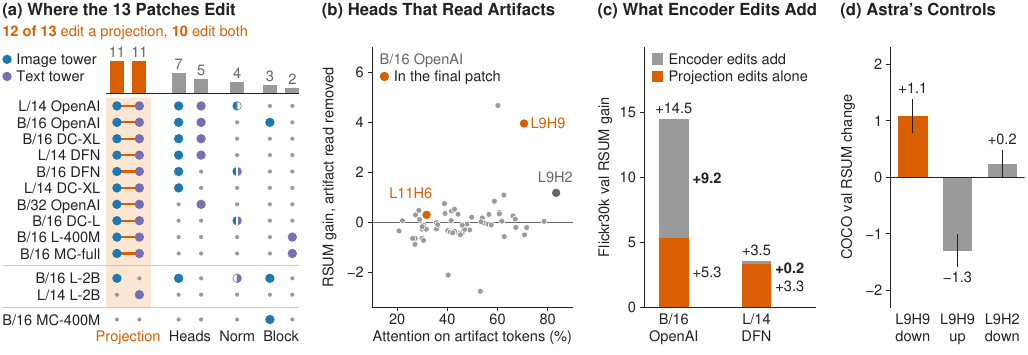}
\caption{\textbf{Where the agent edits and what changes retrieval.}
  (a) Final edit sites in the image (blue) and text (purple) towers. Norm denotes the final LayerNorm bias (image on the left, text on the right), and Block denotes a last-block rewrite.
  (b) For each OpenAI B/16 head, the horizontal axis gives the share of class-token attention on high-norm artifact tokens. The vertical axis gives the gain when its read of those tokens is removed, on 5{,}000 COCO restval images. Orange marks heads in the final patch.
  (c) Projection edits (orange) and the incremental gain from encoder edits (gray) sum to the full patch's Flickr30k val gain.
  (d) A later B/16 session tested scaling L9H9 in either direction and L9H2 down. Whiskers show the uncertainty reported by that session.}
  \label{fig:interp_full}
\end{figure*}

\subsection{Retrieval Gains and Recurring Edit Locations}
Table~\ref{tab:zoo} gives the 13 sessions' test results: mean gains of $+12.2$ COCO RSUM and $+5.8$ Flickr30k RSUM, significant on 13 and 11 checkpoints, respectively.
Twelve final patches rewrite a projection using $\dD$: nine attenuate negative directions and three rotate the towers.
The remaining patch edits the last vision MLP alone (Fig.~\ref{fig:interp_full}a).

\noindent\textbf{What the rotation patches change.}
A common orthogonal map on both towers leaves their cosine similarities unchanged.
The three rotation-based patches instead change the relative alignment of the towers.
On B/16 LAION-2B, the image projection alone receives $0.75I+0.25R$, where $R$ is an orthogonal Procrustes map that fixes the two modality means; this interpolation is generally nonorthogonal.
On L/14 DataComp-XL and B/16 DataComp-L, the projection maps are opposite orthogonal rotations, one on each tower.
For example, maps $Q^{\top}$ and $Q$ give cross-modal numerator $\hat e_v^{\top}Q^2\hat e_t$, which can change rankings while preserving each embedding's norm.
Figure~\ref{fig:interp_full}a shows where the complete patches also edit encoder components and biases.

\subsection{From Agent Sessions to a Fixed Recipe}
The B/32 program chose a uniform 10\% shrink over the lowest quarter of directions. On the 13 checkpoints with sessions, this one-round rule gains $13.7$ COCO val RSUM. Adding our fixed second round yields $18.2$ COCO test RSUM, compared with $12.2$ for the original patches. Every paired test difference favors \ours (Table~\ref{tab:ci_ckpt}).

\noindent\textbf{Search without the agent.}
We give two searches the same statistic, evaluation budgets and selection rule as the sessions. Their fixed grammar contains about 4{,}900 candidates per checkpoint, including spectral projection edits, rotations, mean and gap edits, and head scaling. The agent can also rewrite MLPs.

Over three seeds, uniform random search gains $12.8$ COCO val and $6.3$ Flickr30k val RSUM, compared with $12.0$ and $5.5$ for the agent. The Flickr30k test means before fallback are $5.7$ and $5.8$, respectively. All 39 random-search patches include a projection edit derived from $\mathrm{sym}(\dD)$, and 24 use the negative-direction shrink family seen in nine agent patches. The full ImageNet guard accepts all 13 agent patches and 34 of the 39 random-search patches. Reverting the other five gives random-search gains of $10.9$ COCO val and $5.4$ Flickr30k val RSUM. Ranking candidates by first-order margin instead favors large strengths, returns zero-shot on seven checkpoints and gains $7.5$ and $4.0$ on those validation sets. Both searches illustrate how the supplied statistic makes the projection correction accessible. The B/32 program specifies the particular shrink from which \ours was built.

\subsection{Edits Within the Encoders}
Encoder edits can provide gains beyond the projection correction on an individual model.
On OpenAI B/16, they contribute most of the Flickr30k validation improvement, adding $+9.2$ RSUM beyond the projection edits; on L/14 DFN-2B, they add $+0.2$ (Fig.~\ref{fig:interp_full}c).
A later-harness B/16 session tests its head choice with its own controls: reducing L9H9 helps, increasing it hurts, and reducing L9H2 has little effect (panel d).
The final B/16 patch halves the vision heads L11H6 and L9H9; that later session also selects L9H9.
Panel b compares these heads with our earlier artifact-token analysis, whose results were withheld from the prompt.
The prompt supplied the artifact-token hypothesis; the sessions selected the particular heads and tested their retrieval effects.
Section~\ref{supp:deep} tests whether the MLP rewrite, and a head correction from a control session, transfer as fixed procedures.

\subsection{The Contribution of Paired Evidence}
\label{supp:controls}
To distinguish the roles of weight access and paired evidence, we repeat sessions with either raw-weight reading or $\dD$ removed.
Table~\ref{tab:controls} reports these controls on OpenAI B/16 and DFN-2B L/14 using a later harness, comparing the arms within each checkpoint.
The harness is fixed within each comparison; differences from the original sessions also reflect harness changes and do not isolate session-to-session variability.

\begin{table*}[t]
\centering
\caption{\textbf{Agent controls under the later harness.} FULL: standard session; NOW: no weight reading (the server applies edits named by the agent); NOD: no $\dD$ or \texttt{crossmap}. Rows give RSUM gains from zero-shot or named contrasts with 95\% CIs. Each B/16 arm has two sessions; L/14 has two FULL and one each NOW/NOD. Pooled contrasts average checkpoints with a joint image-grouped bootstrap. Test scores follow completed sessions.}
\label{tab:controls}
\tabfont\setlength{\tabcolsep}{6pt}
\begin{tabular}{@{}ll r@{\hspace{0.35em}}l r@{\hspace{0.35em}}l r@{\hspace{0.35em}}l@{}}
\toprule
Checkpoint & Arm / contrast & \multicolumn{2}{c}{Flickr30k val} & \multicolumn{2}{c}{COCO val} & \multicolumn{2}{c@{}}{Flickr30k test} \\
\midrule
B/16 OpenAI & FULL & $+15.4$ & & $+26.5$ & & $+13.1$ & \\
 & NOW & $+14.9$ & & $+24.0$ & & $+14.2$ & \\
 & NOD & $+9.1$ & & $+12.0$ & & $+6.5$ & \\
 & FULL $-$ NOW & $+0.5$ & \cit{-0.8}{+1.8} & $+2.5$ & \cit{+1.5}{+3.5} & $-1.0$ & \cit{-2.3}{+0.2} \\
 & FULL $-$ NOD & $+6.3$ & \cit{+4.4}{+8.2} & $+14.5$ & \cit{+12.9}{+16.0} & $+6.7$ & \cit{+4.8}{+8.6} \\
\midrule
L/14 DFN-2B & FULL & $+2.9$ & & $+8.8$ & & $+2.2$ & \\
 & NOW & $+1.6$ & & $+6.3$ & & $+1.8$ & \\
 & NOD & $+1.2$ & & $+2.6$ & & $+1.0$ & \\
 & FULL $-$ NOW & $+1.3$ & \cit{+0.0}{+2.5} & $+2.5$ & \cit{+1.4}{+3.6} & $+0.4$ & \cit{-1.1}{+1.9} \\
 & FULL $-$ NOD & $+1.7$ & \cit{+0.3}{+3.1} & $+6.2$ & \cit{+5.0}{+7.4} & $+1.2$ & \cit{-0.3}{+2.8} \\
\midrule
Pooled & FULL $-$ NOW & $+0.9$ & \cit{+0.0}{+1.8} & $+2.5$ & \cit{+1.8}{+3.3} & \multicolumn{1}{c}{--} & \\
 & FULL $-$ NOD & $+4.0$ & \cit{+2.9}{+5.2} & $+10.3$ & \cit{+9.3}{+11.3} & \multicolumn{1}{c}{--} & \\
\bottomrule
\end{tabular}
\end{table*}

Removing $\dD$ costs more than removing weight reading on both models: pooled losses of $4.0$ versus $0.9$ Flickr30k val RSUM and $10.3$ versus $2.5$ COCO val RSUM.
Weight access adds $2.5$ COCO val RSUM on each checkpoint, but neither Flickr30k test contrast resolves an additional benefit (Table~\ref{tab:controls}).
On Flickr30k val, its B/16 effect lies within the $\pm2$ RSUM band of no detectable effect; the L/14 effect meets our criterion.

Neither B/16 session without $\dD$ constructs a hard-negative statistic.
One aligns projections with matched pairs alone; the other finds an encoder edit that does not transfer (Sec.~\ref{supp:deep}).
Paired evidence is the stronger contributor within this workflow, and the fixed recipe makes it usable without another session.

\subsection{A Fixed Expert Recipe as a Reference}
Table~\ref{tab:zoo} also compares the sessions with a fixed expert recipe (R), consisting of four closed-form components assembled on OpenAI B/16 in about 2.5 days and instantiated on other checkpoints by the same rules.
These components suppress vision-specific writing directions, rank heads using text descriptions, remove artifact-token reads from the selected heads, and filter the output projections.
One component, a projection filter, came from an earlier \astra session, so this is a reference for fixed-rule transfer rather than a purely human baseline.
The agent's model-specific patches outperform this recipe on COCO test on 12 of 13 checkpoints, by $+8.2$ RSUM on average.
On OpenAI B/16, where the reference recipe was developed, the difference is not significant.

\subsection{Agent Discovery with the Additional Objective}
\label{supp:agent_composition}
Four sessions receive $\dD_{\mathrm{add}}$ and a held-out add check.
Each meets both goals on its own checkpoint, gaining $+10.0$ to $+11.9$ COCO val RSUM and $+1.0$ to $+2.3$ points on the held-out add check, using different patches.
The recurring components are a protection subspace (4 of 4), a rotation of one tower (3 of 4), and metric reweighting (2 of 4).
We distill these components with the sessions' median constants and freeze the recipe before scoring the fresh checkpoints.
This recipe gains $+4.6$ COCO val RSUM on the deployed rows and fails the accuracy guard on two of four checkpoints.
For this additional objective, directly constraining the paired-statistic correction transfers more effectively than the shared components extracted from these sessions.

\subsection{Transfer of Encoder Edits}
\label{supp:deep}
Some encoder edits give substantial gains in individual sessions (Fig.~\ref{fig:interp_full}c).
To determine whether they should also enter a transferable recipe, we reapply the OpenAI B/16 session's largest encoder edit with its constants fixed.
It rewrites the last vision MLP's output matrix by mapping its contribution into embedding space, attenuating directions that favor hard negatives in proportion to their eigenvalues, and mapping back.

Let $W$ be the MLP output matrix and $T$ the centered linear map formed from the final LayerNorm gain and projection, used to express the MLP output in embedding coordinates, with pseudo-inverse $T^{+}$.
This direct-contribution approximation holds each example's LayerNorm normalization fixed; it does not model all downstream effects of changing the MLP.
Let $\dD_{\mathrm{mlp}}$ be the statistic in Eq.~\eqref{eq:D} restricted to this MLP's direct contribution, and let $\mathrm{sym}(\dD_{\mathrm{mlp}})$ have eigenvalues $e_1\le\dots\le e_d$ and orthonormal eigenvectors given by the columns of $Q$.
The rewrite is
\begin{equation}
\label{eq:supp_deep}
\begin{aligned}
W' &= W - 0.75\;T^{+}Q\,\mathrm{diag}(a)\,Q^{\top}T\,W,\\
a_i &= \max(-e_i, 0)\,/\,|e_1| ,
\end{aligned}
\end{equation}
where $e_1<0$ for the source edit and $a_i\in[0,1]$ is zero for non-negative eigenvalues and one for the smallest eigenvalue.
Mapping back through $T^{+}$ realizes the requested embedding-space change only within the range of $T$.

On each of five further checkpoints, we recompute $T$, $\dD_{\mathrm{mlp}}$ and the rewrite from that checkpoint's own weights and construction pairs, keeping the strength $0.75$ fixed.
An additional gain counts when DEEP+TB $-$ TB has a confidence interval above zero and passes the accuracy guard; the transfer criterion requires four of five checkpoints.
One meets it (Table~\ref{tab:deep}).
The rewrite alone improves the OpenAI checkpoints and LAION-2B B/16, but adds little beyond \ours on most models.
Its substantial overlap with the subspace already shrunk by \ours is consistent with this result.

\begin{table}[t]
\centering
\caption{\textbf{Transfer of the last-MLP rewrite.} Flickr30k val RSUM contrasts with paired 95\% CIs, one construction per checkpoint. DEEP: rewrite alone; TB: \ours; DEEP+TB: rewrite followed by \ours recomputed on the edited model; Z: zero-shot. $^{\circ}$Source checkpoint, excluded from the five-model transfer test. The last column gives eigenvalue-weighted overlap with TB's $d/4$ directions (chance $0.25$).}
\label{tab:deep}
\tabfont
\begin{tabular*}{\linewidth}{@{\extracolsep{\fill}}l E E r@{}}
\toprule
Checkpoint & \multicolumn{2}{c}{DEEP $-$ Z} & \multicolumn{2}{c}{DEEP+TB $-$ TB} & In TB's subspace \\
\midrule
B/16 OpenAI$^{\circ}$ & $+8.0$ & \cit{+5.7}{+10.3} & $+0.9$ & \cit{-0.5}{+2.3} & $0.60$ \\
B/32 OpenAI & $+5.6$ & \cit{+3.4}{+7.8} & $+0.4$ & \cit{-1.3}{+2.1} & $0.60$ \\
L/14 OpenAI & $+3.2$ & \cit{+1.0}{+5.3} & $+2.6$ & \cit{+1.2}{+4.0} & $0.60$ \\
B/16 LAION-2B & $+2.4$ & \cit{+0.6}{+4.1} & $-0.6$ & \cit{-1.9}{+0.6} & $0.56$ \\
L/14 DFN-2B & $-3.1$ & \cit{-5.7}{-0.5} & $-0.6$ & \cit{-2.0}{+0.8} & $0.42$ \\
B/32 LAION-2B & $-2.3$ & \cit{-5.4}{+0.6} & $-0.1$ & \cit{-1.7}{+1.5} & $0.59$ \\
\bottomrule
\end{tabular*}
\end{table}

\noindent\textbf{Transfer of a head-editing procedure.}
The B/16 OpenAI session without $\dD$ produces a last-block patch that gains $+8.8$ Flickr30k val RSUM on its source model.
It combines a reflection of L11H0's output, a value-bias shift for L11H5, and scaling of L11H9's key weights.
Recomputing these edits on the other seven B/16 checkpoints, with the head indices and strengths fixed, gives a mean gain of $-1.2$, compared with $+6.4$ for \ours.
These comparisons explain the choice of a projection recipe: it captures a correction that transfers across models, while the tested internal edits provide gains mainly in their source sessions.

\section{Agent Protocol and Paired Evidence}
\label{supp:protocol}
The harness fixes tools, permitted edits and feedback while leaving the choice of intervention to the agent.

\subsection{The Prompt and Paired Evidence}
\astra denotes GPT-6-Astra (configured as \texttt{gpt-6-astra}) operating through our harness with its native tools disabled. The prompt describes both towers, reports baseline recalls and lists known CLIP issues as hypotheses. The agent chooses which components to edit. Section~\ref{supp:inputs} reproduces the prompt excerpts, tool description and B/32 program.

For up to 16 chosen image/text component pairs, \texttt{crossmap} returns $d\times d$ blocks of our matched-versus-hard-negative statistic (Eq.~\eqref{eq:D}). It uses 5 wrong images per caption and 25 wrong captions per image. The B/32 program sums the supplied component blocks to recover $\dD$. The interface permits edits along their singular directions, shared output maps and re-mining on a patched model. Before these sessions we had used $\dD$ on OpenAI B/16 to choose components for expert patches. An earlier signed eight-direction edit gained $16.1$ COCO val RSUM. The prompt included neither result.

Nine sessions attenuated negative-eigenvalue directions. The B/32 program selected the lowest $d/4$ and applied the mild uniform shrink used in round one. All 13 sessions requested fresh negatives, mostly to re-check the margin. B/16 DFN-2B also used them for its final projection rewrite. We added a fixed second round with re-mining and strength $0.05$ (Sec.~\ref{sec:edit}).

\subsection{Tools and Data Access}
The tools let the agent inspect the model, construct an edit and assess its effect within fixed budgets:
\begin{center}
\begingroup
\tabfont
\begin{tabular}{@{}lp{0.55\linewidth}r@{}}
\toprule
Tool & Function & Calls \\
\midrule
\texttt{python} & Compute over weights without gradients & 300 \\
\texttt{probe} & Inspect activations and head contributions & 30 \\
\texttt{crossmap} & Obtain componentwise blocks of $\dD$ & 20 \\
\texttt{evaluate} & Score patches on the selection set & 100 \\
\texttt{test} & Score COCO val and the ImageNet proxy & 50 \\
\bottomrule
\end{tabular}
\endgroup
\end{center}
Edits are computed from the \emph{construction} set of 35{,}000 COCO Karpathy-train images with five captions each, which the agent can read item by item.
The \emph{selection} set contains 5{,}000 other Karpathy-train images; \texttt{evaluate} returns only aggregate scores on it.
The \texttt{test} tool evaluates COCO Karpathy val and a 5{,}000-image ImageNet proxy.
No tool reports Flickr30k, full ImageNet or SugarCrepe results.
These access rules describe the original 13 sessions.
The later controls in Sec.~\ref{supp:controls} use a different harness, and the sessions in Sec.~\ref{supp:agent_composition} use that later harness with $\dD_{\mathrm{add}}$ and a held-out composition check added; neither changes the protocol of the original sessions.

\subsection{Permitted Edits and Patch Selection}
The editing interface covers both output projections and internal encoder components, so the agent can test where a correction is most useful.
It supports seven kinds of edit:
\begin{center}
\begingroup
\tabfont
\begin{tabular}{@{}lp{0.84\linewidth}@{}}
\toprule
Kind & Permitted operation \\
\midrule
H1 & Low-rank update of a projection \\
H2 & Scale a head or remove a subspace from its output \\
H3 & Masked rank-one edit of an MLP output \\
H4 & Rank-one rewrite on a projection or head \\
H5 & Low-rank update of a block's attention or MLP output \\
H6 & Apply one shared matrix to both projections \\
H8 & Rewrite up to four whole tensors \\
\bottomrule
\end{tabular}
\endgroup
\end{center}
Together, H1--H6 edits may use total rank at most 8 and at most 12 heads.
The four whole-tensor rewrites allowed by H8 are counted separately.
Internal edits (H2, H3, H5, H4 on a head, and H8 on a block) are available in every session; the agent chooses whether to use them or edit the projections.
All edits must be computed in closed form; gradients, iterative fitting and regression to data targets are disallowed.

A tested patch replaces the current \emph{incumbent} only if it improves COCO validation RSUM by more than two paired standard errors and passes the accuracy guard: at most a 0.5-point drop on the ImageNet proxy.
The final incumbent is the session's output.

\subsection{Test Evaluation and Audit}
COCO Karpathy test and Flickr30k test are sealed: no tool can read them.
Each final patch is evaluated once after its session, with the evaluation rows committed before test data are loaded.
For the frozen \ours recipe, correction matrices are hash-locked before the single COCO test evaluation.
Flickr30k test results are reported for both agent-free methods and completed agent sessions; no test score is returned to the agent.

We statically audit every Python block for prohibited fitting operations and manually trace every array in the 13 final patches to the code that produced it.
All 13 patches satisfy the editing rules.
Table~\ref{tab:zoo} records 8 as valid and 5 as valid with a selection caveat: many development tests, including a panel of up to 40 candidates; an incumbent accepted by a narrow margin; or, in one session, an edit family tested after the session's own gate for it had failed.
These caveats concern selection optimism, not rule violations.
Each COCO test gain is within 1.5 RSUM of its development gain.

The original study has one session per checkpoint with a proprietary agent. Its intervals measure evaluation uncertainty.
Later repeated controls test evidence access within their harness, rather than rediscovery across the original 13 models.
The archived B/32 program reproduces the first-round projection edit. Section~\ref{supp:edit} specifies the complete two-round recipe.

\section{First-Order Analysis of the Correction}
\label{supp:theory}
For fixed negatives, $\mathrm{sym}(\dD)$ gives the gradient of the bilinear margin. With fixed embeddings and re-mined negatives, it gives a supergradient. A bounded linear optimization then gives the form of the shrink. The experiments determine the rank, strengths and number of rounds and test the resulting edit under cosine retrieval.

\subsection{How the Edit Changes Similarity}
Applying the same map $S$ to both towers changes unit image and caption embeddings $\hat e_v,\hat e_t\in\mathbb{R}^d$ into $S\hat e_v,S\hat e_t$.
Their dot product is $\hat e_v^{\top}S^{\top}S\hat e_t$.
Thus one matrix, $M=S^{\top}S$, describes the combined effect of the two edits on similarity.
We call $M$ the \emph{metric} and write the resulting bilinear score as $b_M(a,b)=a^{\top}Mb$.
Retrieval also normalizes the edited embeddings, giving
\begin{equation}
\label{eq:supp_cos}
\cos_M(\hat e_v,\hat e_t)=
\frac{\hat e_v^{\top}M\hat e_t}
{\sqrt{\hat e_v^{\top}M\hat e_v}\,\sqrt{\hat e_t^{\top}M\hat e_t}}.
\end{equation}
This expression is defined whenever both edited embeddings are nonzero.
Before editing, $M=I$ and the norms are one, so the bilinear and cosine scores coincide.

\noindent\textbf{What a shared shrink does.}
Let $U$ contain $k$ orthonormal editing directions, so $U^{\top}U=I$.
The map $S=I-sUU^{\top}$ scales each of these directions by $1-s$ and leaves the others unchanged.
For $0\le s\le1$, its metric is
\begin{equation}
\label{eq:supp_metric}
\begin{aligned}
M=S^{\top}S
&=I-2sUU^{\top}+s^2U(U^{\top}U)U^{\top}\\
&=I-cUU^{\top},\qquad c=2s-s^2.
\end{aligned}
\end{equation}
At $s=0.10$, the selected directions receive weight $0.81$ in the dot product; the others retain weight one.

\subsection{Why the Paired Statistic Guides the Edit}
We want matched pairs to score above the incorrect candidates that compete with them.
For now, hold these hard negatives fixed: $\mathcal{H}(y)$ contains the five hardest wrong images for caption $y$, and $\mathcal{H}(x)$ contains the 25 hardest wrong captions for image $x$.
Using the matched pairs $\mathcal{P}$, define their mean bilinear margin as
\begin{equation}
\label{eq:supp_margin}
\begin{aligned}
m(M)={}&\mathbb{E}_{\mathcal{P}}\big[b_M(\hat e_v(x),\hat e_t(y))\big]\\
&-\tfrac12\,\mathbb{E}_{y,\,x'\in\mathcal{H}(y)}
\big[b_M(\hat e_v(x'),\hat e_t(y))\big]\\
&-\tfrac12\,\mathbb{E}_{x,\,y'\in\mathcal{H}(x)}
\big[b_M(\hat e_v(x),\hat e_t(y'))\big].
\end{aligned}
\end{equation}

\noindent\textbf{Statement 1 (the statistic determines the margin).}
For fixed embeddings and negatives, the statistic $\dD$ in Eq.~\eqref{eq:D} gives, for every symmetric $M$,
\begin{equation}
\label{eq:supp_s1}
m(M)=\langle M,\dD\rangle
=\langle M,\mathrm{sym}(\dD)\rangle.
\end{equation}
Here $\langle A,B\rangle=\mathrm{tr}(A^{\top}B)$ is the matrix inner product and $\mathrm{sym}(\dD)=(\dD+\dD^{\top})/2$.

\noindent\emph{Proof.}
Each score can be written as $b_M(a,b)=\langle M,ab^{\top}\rangle$.
Taking the three averages in Eq.~\eqref{eq:supp_margin} collects their outer products into exactly $\dD$, giving $m(M)=\langle M,\dD\rangle$.
Since $M$ is symmetric, $\langle M,\dD\rangle=\langle M,\dD^{\top}\rangle$; averaging these expressions gives the second equality.
\hfill$\square$

\noindent\textbf{Why negative eigenvalues identify useful edits.}
At $M=I$, Eq.~\eqref{eq:supp_s1} gives $m(I)=\mathrm{tr}\,\dD$.
Let $\lambda_1\le\dots\le\lambda_d$ be the eigenvalues of $\mathrm{sym}(\dD)$, with orthonormal eigenvectors $u_1,\dots,u_d$.
Each $\lambda_i$ is the contribution of direction $u_i$ to this margin.
A negative contribution means that the direction favors hard negatives over matched pairs.
If $U=[u_1,\dots,u_k]$, shrinking these directions changes the margin by
\begin{equation}
\label{eq:supp_gain}
\begin{aligned}
m(M)-m(I)
&=-c\,\langle UU^{\top},\mathrm{sym}(\dD)\rangle\\
&=-c\sum_{i\le k}\lambda_i.
\end{aligned}
\end{equation}
For $c>0$, this gain is positive when the selected eigenvalues are negative: the edit reduces the influence of directions that were lowering the margin.
An additive step $M=I+\eta\,\mathrm{sym}(\dD)$ would also use this gradient, but would amplify positive directions and weight each change by its eigenvalue.
The contraction below instead bounds every directional change and allocates a fixed total budget.
Section~\ref{supp:uniform} tests eigenvalue-weighted contraction within the selected negative subspace; it does not evaluate the unrestricted additive step.

\subsection{Why the Shrink Is Uniform}
We bound contraction in each direction and in total.
Under these bounds, the best bilinear margin allocates contraction to the most negative directions first.

\noindent\textbf{Statement 2 (the optimal bounded contraction).}
Let $A$ be any symmetric $d\times d$ matrix with eigenvalues $\lambda_1\le\dots\le\lambda_d$ and orthonormal eigenvectors $u_1,\dots,u_d$.
For $0<c\le1$ and an integer $1\le k\le d$, consider
\begin{equation}
\label{eq:supp_K}
\begin{aligned}
\mathcal{K}=\big\{M=M^{\top}:\;&(1-c)I\preceq M\preceq I,\\
&\mathrm{tr}(I-M)\le ck\big\}.
\end{aligned}
\end{equation}
Here $A\preceq B$ means that $B-A$ is positive semidefinite.
The first constraint allows at most $c$ contraction in any direction; the trace constraint allows total contraction at most $ck$.
Then
\begin{equation}
\label{eq:supp_s2}
\max_{M\in\mathcal{K}}\langle M,A\rangle
=\mathrm{tr}A+c\sum_{i\le k}\max(-\lambda_i,0).
\end{equation}
If $\lambda_k<0$, a maximizer is $M^{\star}=I-cU_kU_k^{\top}$, where $U_k=[u_1,\dots,u_k]$.
It contracts the $k$ selected directions equally and leaves the rest unchanged.
Under this negative-eigenvalue condition, the maximizer is unique if $k=d$, or if $k<d$ and $\lambda_k<\lambda_{k+1}$.

\noindent\emph{Proof: express the contraction in each direction.}
Write $\Pi=(I-M)/c$.
The constraints become $0\preceq\Pi\preceq I$ and $\mathrm{tr}\,\Pi\le k$.
In the eigenbasis of $A$, let $\pi_i=u_i^{\top}\Pi u_i$ be the fraction of the allowed contraction used along $u_i$.
These fractions satisfy $0\le\pi_i\le1$ and $\sum_i\pi_i\le k$.
The objective is
\begin{equation}
\label{eq:supp_obj}
\langle M,A\rangle
=\mathrm{tr}A-c\langle\Pi,A\rangle
=\mathrm{tr}A-c\sum_{i=1}^{d}\lambda_i\pi_i.
\end{equation}
Thus contraction helps when $\lambda_i<0$, and its benefit is larger for more negative eigenvalues.

\noindent\emph{Allocate the contraction and attain the bound.}
An optimal allocation sets $\pi_i=1$ on the negative eigenvalues among the first $k$, and zero elsewhere.
To verify that no other allocation does better, set $\mu_i=\min(\lambda_i,0)$.
Since $\pi_i\ge0$, $\sum_i\lambda_i\pi_i\ge\sum_i\mu_i\pi_i$.
Also,
\begin{equation}
\label{eq:supp_lp}
\begin{aligned}
\sum_{i\le k}\mu_i-\sum_i\mu_i\pi_i
&=\sum_{i\le k}\mu_i(1-\pi_i)-\sum_{i>k}\mu_i\pi_i\\
&\le\mu_k\sum_{i\le k}(1-\pi_i)-\mu_k\sum_{i>k}\pi_i\\
&=\mu_k\Big(k-\sum_i\pi_i\Big)\le0.
\end{aligned}
\end{equation}
The first inequality follows from the ordered $\mu_i$ and $0\le\pi_i\le1$; the last follows from $\mu_k\le0$ and $\sum_i\pi_i\le k$.
Hence $\sum_i\lambda_i\pi_i\ge\sum_{i\le k}\mu_i$, which gives the upper bound in Eq.~\eqref{eq:supp_s2}.
It is attained by taking $\Pi$ to be the orthogonal projector onto the eigenvectors with $i\le k$ and $\lambda_i<0$.
When $\lambda_k<0$, this projector is $U_kU_k^{\top}$.

\noindent\emph{Uniqueness.}
If $\lambda_k<0$, equality in Eq.~\eqref{eq:supp_lp} requires $\sum_i\pi_i=k$.
If also $k<d$ and $\lambda_k<\lambda_{k+1}$, then $\mu_i>\mu_k$ for $i>k$, forcing $\pi_i=0$ there and $\pi_i=1$ for $i\le k$.
For $k=d$, the same trace condition forces all $\pi_i=1$.
These diagonal values determine the full matrix: for a positive semidefinite $B$, $u^{\top}Bu=0$ implies $Bu=0$.
Applying this to $\Pi$ and $I-\Pi$ gives $\Pi u_i=0$ for $i>k$ and $\Pi u_i=u_i$ for $i\le k$.
Thus $\Pi=U_kU_k^{\top}$.
\hfill$\square$

\noindent\textbf{Recovering the TowerBridge edit.}
Taking $A=\mathrm{sym}(\dD)$ makes the objective exactly the margin in Statement~1.
The recipe uses $k=d/4$ and $c=0.19$; the selected eigenvalues are negative on all 21 checkpoints.
By Eq.~\eqref{eq:supp_metric}, the optimal metric is realized by the shared map $S=I-0.10UU^{\top}$, recovering the agent's uniform shrink.
This is a variant of Ky Fan's maximum principle~\cite{fan1949weyl} with a trace bound. At the identity, the maximization in Statement~2 is also the linear subproblem of a Frank--Wolfe step for the bilinear margin. Round two builds a new subproblem on re-normalized embeddings (Sec.~\ref{supp:roundtwo}).
The result determines the metric $M$ for the chosen $k$ and $c$; the shared map is the recipe's way of applying it across the towers (Sec.~\ref{supp:edit}).

\noindent\textbf{Why both bounds matter.}
Without the total-contraction bound, every negative-eigenvalue direction would shrink by $c$.
Without the per-direction bound, retaining only $M\preceq I$ and the trace bound allows all contraction to be placed on $u_1$ when $\lambda_1<0$.
The per-direction bound also bounds each score change.
For unit $\hat e_v,\hat e_t$ and $M\in\mathcal{K}$,
\begin{equation}
\label{eq:supp_op}
\begin{aligned}
\big|b_M(\hat e_v,\hat e_t)-b_I(\hat e_v,\hat e_t)\big|
&=\big|\hat e_v^{\top}(M-I)\hat e_t\big|\\
&\le\lVert M-I\rVert_{\mathrm{op}}\le c.
\end{aligned}
\end{equation}
Here $\lVert\cdot\rVert_{\mathrm{op}}$ is the largest singular value.
A reversal of bilinear scores therefore requires an original bilinear score difference of at most $2c$.

\subsection{Why Round Two Needs Fresh Negatives}
\label{supp:roundtwo}
The first edit can change which wrong candidates are hardest.
Allowing the negative sets to change with $M$ shows how re-mining refreshes $\dD$.

\noindent\textbf{The margin with re-mined negatives.}
For a caption $y$, let $\mathcal{X}_y$ be its incorrect image candidates.
The mean score of its five hardest negatives is
\begin{equation}
\label{eq:supp_top}
h_y(M)=\max_{\substack{H\subseteq\mathcal{X}_y\\|H|=5}}
\frac{1}{5}\sum_{x'\in H}b_M\big(\hat e_v(x'),\hat e_t(y)\big).
\end{equation}
Define $h_x(M)$ in the same way for an image and its 25 hardest wrong captions.
Replacing the fixed-negative averages in Eq.~\eqref{eq:supp_margin} gives
\begin{equation}
\label{eq:supp_f}
\begin{aligned}
f(M)={}&\mathbb{E}_{\mathcal{P}}\big[b_M(\hat e_v(x),\hat e_t(y))\big]\\
&-\tfrac12\mathbb{E}_y[h_y(M)]-\tfrac12\mathbb{E}_x[h_x(M)].
\end{aligned}
\end{equation}
Unlike $m(M)$, this function chooses the hardest negatives for each metric.

\noindent\textbf{Statement 3 (the current statistic gives a linear upper bound).}
For fixed embeddings, $f$ is concave on the symmetric matrices.
Let $\dD(M_0)$ be the statistic formed from negatives mined under the current score $b_{M_0}$.
Then, for every symmetric $M$,
\begin{equation}
\label{eq:supp_s3}
f(M)\le f(M_0)
+\big\langle\mathrm{sym}(\dD(M_0)),M-M_0\big\rangle.
\end{equation}

\noindent\emph{Proof.}
For any fixed negative set, its mean score is linear in $M$.
Taking the maximum over these sets makes $h_y$ and $h_x$ convex.
The matched-pair term is linear, so subtracting the two convex terms makes $f$ concave.

To establish the bound, hold the negatives mined at $M_0$ fixed and call their margin $m_0(M)$.
Re-mining can only raise the negative scores, so $f(M)\le m_0(M)$, with equality at $M=M_0$.
Statement~1 gives $m_0(M)=\langle M,\mathrm{sym}(\dD(M_0))\rangle$.
Writing this linear function as its value at $M_0$ plus its change gives Eq.~\eqref{eq:supp_s3}.
\hfill$\square$

\noindent This bound makes $\mathrm{sym}(\dD(M_0))$ a \emph{supergradient}.
Where the mined sets remain unchanged locally, it is the ordinary gradient; the bound remains valid at ties.
It supplies first-order information for choosing an edit, although increasing the upper bound alone does not guarantee an increase in $f$.

\noindent\textbf{Applying the result after the first shrink.}
The recipe re-normalizes the edited embeddings before re-mining.
Write them as $z_v=S_1\hat e_v/\lVert S_1\hat e_v\rVert$ and $z_t=S_1\hat e_t/\lVert S_1\hat e_t\rVert$.
Apply Statement~3 with these updated unit embeddings as its fixed inputs, and call the resulting margin $f_2(M_2)$, where $M_2=S_2^{\top}S_2$.
At $M_2=I$, their dot product equals their cosine, so mining gives the recipe's $\dD'$ and
\[
f_2(M_2)\le f_2(I)+\langle\mathrm{sym}(\dD'),M_2-I\rangle.
\]
With the new negatives fixed, the inner product is the exact bilinear-margin change.
Statement~2 again gives a uniform shrink when the selected eigenvalues are negative, as on all 21 checkpoints.
This explains the second round's form, not its fixed strength $s_2=0.05$.

\noindent\textbf{Why retaining the old negatives is different.}
Without re-mining or re-normalization, the updated statistic would be $S_1\dD S_1$.
Its symmetric part shares the original eigenvectors because $S_1$ acts along those same directions:
\begin{equation}
\label{eq:supp_old}
S_1\mathrm{sym}(\dD)S_1u_i=
\begin{cases}
(1-s)^2\lambda_i u_i,&i\le k,\\
\lambda_i u_i,&i>k.
\end{cases}
\end{equation}
The selected eigenvalues are scaled by $0.81$; the others do not change.
Selecting the $k$ smallest therefore returns the same subspace unless a scaled eigenvalue rises above an unscaled one.
In the implemented control, which does re-normalize but retains the old negatives, the selected subspace also changes little: its mean overlap with the original is $0.97$ across 21 checkpoints, matching the value that reordering at the cut predicts (within $0.004$ on the 13 session checkpoints).
It gives no additional mean retrieval gain over a second shrink along the original subspace (the larger step in Fig.~\ref{fig:transfer}b, which equals $I-0.145\,UU^{\top}$).
Here overlap between orthonormal bases $U_a,U_b$ is $\lVert U_a^{\top}U_b\rVert_F^2/k$, equal to one for identical subspaces and $k/d$ on average for random ones.

\noindent\textbf{The combined map on the original embeddings.}
The two rounds can still be deployed as one weight edit, $S=S_2S_1$.
Intermediate normalization only rescales each embedding, so it does not change the final cosine after both maps are applied.
For the bilinear score before normalization, the combined metric is $M=S_1M_2S_1$, and
\begin{equation}
\label{eq:supp_r2}
b_M(\hat e_v,\hat e_t)
=(S_1\hat e_v)^{\top}M_2(S_1\hat e_t).
\end{equation}
This also makes $M_2\mapsto f(S_1M_2S_1)$ concave, since its argument is linear in $M_2$.
At $M_2=I$, Eq.~\eqref{eq:supp_s3} gives supergradient $S_1\mathrm{sym}(\dD(M_0))S_1$, with $M_0=S_1^2$.
Moving the two $S_1$ factors across the matrix inner product gives this expression.
This identity uses the original embeddings and bilinear mining under $M_0$; the recipe's $\dD'$ instead uses the normalized $z_v,z_t$ described above.

\subsection{Connection to Cosine Retrieval}
\label{supp:cosine_derivation}
The analysis so far isolates the change in the dot product.
Cosine retrieval divides this dot product by the two edited embedding norms, as shown in Eq.~\eqref{eq:supp_cos}.
To measure their contribution, consider a small metric change $M=I+\varepsilon E$, where $E$ is symmetric.
The numerator changes by $\varepsilon\hat e_v^{\top}E\hat e_t$, while each norm has expansion $1+\tfrac{\varepsilon}{2}\hat e^{\top}E\hat e+O(\varepsilon^2)$.
Differentiating gives
\begin{equation}
\label{eq:supp_dcos}
\begin{aligned}
\frac{d}{d\varepsilon}\cos_M(\hat e_v,\hat e_t)\Big|_{\varepsilon=0}
={}&\hat e_v^{\top}E\hat e_t\\
&-\tfrac12(\hat e_v^{\top}\hat e_t)
\big(\hat e_v^{\top}E\hat e_v+\hat e_t^{\top}E\hat e_t\big).
\end{aligned}
\end{equation}
The first term is the bilinear change; the second comes from normalization.
Averaging over matched pairs and fixed negatives gives the first-order cosine-margin change, with bilinear part $\langle E,\mathrm{sym}(\dD)\rangle$.

For the uniform edit, set $E=-cUU^{\top}$.
Across 30 models, the first-order cosine-margin change on the construction pairs is $0.85$--$1.19$ times the corresponding bilinear change.
Thus the measured effect of normalization is consistent with the improvement predicted by the paired statistic.
The optimality and concavity results concern the bilinear margin; Sec.~\ref{supp:uniform} tests the resulting uniform shrink directly under cosine retrieval.

The normalization-aware control in Sec.~\ref{supp:cosine} uses this derivative directly and tests its effect on retrieval.
\section{Complete Results, Cohorts and Uncertainty}
\label{supp:ci}
\begin{table}[!htbp]
\centering
\caption{\textbf{All six recall changes on sealed COCO test 5k.} Frozen two-round \ours minus zero-shot, in percentage points, for all 21 CLIPs. Z is zero-shot RSUM. Text-to-image retrieval accounts for 51\% of the total RSUM gain; every recall improves on every checkpoint. $^{\dagger}$No agent session.}
\label{tab:recall_all}
\tabdense
\begin{tabular*}{\linewidth}{@{\extracolsep{\fill}}l*{7}{r}@{}}
\toprule
 & & \multicolumn{3}{c}{Image $\to$ text} & \multicolumn{3}{c@{}}{Text $\to$ image} \\
\cmidrule(lr){3-5}\cmidrule(l){6-8}
Checkpoint & \hd{Z} & \hd{R@1} & \hd{R@5} & \hd{R@10} & \hd{R@1} & \hd{R@5} & \multicolumn{1}{c@{}}{R@10} \\
\midrule
B/16 DataComp-L & $362.3$ & $+3.2$ & $+3.8$ & $+2.5$ & $+2.1$ & $+2.8$ & $+2.6$ \\
B/16 DataComp-XL & $408.4$ & $+4.0$ & $+2.5$ & $+1.5$ & $+2.6$ & $+2.7$ & $+2.8$ \\
B/16 DFN-2B & $424.4$ & $+3.4$ & $+1.9$ & $+1.4$ & $+2.1$ & $+2.2$ & $+1.7$ \\
B/16 LAION-2B & $417.1$ & $+4.0$ & $+2.8$ & $+2.3$ & $+2.4$ & $+2.5$ & $+2.4$ \\
B/16 LAION-400M & $397.8$ & $+3.0$ & $+2.3$ & $+1.6$ & $+2.2$ & $+3.1$ & $+2.5$ \\
B/16 MetaCLIP-400M & $404.1$ & $+3.7$ & $+2.9$ & $+1.8$ & $+2.7$ & $+3.3$ & $+2.6$ \\
B/16 MetaCLIP-fullCC & $413.4$ & $+3.6$ & $+3.2$ & $+1.8$ & $+2.8$ & $+2.7$ & $+2.4$ \\
B/16 OpenAI & $374.3$ & $+3.9$ & $+4.1$ & $+3.2$ & $+5.7$ & $+6.4$ & $+5.9$ \\
B/32 OpenAI & $362.0$ & $+4.8$ & $+3.1$ & $+3.0$ & $+5.3$ & $+5.8$ & $+5.4$ \\
L/14 DataComp-XL & $433.0$ & $+3.5$ & $+2.5$ & $+1.7$ & $+2.3$ & $+2.5$ & $+2.1$ \\
L/14 DFN-2B & $447.5$ & $+3.7$ & $+2.1$ & $+1.6$ & $+2.2$ & $+2.2$ & $+1.7$ \\
L/14 LAION-2B & $435.6$ & $+3.5$ & $+2.9$ & $+1.7$ & $+2.1$ & $+2.1$ & $+1.9$ \\
L/14 OpenAI & $391.0$ & $+5.0$ & $+3.1$ & $+2.7$ & $+5.1$ & $+5.5$ & $+5.2$ \\
L/14 MetaCLIP-fullCC$^{\dagger}$ & $439.6$ & $+3.3$ & $+2.4$ & $+1.3$ & $+2.9$ & $+3.3$ & $+2.7$ \\
H/14 LAION-2B$^{\dagger}$ & $448.3$ & $+3.8$ & $+2.4$ & $+1.3$ & $+1.8$ & $+1.9$ & $+1.9$ \\
B/32 LAION-2B$^{\dagger}$ & $403.7$ & $+4.0$ & $+2.5$ & $+1.8$ & $+2.1$ & $+2.5$ & $+2.2$ \\
B/32 DataComp-XL$^{\dagger}$ & $389.5$ & $+4.1$ & $+2.6$ & $+1.8$ & $+2.5$ & $+3.1$ & $+3.2$ \\
H/14 DFN-5B$^{\dagger}$ & $467.8$ & $+3.1$ & $+1.6$ & $+0.7$ & $+2.4$ & $+1.8$ & $+1.4$ \\
B/32 LAION-2B (e16)$^{\dagger}$ & $401.4$ & $+3.6$ & $+2.7$ & $+1.8$ & $+2.6$ & $+2.6$ & $+2.4$ \\
L/14 CommonPool-XL$^{\dagger}$ & $437.9$ & $+4.1$ & $+2.1$ & $+1.3$ & $+2.1$ & $+2.5$ & $+1.9$ \\
H/14 MetaCLIP-2.5B$^{\dagger}$ & $447.7$ & $+3.5$ & $+1.5$ & $+1.3$ & $+2.8$ & $+2.6$ & $+2.4$ \\
\midrule
Mean (21) &  & $+3.8$ & $+2.6$ & $+1.8$ & $+2.8$ & $+3.1$ & $+2.7$ \\
Higher after edit &  & $21/21$ & $21/21$ & $21/21$ & $21/21$ & $21/21$ & $21/21$ \\
\bottomrule
\end{tabular*}
\end{table}

The main cohort comprises 13 checkpoints with agent sessions and eight without. Three of the eight were downloaded after the two-round rule was fixed (group F in Table~\ref{tab:cohorts}). In table order their pinned revisions are \texttt{825e019}, \texttt{f923444} and \texttt{28cc5ad}. The four prospective composition checkpoints form a separate group C.

Two 15-checkpoint comparisons exclude different development sets. For the fixed hinge schedule we exclude the six pilot checkpoints in Sec.~\ref{supp:development}. For the constrained edit we exclude the six used to explore constraint targets, B/16 OpenAI, DataComp-XL, DFN-2B and LAION-2B, and L/14 DFN-2B and LAION-2B.

Table~\ref{tab:recall_all} gives all six COCO test recall changes. Table~\ref{tab:ci_ckpt} compares the frozen edits on the same test images. On Flickr30k test, 11 of the 13 agent gains have intervals above zero. The other two are B/16 DFN-2B ($+1.5$\ci{-0.3}{+3.3}) and L/14 DataComp-XL ($+1.3$\ci{-0.1}{+2.8}).

\subsection{Uncertainty Protocol}
Retrieval intervals use a paired bootstrap over images with the candidate gallery fixed. Each sampled image carries its image-to-text query and its caption queries. Methods and checkpoints share the same resampled queries. ECCV Caption resamples its annotated queries within each direction, and SugarCrepe resamples items.

The pooled shrink, agent-control, composition, attention-head and other-family comparisons use 10{,}000 joint resamples; the pooled ECCV Caption, centering, mean-span and 13-checkpoint web intervals use 2{,}000, and the per-checkpoint agent-control intervals 5{,}000. We average checkpoints within each resample and sessions within each control arm (Table~\ref{tab:controls}). These intervals condition on the observed checkpoints, sessions and construction samples. Section~\ref{supp:samples} separately measures construction variation. The three-seed fine-tuning comparison in Sec.~\ref{supp:ft} uses a normal interval combining evaluation and training-seed variation.

On the three fresh checkpoints, two rounds gain $0.95$ Flickr30k val RSUM over one. Holding these checkpoints fixed gives a joint image-bootstrap interval of $[+0.21,+1.68]$ (normal approximation from the individual standard errors, $[+0.25,+1.64]$). Resampling the three checkpoints as well as their images widens the interval to $[-0.10,+2.05]$.

\begin{table*}[!tp]
\centering
\caption{\textbf{Estimates and 95\% confidence intervals for supporting comparisons.} Rows identify the metric, split and checkpoint cohort. A change is relative to zero-shot; a difference compares the named methods. Pooled estimates weight the listed checkpoints equally. Accuracy changes are in percentage points. Other-family intervals appear in Table~\ref{tab:nonclip} and Sec.~\ref{supp:otherfamilies}.}
\label{tab:ci}
\tabfont\setlength{\tabcolsep}{5pt}
\begin{tabular}{@{}lll r@{\hspace{0.35em}}l@{}}
\toprule
Quantity & Data set & Checkpoints & \multicolumn{2}{c@{}}{Estimate, 95\% CI} \\
\midrule
\multicolumn{5}{@{}l}{\textit{\ours edit}} \\
\gp RSUM change, mean & COCO val & 21 CLIPs & $+16.1$ & \cit{+15.4}{+16.8} \\
\gp Accuracy change & SugarCrepe overall & 21 CLIPs & $+0.06$ & \cit{-0.10}{+0.23} \\
\gp Accuracy change & SugarCrepe add & 21 CLIPs & $-1.9$ & \cit{-2.2}{-1.7} \\
\midrule
\multicolumn{5}{@{}l}{\textit{The \ours edit on the three fresh checkpoints}} \\
\gp RSUM change & Flickr30k val & B/32 LAION-2B (e16) & $+4.3$ & \cit{+2.4}{+6.2} \\
\gp RSUM change & Flickr30k val & L/14 CommonPool-XL & $+3.8$ & \cit{+2.2}{+5.4} \\
\gp RSUM change & Flickr30k val & H/14 MetaCLIP-2.5B & $+3.6$ & \cit{+2.0}{+5.2} \\
\gp RSUM difference, two rounds $-$ one round & Flickr30k val & The 3 fresh CLIPs, pooled & $+0.95$ & \cit{+0.21}{+1.68} \\
\gp Accuracy difference, two rounds $-$ one round & SugarCrepe add & The 3 fresh CLIPs, pooled & $+4.78$ & \cit{+4.25}{+5.33} \\
\midrule
\multicolumn{5}{@{}l}{\textit{Ablations of the \ours edit}} \\
\gp RSUM difference, uniform shrink $-$ eigenvalue-weighted step & COCO val & 13 CLIPs & $+8.0$ & \cit{+7.4}{+8.7} \\
\gp RSUM difference, re-mined $-$ fixed negatives in round two & COCO val & 21 CLIPs & $+4.13$ & \cit{+3.64}{+4.64} \\
\midrule
\multicolumn{5}{@{}l}{\textit{Constrained variant of the \ours edit}} \\
\gp RSUM difference, constrained variant $-$ \ours edit & Flickr30k val & 4 CLIPs & $-0.2$ & \cit{-0.7}{+0.3} \\
\gp Accuracy change of the constrained variant & SugarCrepe add & 4 CLIPs & $+0.2$ & \cit{-0.1}{+0.6} \\
\midrule
\multicolumn{5}{@{}l}{\textit{\astra sessions}} \\
\gp RSUM change, attention-head fix of one session applied to other checkpoints & Flickr30k val & The other 7 B/16 CLIPs & $-1.2$ & \cit{-1.8}{-0.6} \\
\gp RSUM lost by the agent without $\dD$ & Flickr30k val & B/16 OpenAI & $6.3$ & \cit{4.4}{8.2} \\
\gp RSUM lost by the agent without $\dD$ & Flickr30k val & L/14 DFN-2B & $1.7$ & \cit{0.3}{3.1} \\
\gp RSUM lost by the agent without weight reading & Flickr30k val & B/16 OpenAI & $0.5$ & \cit{-0.8}{1.8} \\
\gp RSUM lost by the agent without weight reading & Flickr30k val & L/14 DFN-2B & $1.3$ & \cit{0.0}{2.5} \\
\midrule
\multicolumn{5}{@{}l}{\textit{RSUM difference, \ours edit $-$ other method}} \\
\gp Text-only variant (25{,}000 labelled COCO pairs) & Flickr30k val & 8 CLIPs & $+0.2$ & \cit{-0.2}{+0.5} \\
\gp KISSME-K (original grid; 25{,}000 COCO pairs) & Flickr30k val & 8 CLIPs & $+5.1$ & \cit{+4.2}{+6.0} \\
\gp CSLS, COCO test-time bank & Flickr30k val & 8 CLIPs & $+1.1$ & \cit{+0.3}{+1.9} \\
\gp NNN, COCO test-time bank & Flickr30k val & 8 CLIPs & $+2.2$ & \cit{+1.1}{+3.4} \\
\gp QB-Norm, COCO test-time bank & Flickr30k val & 8 CLIPs & $+6.0$ & \cit{+4.6}{+7.3} \\
\gp DN, COCO test-time bank & Flickr30k val & 8 CLIPs & $+4.8$ & \cit{+4.0}{+5.6} \\
\gp CSLS, Flickr30k test-time bank & Flickr30k val & 8 CLIPs & $-2.1$ & \cit{-3.0}{-1.3} \\
\gp NNN, Flickr30k test-time bank & Flickr30k val & 8 CLIPs & $-3.6$ & \cit{-4.7}{-2.5} \\
\gp QB-Norm, Flickr30k test-time bank & Flickr30k val & 8 CLIPs & $-1.3$ & \cit{-2.5}{0.0} \\
\gp DN, Flickr30k test-time bank & Flickr30k val & 8 CLIPs & $+3.8$ & \cit{+3.0}{+4.6} \\
\bottomrule
\end{tabular}
\end{table*}

\begin{table*}[!tp]
\centering
\caption{\textbf{Checkpoint groups and analysis cohorts.} $n$ gives the checkpoint count. $^{a}$CommonPool-XL tags share prefix \texttt{commonpool\_xl\_}, with suffixes \texttt{s13b\_b90k} (A), \texttt{laion\_s13b\_b90k} (F), and \texttt{clip\_s13b\_b90k} (C). The g/14 LAION-2B tags are \texttt{laion2b\_s34b\_b88k} (A) and \texttt{laion2b\_s12b\_b42k} (C). $^{b}$MetaCLIP-fullCC and MetaCLIP-2.5B name the same pretraining data.}
\label{tab:cohorts}
\tabfont\renewcommand{\arraystretch}{1.0}
\begin{tabular}{@{}p{0.17\linewidth}rp{0.48\linewidth}p{0.26\linewidth}@{}}
\toprule
Group or cohort & $n$ & Checkpoints & Used in \\
\midrule
S: with agent sessions & 13 & B/32 OpenAI; B/16 OpenAI, LAION-400M, LAION-2B, DataComp-L, DataComp-XL, DFN-2B, MetaCLIP-400M, MetaCLIP-fullCC$^{b}$; L/14 OpenAI, LAION-2B, DataComp-XL, DFN-2B & Table~\ref{tab:zoo}, Fig.~\ref{fig:zoo}, Secs.~\ref{supp:shrink}, \ref{supp:uniform}, \ref{supp:bridge} and~\ref{supp:ft} \\
U: without sessions & 5 & L/14 MetaCLIP-fullCC$^{b}$, H/14 LAION-2B, B/32 LAION-2B, B/32 DataComp-XL, H/14 DFN-5B & Table~\ref{tab:transfer} \\
F: fresh & 3 & B/32 LAION-2B (e16), L/14 CommonPool-XL (LAION filter)$^{a}$, H/14 MetaCLIP-2.5B$^{b}$ & Table~\ref{tab:transfer}; chosen before download \\
A: additional & 4 & B/32 MetaCLIP-fullCC$^{b}$, L/14 CommonPool-XL$^{a}$, g/14 LAION-2B$^{a}$, ConvNeXt-Base-W LAION-2B & Only in the cohorts of 24, 25 and 30 \\
B: additional & 2 & B/16 CommonPool-L (CLIP filter), bigG/14 LAION-2B & Only in the cohort of 24 \\
C: constrained variant & 4 & B/32 MetaCLIP-400M, L/14 LAION-400M, L/14 CommonPool-XL (CLIP filter)$^{a}$, g/14 LAION-2B$^{a}$ & Table~\ref{tab:addcost} \\
O: other families & 5 & EVA02-B/16, MobileCLIP-B, CoCa ViT-B/32, SigLIP B/16, SigLIP2 B/16 & Tables~\ref{tab:nonclip} and~\ref{tab:bridge} \\
H: recent encoders & 4 & MetaCLIP 2 H/14, EVA02-E/14+, SigLIP2 So400m/14, Perception Encoder Core L/14 (336 px) & Table~\ref{tab:strong} \\
\midrule
Main evaluation & 21 & S + U + F & Table~\ref{tab:transfer}, Fig.~\ref{fig:transfer}a,b, Secs.~\ref{supp:f2}--\ref{supp:geometry} \\
Without sessions & 8 & U + F & Table~\ref{tab:transfer} \\
Ablations & 18 & S + U & Tables~\ref{tab:ablation} and~\ref{tab:gap}; Figs.~\ref{fig:anatomy}b and~\ref{fig:transfer}c; Sec.~\ref{supp:placement} \\
Random negatives & 16 & S + first three checkpoints in group U, in listed order & Table~\ref{tab:ablation}, row marked $\dagger$ \\
Shrink and removal & 24 & S + U + A + B & Secs.~\ref{supp:shrink} and~\ref{supp:centering}; Figs.~\ref{fig:anatomy}a and~\ref{fig:dose}; Table~\ref{tab:centering} \\
Theory checks, CLIP & 25 & S + U + F + A & Secs.~\ref{supp:shrink}, \ref{supp:shared_directions} and~\ref{supp:ft} \\
Theory checks, all & 30 & S + U + F + A + O & Sec.~\ref{supp:theory} (cosine versus bilinear first-order change) \\
Bank/metric baselines & 8 & B/16 OpenAI, B/32 OpenAI, L/14 OpenAI, B/16 LAION-2B, L/14 DFN-2B, and F & Table~\ref{tab:alt} \\
Constrained variant & 25 & S + U + F + C & Table~\ref{tab:addcost} (21 main, 4 prospective) \\

\bottomrule
\end{tabular}
\end{table*}

\begin{table*}[!tp]
\centering
\caption{\textbf{Paired COCO test RSUM contrasts.} Each column compares two methods on the same test images. Entries are RSUM differences with image-grouped 95\% CIs. All intervals are above zero. Agent comparisons apply to the first 13 checkpoints.}
\label{tab:ci_ckpt}
\tabfont\renewcommand{\arraystretch}{1.0}\setlength{\tabcolsep}{8pt}
\begin{tabular}{@{}l r@{\hspace{0.35em}}l r@{\hspace{0.35em}}l r@{\hspace{0.35em}}l@{}}
\toprule
Checkpoint & \multicolumn{2}{c}{\astra's patch $-$ zero-shot} & \multicolumn{2}{c}{\ours $-$ zero-shot} & \multicolumn{2}{c@{}}{\ours $-$ \astra's patch} \\
\midrule
B/32 OpenAI & $+21.5$ & \cit{+19.5}{+23.6} & $+27.5$ & \cit{+25.6}{+29.4} & $+5.9$ & \cit{+4.6}{+7.3} \\
B/16 OpenAI & $+19.9$ & \cit{+17.8}{+21.9} & $+29.3$ & \cit{+27.3}{+31.2} & $+9.4$ & \cit{+7.7}{+11.1} \\
L/14 OpenAI & $+17.9$ & \cit{+16.0}{+19.9} & $+26.7$ & \cit{+24.8}{+28.5} & $+8.8$ & \cit{+7.2}{+10.3} \\
B/16 LAION-400M & $+13.1$ & \cit{+11.4}{+14.7} & $+14.7$ & \cit{+13.2}{+16.2} & $+1.6$ & \cit{+0.4}{+2.7} \\
B/16 LAION-2B & $+11.2$ & \cit{+9.6}{+12.7} & $+16.4$ & \cit{+14.8}{+18.0} & $+5.2$ & \cit{+4.0}{+6.4} \\
B/16 DataComp-L & $+11.0$ & \cit{+9.1}{+12.9} & $+17.1$ & \cit{+15.4}{+18.7} & $+6.1$ & \cit{+4.8}{+7.4} \\
B/16 DataComp-XL & $+10.4$ & \cit{+8.7}{+12.1} & $+16.1$ & \cit{+14.6}{+17.7} & $+5.8$ & \cit{+4.5}{+7.1} \\
B/16 DFN-2B & $+8.3$ & \cit{+7.0}{+9.7} & $+12.8$ & \cit{+11.3}{+14.3} & $+4.4$ & \cit{+3.3}{+5.6} \\
B/16 MetaCLIP-400M & $+6.4$ & \cit{+5.0}{+7.9} & $+17.0$ & \cit{+15.4}{+18.6} & $+10.5$ & \cit{+9.2}{+11.9} \\
B/16 MetaCLIP-fullCC & $+12.5$ & \cit{+10.8}{+14.3} & $+16.4$ & \cit{+14.8}{+18.0} & $+3.8$ & \cit{+2.7}{+5.0} \\
L/14 LAION-2B & $+10.1$ & \cit{+8.5}{+11.7} & $+14.1$ & \cit{+12.7}{+15.5} & $+4.0$ & \cit{+2.9}{+5.2} \\
L/14 DataComp-XL & $+6.2$ & \cit{+4.9}{+7.5} & $+14.7$ & \cit{+13.3}{+16.1} & $+8.5$ & \cit{+7.2}{+9.8} \\
L/14 DFN-2B & $+10.3$ & \cit{+9.0}{+11.8} & $+13.5$ & \cit{+12.0}{+14.9} & $+3.1$ & \cit{+2.2}{+4.0} \\
\midrule
L/14 MetaCLIP-fullCC & \multicolumn{1}{c}{--} &  & $+15.8$ & \cit{+14.3}{+17.3} & \multicolumn{1}{c}{--} &  \\
H/14 LAION-2B & \multicolumn{1}{c}{--} &  & $+13.1$ & \cit{+11.8}{+14.4} & \multicolumn{1}{c}{--} &  \\
B/32 LAION-2B & \multicolumn{1}{c}{--} &  & $+15.1$ & \cit{+13.6}{+16.7} & \multicolumn{1}{c}{--} &  \\
B/32 DataComp-XL & \multicolumn{1}{c}{--} &  & $+17.3$ & \cit{+15.7}{+18.9} & \multicolumn{1}{c}{--} &  \\
H/14 DFN-5B & \multicolumn{1}{c}{--} &  & $+11.1$ & \cit{+9.8}{+12.3} & \multicolumn{1}{c}{--} &  \\
B/32 LAION-2B (e16) & \multicolumn{1}{c}{--} &  & $+15.7$ & \cit{+14.2}{+17.3} & \multicolumn{1}{c}{--} &  \\
L/14 CommonPool-XL & \multicolumn{1}{c}{--} &  & $+13.9$ & \cit{+12.5}{+15.3} & \multicolumn{1}{c}{--} &  \\
H/14 MetaCLIP-2.5B & \multicolumn{1}{c}{--} &  & $+14.2$ & \cit{+12.7}{+15.6} & \multicolumn{1}{c}{--} &  \\
\bottomrule
\end{tabular}
\end{table*}

\twocolumn
\makeatletter
\global\@colht\textheight
\global\@colroom\textheight
\global\vsize\textheight
\makeatother
\section{Discovery Inputs and Agent Program}
\label{supp:inputs}
The B/32 program below implements the first-round projection edit. To show what evidence was available when the agent wrote it, we first reproduce four of the 35 paragraphs in its prompt. The excerpts and program are verbatim apart from line wrapping.

\noindent\textbf{The session prompt (excerpt).}
The other 12 sessions use the same instructions with the checkpoint name, architecture and zero-shot COCO val recalls replaced. The system prompt is \texttt{Answer the user's request.} The remaining paragraphs cover tool access, budgets, indexing, analysis and patch selection (Sec.~\ref{supp:protocol}). We will release the complete prompts and session logs with the code. The first excerpt gives the task.

{\scriptsize
\begin{verbatim}
TASK. You are given the weights of an image-text retrieval
model, CLIP ViT-B/32 (the original OpenAI weights). It has two
towers that share no weights. The image tower is a vision
transformer: a 224x224 image is cut into 7x7 patches of 32x32
pixels, plus one class token; 12 layers (blocks) of width 768,
each with 12 attention heads of size 64 and an MLP of width
3,072 (QuickGELU, x * sigmoid(1.702 x), in both towers),
LayerNorm before each sublayer; the class token's final state
goes through a final LayerNorm and a 768x512 projection. The
text tower is a causal transformer over at most 77 BPE tokens
(vocabulary 49,408): 12 layers of width 512, each with 8
attention heads of size 64 and an MLP of width 2,048,
LayerNorm before each sublayer; the state at the end-of-text
token goes through a final LayerNorm and a 512x512 projection.
Both towers end in the same 512-dimensional space, and an
image and a text are matched by the cosine similarity of their
embeddings. Your job is to UNIFY THE TWO TOWERS by EDITING
THEIR WEIGHTS DIRECTLY. Each tower writes into the shared
space through its components (attention heads, MLP neurons,
the projection). Read the pretrained weights of BOTH towers,
find which image-tower components and which text-tower
components express the same content and which disagree or
write tower-specific content, and edit them so that the two
towers agree: one shared representation that generalises to
any dataset of captioned photos, not just COCO.
\end{verbatim}
}

\noindent The second listing is the list of known issues, which the prompt presents as hypotheses.

{\scriptsize
\begin{verbatim}
KNOWN ISSUES OF CLIP-LIKE MODELS. Published work reports
several general issues of CLIP-like image-text models. They
are hypotheses to check, not established facts about this
model and not the answer: check whether each applies here
before relying on it; some may not matter here, and something
not listed may matter more. Retrieval on COCO does not test
every one of them (counting, negation and attribute binding,
for example, are rare in its captions). (1) The modality gap:
image embeddings and text embeddings have been reported to
occupy two separate narrow cones of the shared space (Liang et
al. 2022). (2) Hubness: a few items are the nearest neighbours
of many queries, so they are retrieved for queries they do not
match (Radovanovic et al. 2010; Bogolin et al. 2022). (3)
High-norm artifact tokens: in some vision transformers a few
patch tokens acquire very large norms, carry little local
content and are read by attention as sinks, as if they were
registers (Darcet et al. 2024); whether this happens in this
model, and whether it matters for retrieval, is to be checked.
(4) Bag-of-words behaviour: weak sensitivity to word order and
to which attribute belongs to which object (Yuksekgonul et al.
2023). (5) Weak counting and weak negation. (6) A short
effective text length: it has been reported that most of the
information comes from the first 20 or so tokens of a text
(Zhang et al. 2024, Long-CLIP). (7) Tower-specific dimensions:
directions of the shared space that one tower writes and the
other ignores.
\end{verbatim}
}

\noindent\textbf{The description of \texttt{crossmap}.}
The third listing is the paragraph of the prompt that describes \texttt{crossmap}; the \texttt{pair\_D} matrices it mentions are the blocks of $\dD$.

{\scriptsize
\begin{verbatim}
6. A crossmap tool with a budget of 20 maps: the
correspondence between the two towers' components. Reply with
your explanation followed by exactly one block that starts
with ```crossmap and contains JSON: {"mode": "data", "weights"
or "both" (default "both"), "vision": [component, ...] or
{"layers": [l, ...]}, "text": the same for the text tower,
"k": the number of output directions per component in the
weights map (1 to 64, default 8), "pairs": [[image-tower
component, text-tower component], ...] (optional, at most 16;
data map only), "patch": {...} (optional: map the patched
model instead of the unedited one), "negatives": "fixed" or
"remined" (optional, with a "patch" and the data map; default
"fixed", see below)}. Components are named "LxHy" (head y of
block x, for example "L3H1"), "Lxmlp" (the MLP of block x),
"embed" (the input embedding at the class / end-of-text token)
and "biases" (the attention output biases and the final
LayerNorm's bias); {"layers": [l, ...]} means every head and
the MLP of those blocks ({"layers": "all"} for all twelve;
more names with "add": [...]); at most 160 per tower;
everything of a tower you do not name is collected into one
more component, "rest". The data map (modes "data" and "both";
the photos and captions of the construction half of the
discovery data only; it uses a fixed subsample of 5,000
construction photos with their 25,000 captions, the same in
every map and listed in index.json, since the per-component
arrays of all 35,000 photos would not fit; hard negatives are
mined among these): an item's embedding is the exact sum of
its components' direct contributions (as in a probe's
head_contrib arrays), each divided by the embedding's norm, so
the cosine of a photo and a caption is the sum over all pairs
(i, j) of an image-tower component i and a text-tower
component j of <c_i(photo), c_j(caption)>. This is a
decomposition of one forward pass, not a causal effect: a
direct contribution leaves out everything the component
changes in later blocks of its tower (their inputs, and the
final LayerNorm's statistics), so editing a component can
change the map in ways its own entry does not show. The map
gives match[i, j], the mean of <c_i, c_j> over matched
photo-caption pairs, and M[i, j] = match[i, j] minus the same
mean over hard negatives (each caption with the 5 wrong photos
that score highest for it, and each photo with the 25 wrong
captions that score highest for it), averaged over the two
directions (also given separately): a positive M[i, j] means
the pair adds more similarity to right pairs than to the
hardest wrong ones, a negative one the opposite, and M sums to
the mean hard-negative margin. A negative entry is evidence
about this margin on these photos and captions, not proof that
a component is defective (some "wrong" pairs are valid
matches). For a patched model the hard negatives are by
default those of the unedited model, kept fixed, since moving
negatives change the quantity being compared (the map is then
labelled "fixed negatives (original model)"); with
"negatives": "remined" they are mined again with the patched
model instead (labelled "re-mined negatives (patched model)").
The map also gives each component's mean contribution, and for
each of "pairs" the 512 x 512 matrix (matched minus
hard-negative mean of c_i c_j^T) whose trace is M[i, j]; its
singular vectors are the shared-space directions along which
the pair carries its evidence on these data. In this session
the data map, like any closed-form statistic of the discovery
data, may also enter an edit directly (for example a singular
direction of a pair's matrix; see RULES). The weights map
(modes "weights" and "both"; no data): each named head or MLP
is taken as A = shared_write(tower, layer, head or "mlp") (its
output weights carried into the shared space, below) with its
k strongest output directions (the top k left singular vectors
of A and their singular values). W_sub[i, j] is the mean
squared cosine of the principal angles between the k
directions of i and the k directions of j, and W_energy[i, j]
the same overlap weighted by the squared singular values (the
cosine between the two truncated output covariances A_k
A_k^T); beside each, a null: the same quantity with the text
component's k directions replaced by random directions of the
shared space (200 draws; its mean and 95th percentile), since
subspaces overlap by chance. Both say where two components can
write the same content in the shared space, not whether they
do on real inputs. Everything is saved in /work/crossmap_<k>/
(crossmap.npz, README.txt with every definition, index.json)
and you get a short summary; a data map takes a minute or two,
a weights map seconds. A request that is not valid or fails
does not count against the budget.
\end{verbatim}
}

\noindent The fourth listing gives the editing rules.

{\scriptsize
\begin{verbatim}
RULES. This session is GRADIENT-FREE AND DATA-ADAPTIVE: its
results will be reported as gradient-free, data-adaptive
editing, not as editing from the weights alone. (a) Where the
numbers of an edit may come from. Every numerical vector,
matrix or mask in an edit (u, v, q, A, B, a, b, mask, and
every H8 file) must be computed in closed form from the
weights of either tower or both and from statistics of the
discovery data, of probe activations and of the crossmap's
data map, which may be used directly in the edit. Allowed
statistics: means and differences of means, covariances and
cross-covariances, correlations, whitening (inverse square
roots of covariances), and eigen- and singular-value
decompositions of any of these, including of paired
photo-caption statistics and of the data map (for example its
pair_D matrices); products of such statistics with weight
matrices; and masks built by ranking or thresholding a
statistic. Closed-form moment estimators built from these are
allowed, for example an orthogonal (Procrustes-type) map from
the singular value decomposition of a photo-caption
cross-covariance, or a canonical-correlation-type map from
whitened cross-covariances; if you use one, say so in your
explanation and name it as such. (b) Not allowed, in any step
of a computation whose result enters an edit: gradients of any
kind (backward passes, autograd, finite-difference estimates
of the gradient of a score with respect to many numbers);
optimizers and iterative fitting (SGD, Adam, alternating or
EM-style updates, iterative solvers, any loop that improves a
loss or a score step by step); regression or least squares
against targets (lstsq, normal equations, a pseudo-inverse or
a linear solve of data against data targets, ridge, logistic
or any other regression, and estimators that fit by regression
or by iteration, such as scikit-learn's CCA, PLS, regressors,
KMeans or ICA); and fitting any vector or matrix to a loss.
Pseudo-inverses and linear solves of weight matrices, and
inverses or inverse square roots of covariances for whitening,
are not regression and are allowed. (c) Choosing. Discovery
data, probes, the crossmap, describe, evaluations and tests
may also be used to choose: which components to edit, which of
several candidates to use, and ONE strength per edit (its
alpha or scale; for H6 one common factor that multiplies the
whole alpha list; for H8 at most one scalar, as in WRITING
WHOLE MATRICES), from a dose-response or from a formula.
Measuring many settings and keeping the best is selection: use
it for these choices, not to tune many numbers of an edit one
at a time. (d) Validity. A patch with any array that breaks
(a) or (b), in any of its edits, makes the session invalid as
a success, whatever its scores; this is judged after the
session from your code (every Python run and every script in
/work is kept, see RECORD OF YOUR WORK), independently of the
results. If you are unsure whether a construction is allowed,
do not use it in an edit. (e) No running of the model other
than through the probe, evaluate, test and crossmap tools: do
not implement or simulate either tower's computation on any
input in the kernel; multiplying probe activations by weight
matrices, to analyse what a component writes or to carry a
statistic of activations into the space of a weight matrix, is
allowed; chaining layers to compute the network's output on an
input is not. (f) No internet access or search (there is
none). (g) No change to the model other than through a patch
of H1 to H6 and H8 edits. Allowed: any analysis of the weights
and of probe, map, evaluation and test results; products and
decompositions of the two towers' weights with each other (for
example shared_write of components of both towers); the
closed-form statistics of (a) inside edits; writing whole
matrices for H8 from them; and any patch you compute that is
in scope and follows (a) to (c).
\end{verbatim}
}

\noindent\textbf{The ViT-B/32 session's program.}
The listing is the program, written by the agent, that builds the projection edit of its final patch.
It sums the 16 blocks returned by the session's second \texttt{crossmap} call, computed on the unedited model, into the whole-embedding statistic, and writes the two replacement projections, which the patch installs as whole-tensor (H8) edits.
The code uses the stored projection layout, so right-multiplication by the transposed map implements Eq.~\eqref{eq:S}.
The final patch also scales one text-tower head, L10H3, by $1.5$.
In the listing, a line that continues the previous one is indented four further spaces.

{\scriptsize
\begin{verbatim}
# Hypothesis: attenuating 128 original negative-evidence
    directions preserves useful shared discrimination.
from pathlib import Path
import numpy as np
from scipy.linalg import eigh

work = Path("/work")
with np.load(work / "crossmap_2/crossmap.npz") as source:
    D = source["pair_D"].astype(np.float64).sum(axis=0)
eigenvalues, eigenvectors = eigh((D + D.T) / 2)
assert np.all(eigenvalues[:128] < 0)
weights = np.zeros(512)
weights[:128] = 1.0
negative_projector = (eigenvectors * weights[None, :]) @
    eigenvectors.T
rank128_transform = np.eye(512) - 0.1 * negative_projector

rank128_relative_updates = {}
for tower, tensor_name in [
    ("vision", "visual.proj"),
    ("text", "text_projection"),
]:
    original = load_original(tensor_name).astype(np.float64)
    replacement = (original @
        rank128_transform.T).astype(np.float32)
    target = work / f"spectral_rank128_0.1_{tower}.npy"
    assert replacement.shape == original.shape
    assert np.isfinite(replacement).all()
    assert np.linalg.norm(replacement) <= 3 *
        np.linalg.norm(original)
    if target.exists():
        assert np.array_equal(np.load(target), replacement),
            "Tested-array reconstruction mismatch"
    np.save(target, replacement)
    rank128_relative_updates[tower] = float(
        np.linalg.norm(replacement - original) /
            np.linalg.norm(original)
    )
\end{verbatim}
}
\end{document}